%% file: iclr2016_initialization.tex
\documentclass{article} % 
\usepackage{iclr2016_conference,times}
\usepackage{url}
\iclrfinalcopy
\usepackage{graphicx}
\usepackage{amsmath}
\usepackage{amssymb}
\usepackage{booktabs}
\usepackage{multirow,array}
\usepackage{color}
\usepackage{bm}
\usepackage{subcaption}
\usepackage[font=footnotesize,subrefformat=parens]{subcaption}
\usepackage{tabu}           % 
\usepackage{xcolor}            % 
\usepackage{algorithm}
\usepackage{algorithmicx}
\usepackage{algpseudocode}
\usepackage{subcaption}
\usepackage{pgfplots}
\usepackage{tikz}
\usepackage{array}
\usepackage{xspace}
\usepackage{fnpos}

\usepackage[pagebackref=true,breaklinks=true,letterpaper=true,colorlinks,bookmarks=false,urlcolor=ta4skyblue]{hyperref}

\definecolor{Plum}{RGB}{142, 69, 133}
\definecolor{Cyan}{RGB}{0, 255, 255} % 
\definecolor{Red3}{HTML}{a40000}
\definecolor{Green3}{HTML}{4e9a06} % 

\definecolor{tabutter}{rgb}{0.98824, 0.91373, 0.30980}      % 
\definecolor{ta2butter}{rgb}{0.92941, 0.83137, 0}     % 
\definecolor{ta3butter}{rgb}{0.76863, 0.62745, 0}     % 

\definecolor{taorange}{rgb}{0.98824, 0.68627, 0.24314}      % 
\definecolor{ta2orange}{rgb}{0.96078, 0.47451, 0}     % 
\definecolor{ta3orange}{rgb}{0.80784, 0.36078, 0}     % 

\definecolor{tachocolate}{rgb}{0.91373, 0.72549, 0.43137}   % 
\definecolor{ta2chocolate}{rgb}{0.75686, 0.49020, 0.066667} % 
\definecolor{ta3chocolate}{rgb}{0.56078, 0.34902, 0.0078431}   % 

\definecolor{tachameleon}{rgb}{0.54118, 0.88627, 0.20392}   % 
\definecolor{ta2chameleon}{rgb}{0.45098, 0.82353, 0.086275} % 
\definecolor{ta3chameleon}{rgb}{0.30588, 0.60392, 0.023529} % 

\definecolor{taskyblue}{rgb}{0.44706, 0.56078, 0.81176}     % 
\definecolor{ta2skyblue}{rgb}{0.20392, 0.39608, 0.64314} % 
\definecolor{ta3skyblue}{rgb}{0.12549, 0.29020, 0.52941} % 
\definecolor{ta4skyblue}{rgb}{0.06274, 0.14510, 0.26470} % 

\definecolor{taplum}{rgb}{0.67843, 0.49804, 0.65882}     % 
\definecolor{ta2plum}{rgb}{0.45882, 0.31373, 0.48235}    % 
\definecolor{ta3plum}{rgb}{0.36078, 0.20784, 0.4}     % 

\definecolor{tascarletred}{rgb}{0.93725, 0.16078, 0.16078}  % 
\definecolor{ta2scarletred}{rgb}{0.8, 0, 0}        % 
\definecolor{ta3scarletred}{rgb}{0.64314, 0, 0}       % 

\definecolor{taaluminium}{rgb}{0.93333, 0.93333, 0.92549}   % 
\definecolor{ta2aluminium}{rgb}{0.82745, 0.84314, 0.81176}  % 
\definecolor{ta3aluminium}{rgb}{0.72941, 0.74118, 0.71373}  % 

\definecolor{tagray}{rgb}{0.53333, 0.54118, 0.52157}     % 
\definecolor{ta2gray}{rgb}{0.33333, 0.34118, 0.32549}    % 
\definecolor{ta3gray}{rgb}{0.18039, 0.20392, 0.21176}    % 

\newcommand{\reffig}[1]{Figure~\ref{fig:#1}}

\newcommand{\refsec}[1]{Section~\ref{sec:#1}}

\newcommand{\reftbl}[1]{Table~\ref{tbl:#1}}
\newcommand{\refalg}[1]{Algorithm~\ref{alg:#1}}

\newcommand{\shortrefsec}[1]{\S~\ref{sec:#1}}
\newcommand{\refeq}[1]{Equation~\eqref{eq:#1}}

\newcommand{\lblfig}[1]{\label{fig:#1}}
\newcommand{\lblsec}[1]{\label{sec:#1}}
\newcommand{\lbleq}[1]{\label{eq:#1}}
\newcommand{\lbltbl}[1]{\label{tbl:#1}}
\newcommand{\lblalg}[1]{\label{alg:#1}}

\newcommand{\etal}{\mbox{\emph{et al.\ }}}

\newcommand{\expect}[2]{\mathbb{E}_{#1}\left[#2\right]}
\newcommand{\expectbig}[2]{\mathbb{E}_{#1}\bigg[#2\bigg]}

\newcommand{\kmeans}{$k$-means\xspace}

\title{Data-dependent Initializations of \\ Convolutional Neural Networks}

\author{Philipp Kr\"ahenb\"uhl$^{1}$, Carl Doersch$^{1,2}$, Jeff Donahue$^{1}$, Trevor Darrell$^{1}$\\
$^{1}$Department of Electrical Engineering and Computer Science, UC Berkeley\\
$^{2}$Machine Learning Department, Carnegie Mellon\\
\texttt{\{philkr,jdonahue,trevor\}@eecs.berkeley.edu; cdoersch@cs.cmu.edu} \\
}

\begin{document}

\maketitle

\begin{abstract}
Convolutional Neural Networks spread through computer vision like a wildfire, impacting almost all visual tasks imaginable.
Despite this, few researchers dare to train their models from scratch.
Most work builds on one of a handful of ImageNet pre-trained models, and fine-tunes or adapts these for specific tasks.
This is in large part due to the difficulty of properly initializing these networks from scratch.
A small miscalibration of the initial weights leads to vanishing or exploding gradients, as well as poor convergence properties.
In this work we present a fast and simple data-dependent initialization procedure, that sets the weights of a network such that all units in the network train at roughly the same rate, avoiding vanishing or exploding gradients.
Our initialization matches the current state-of-the-art unsupervised or self-supervised pre-training methods on standard computer vision tasks, such as image classification and object detection, while reducing the pre-training time by three orders of magnitude.
When combined with pre-training methods, our initialization significantly outperforms prior work, narrowing the gap between supervised and unsupervised pre-training.
\let\thefootnote\relax\footnote{Code available: \url{https://github.com/philkr/magic_init}}
\end{abstract}
\section{Introduction}

In recent years, Convolutional Neural Networks (CNNs) have improved performance across a wide variety of computer vision tasks~\citep{googlenet,vgg,fastrcnn}.
Much of this improvement stems from the ability of CNNs to use large datasets better than previous methods.
In fact, good performance seems to require large datasets: the best-performing methods usually begin by ``pre-training'' CNNs to solve the million-image ImageNet classification challenge~\citep{imagenet}.
This ``pre-trained'' representation is then ``fine-tuned'' on a smaller dataset where the target labels may be more expensive to obtain.
These fine-tuning datasets generally do not fully constrain the CNN learning: different initializations can be trained until they achieve equally high training-set performance, but they will often perform very differently at test time.
For example, initialization via ImageNet pre-training is known to produce a better-performing network at test time across many problems.
However, little else is known about which other factors affect a CNN's generalization performance when trained on small datasets.
There is a pressing need to understand these factors, first because we can potentially exploit them to improve performance on tasks where few labels are available.
Second they may already be confounding our attempts to evaluate pre-training methods.
A pre-trained network which extracts useful semantic information but cannot be fine-tuned for spurious reasons can be easily overlooked.
Hence, this work aims to explore how to better fine-tune CNNs.
We show that simple statistical properties of the network, which can be easily measured using training data, can have a significant impact on test time performance.
Surprisingly, we show that controlling for these statistical properties leads to a fast and general way to improve performance when training on relatively little data.

Empirical evaluations have found that when transferring deep features across tasks, freezing weights of some layers during fine-tuning generally harms performance~\citep{yosinski2014transferable}.  
These results suggest that, given a small dataset, it is better to adjust all of the layers a little rather than to adjust just a few layers a large amount, and so perhaps the ideal setting will adjust all of the layers the same amount.
While these studies did indeed set the learning rate to be the same for all layers, somewhat counterintuitively this does not actually enforce that all layers learn at the same rate.
To see this, say we have a network where there are two convolution layers separated by a ReLU.  
Multiplying the weights and bias term of the first layer by a scalar $\alpha>0$, and then dividing the weights (but not bias) of the next (higher) layer by the same constant $\alpha$ will result in a network which computes exactly the same function.
However, note that the gradients of the two layers are not the same: they will be divided by $\alpha$ for the first layer, and multiplied by $\alpha$ for the second.
Worse, an update of a given magnitude will have a smaller effect on the lower layer than the higher layer, simply because the lower layer's norm is now larger.
Using this kind of reparameterization, it is easy to make the gradients for certain layers vanish during fine-tuning, or even to make them explode, resulting in a network that is impossible to fine-tune despite representing exactly the same function.
Conversely, this sort of re-parameterization gives us a tool we can use to calibrate layer-by-layer learning to improve fine-tuning performance, provided we have an appropriate principle for making such adjustments.

Where can we look to find such a principle?
A number of works have already suggested that statistical properties of network activations can impact network performance.
Many focus on initializations which control the variance of network activations.
\cite{krizhevsky2012imagenet} carefully designed their architecture to ensure gradients neither vanish nor explode.
However, this is no longer possible for deeper architectures such as VGG~\citep{vgg} or GoogLeNet~\citep{googlenet}.
\cite{glorot2010understanding, saxe2013exact, sussillo2014random, he2015delving, bradley2010learning} show that properly scaled random initialization can deal with the vanishing gradient problem, if the architectures are limited to linear transformations, followed by a very specific non-linearities.
\cite{saxe2013exact} focus on linear networks, \cite{glorot2010understanding} derive an initialization for networks with tanh non-linearities, while \cite{he2015delving} focus on the more commonly used ReLUs.
However, none of the above papers consider more general network including pooling, dropout, LRN layers~\citep{krizhevsky2012imagenet}, or DAG-structured networks~\citep{googlenet}.
We argue that initializing the network with real training data improves these approximations and achieves a better performance.
Early approaches to data-driven initializations showed that whitening the activations at all layers can mitigate the vanishing gradient problem~\citep{lecun-98b}, but it does not ensure all layers train at an equal rate.
More recently, batch normalization~\citep{ioffe2015batch} enforces that the output of each convolution and fully-connected layer are zero mean with unit variance for every batch.
In practice, however, this means that the network's behavior on a single example depends on the other members of the batch, and removing this dependency at test-time relies on approximating batch statistics.  
The fact that these methods show improved convergence speed at training time suggests we are justified in investigating the statistics of activations.  
However, the main goal of our work differs in two important respects.
First, these previous works pay relatively little attention to the behavior on smaller training sets, instead focusing on training speed.  
Second, while all above initializations require a random initialization, our approach aims to handle structured initialization, and even improve pre-trained networks.

\section{Preliminaries}
We are interested in parameterizing (and re-parameterizing) CNNs, where the output is a highly non-convex function of both the inputs and the parameters.
Hence, we begin with some notation which will let us describe how a CNN's behavior will change as we alter the parameters. 
We focus on feed-forward networks of the form
$$
  z_k = f_k(z_{k-1}; \theta_k),
$$
where $z_k$ is a vector of hidden activations of the network, and $f_k$ is a transformation with parameters $\theta_k$. 
$f_k$ may be a linear transformation $f_k(z^{\prime}_{k}; \theta_k) = W_k z_{k-1} + b_k$, or it may be a non-linearity $f_{k+1}(z_k; \theta_k) = \sigma_{k+1}(z^{\prime}_k)$ such as a rectified linear unit (ReLU) $\sigma(x) = \max(x,0)$.  
Other common non-linearities include local response normalization or pooling \citep{krizhevsky2012imagenet,googlenet,vgg}.
However, as is common in neural networks, we assume these nonlinearities are not parametrized and kept fixed during training.
Hence, $\theta_k$ contains only $(W_k,b_k)$ for each affine layer $k$.

To deal with spatially-structured inputs like images, most hidden activations $z_k \in \mathbb{R}^{C_k \times A_k \times B_k}$ are arranged in a two dimensional grid of size $A_k \times B_k$ (for image width $A_k$ and height $B_k$) with $C_k$ channels per grid cell.  
We let $z_0$ denote the input image.
The final output, however, is generally not spatial, and so later layers are reduced to the form $z_N = \mathbb{R}^{C_N \times 1 \times 1}$, where $C_N$ is the number of output units.
The last of these outputs is converted into a loss with respect to some label; for classification, the approach is to convert the final output into a probability distribution over labels via a Softmax function.  
Learning aims to minimize the expected loss over the training dataset.
Despite the non-convexity of this learning problem, backpropagation and Stochastic Gradient Descent often finds good local minima if initialized properly~\citep{lecun-98b}.
Given an arbitrary neural network, we next aim for a good parameterization.
A good parameterization should be able to learn all weights of a network equally well.
We measure how well a certain weight in the network learns by how much the gradient of a loss function would change it.
A large change means it learns more quickly, while a small change implies it learns more slowly.
We initialize our network such that all weights in all layers learn equally fast.

\section{Data-dependent initialization}
\lblsec{data_init}
Given an $N$-layer neural network with loss function $\ell(z_N)$, we first define $C_{i,j,k}^{2}$ to be the expected norm of the gradient with respect to weights $W_k(i,j)$ in layer $k$:
\begin{equation}
C_{k,i,j}^2 = \expect{z_0 \sim D}{\left(\frac{\partial}{\partial W_k(i,j)} \ell(z_N)\right)^2} =
\expectbig{z_0 \sim D}{\bigg(z_{k-1}(j) \underbrace{\frac{\partial}{\partial z_{k}(i)} \ell(z_N)}_{y_k(i)} \bigg)^2}, \lbleq{change_rate}
\end{equation}
where $D$ is a set of input images and $y_k$ is the backpropagated error.
Similar reasoning can be applied to the biases $b_k$, but where the activations are replaced by the constant $1$.
To not rely on any labels during initialization, we use a random linear loss function $\ell(z_N) = \eta^\top z_N$, where $\eta \sim \mathcal{N}(0,I)$ is sampled from a unit Gaussian distribution.
In other words, we initialize the top gradient to a random Gaussian noise vector $\eta$ during backpropagation.
We sample a different random loss $\eta$ for each image.

In order for all parameters to learn at  the same ``rate,'' we require the change in eq. \ref{eq:change_rate} to be proportional to the magnitude of the weights $\|W_k\|_2^2$ of the current layer; i.e.,
\begin{equation}
 \tilde C_{k,i,j}^2 = \frac{C_{k,i,j}^2}{\|W_k\|_2^2} \lbleq{rel_change}
\end{equation}
is constant for all weights.
However this is hard to enforce, because for non-linear networks the backpropagated error $y_k$ is a function of the activations $z_{k-1}$.
A change in weights that affects the activations $z_{k-1}$ will indirectly change $y_k$.
This effect is often non-linear and hard to control or predict.

We thus simplify \refeq{rel_change}:
rather than enforce that the individual weights all learn at the same rate, we enforce that the \textit{columns} of weight matrix $W_k$ do so, i.e.:
\begin{equation}
 \tilde C_{k,j}^2 = \frac{1}{N} \sum_i \tilde C_{k,i,j}^2 = \frac{1}{N\|W_k\|_2^2} \expect{z_0 \sim D}{z_{k-1}(j)^2 \|y_k\|_2^2}, \lbleq{change_rate_col}
\end{equation}
should be approximately constant, where $N$ is the number of rows of the weight matrix.
As we will show in \refsec{scaling}, all weights tend to train at roughly the same rate even though the objective does not enforce this.
Looking at~\refeq{change_rate_col}, the relative change of a column of the weight matrix is a function of 1) the magnitude of a single activation of the bottom layer, and 2) the norm of the backpropagated gradient.
The value of a single input to a layer will generally have a relatively small impact on the norm of the gradient to the entire layer.  
Hence, we assume $z_{k-1}(j)$ and $\|y_k\|$ are independent, leading to the following simplification of the objective:
\begin{equation}
\tilde C_{k,j}^2 \approx \expect{z_0 \sim D}{z_{k-1}(j)^2} \frac{\expect{z_0 \sim D}{\|y_k\|_2^2}}{N \|W_k\|_2^2}. \lbleq{change_rate_approx}
\end{equation}
This approximation conveniently decouples the change rate per column, which depends on $z_{k-1}(j)^2$, from the global change rate per layer, which depends on the gradient magnitude $\|y_k\|_2^2$, allowing us to correct them in two separate steps.

In \refsec{weight_norm}, we show how to % 
satisfy $\expect{z_0 \sim D}{z_{k-1}(i)^2} = c_k$ for a layer-wise constant $c_k$.
In \refsec{weight_scale}, we then adjust this layer-wise constant $c_k$ to ensure that all gradients are properly calibrated between layers, in a way that can be applied to pre-initialized networks. % 
Finally, in \refsec{kmeans} we present multiple data-driven weight initializations.

\subsection{Within-Layer weight normalization}
\lblsec{weight_norm}
We aim to ensure that each channel that a layer $k+1$ receives a similarly distributed input.
It is straightforward to initialize weights in affine layers such that the units have outputs following similar distributions.  E.g., we could enforce that % 
layer $k$ activations $z_{k}(i,a,b)$ have $
\expect{z_0 \sim D, a, b}{z_{k}(i,a,b)} = \beta
$ and $
\expect{z_0 \sim D, a, b}{(z_{k}(i,a,b)-\beta)^2} = 1
$ simply via properly-scaled random projections, where $a$ and $b$ index over the 2D spatial extent of the feature map.
However, we next have to contend with the nonlinearity $\sigma(.)$.  
Thankfully, most nonlinearities (such as sigmoid or ReLU) operate independently on different channels.
Hence, the different channels will undergo the same transformation, and the output channels will follow the same distribution if the input channels do (though the outputs will generally not be the same distribution as the inputs).
In fact, most common CNN layers that apply a homogeneous operation to uniformly-sized windows of the input with regular stride, such as local response normalization, and pooling, empirically preserve this identical distribution requirement as well, making it broadly applicable.

We normalize the network activations using empirical estimates of activation statistics
obtained from actual data samples $z_0 \sim D$.
In particular, for each affine layer $k \in \{1, 2, \dots, N\}$ in a topological ordering of the network graph,
we compute the empirical mean and standard deviations for all outgoing activations and normalize the weights $W_k$ such that all activations have unit variance and mean $\beta$.
This procedure is summarized in \refalg{layer_init}. 

\begin{algorithm}[t]
 \begin{algorithmic}
  \For{each affine layer $k$}
   \State Initialize weights from a zero-mean Gaussian $W_k \sim \mathcal{N}(0, I)$ and biases $b_k = 0$
   \State Draw samples $z_0 \in \tilde{D} \subset D$ and pass them through the first $k$ layers of the network
   \State Compute the per-channel sample mean $\hat{\mu}_k(i)$ and variance $\hat{\sigma}_k(i)^2$ of $z_k(i)$
   \State Rescale the weights by $W_k(i, :) \gets W_k(i, :) / \hat{\sigma}_k(i)$
   \State Set the bias $b_k(i) \gets \beta - \hat{\mu}_k(i) / \hat{\sigma}_k(i)$ to center activations around $\beta$
  \EndFor
 \end{algorithmic}
 \caption{Within-layer initialization.}
 \lblalg{layer_init}
\end{algorithm}
The variance of our estimate of the sample statistics falls with the size of the sample $|\tilde{D}|$.
In practice, for CNN initialization, we find that on the order of just dozens of samples is typically sufficient.

Note that this simple empirical initialization strategy
guarantees affine layer activations with a particular center and scale
while making no assumptions (beyond non-zero variance) about the inputs to the layer,
making it robust to any exotic choice of non-linearity or other intermediate operation.
This is in contrast with existing approaches
designed for particular non-linearities and with architectural constraints.
Extending these methods to handle operations for which they weren't designed
while maintaining the desired scaling properties may be possible,
but it would at least require careful thought,
while our simple empirical initialization strategy generalizes to any operations
and DAG architecture with no additional implementation effort.

On the other hand, note that for architectures which are not purely feed-forward, the assumption of identically distributed affine layer inputs may not hold.
GoogLeNet~\citep{googlenet}, for example, concatenates layers which are computed via different operations on the same input, and hence may not be identically distributed, before feeding the result into a convolution.
Our method cannot guarantee identically distributed inputs for arbitrary DAG-structured networks, so it should be applied to non-feed-forward networks with care.

\subsection{Between-layer scale adjustment}
\lblsec{global_scale}
Because the initialization given in \refsec{weight_norm}
results in activations $z_k(i)$ with unit variance,
the expected change rate $C_{k,i}^2$ of a column $i$ of the weight matrix $W_k$ is constant
across all columns $i$, under the approximation given in \refeq{change_rate_approx}.
However, this does not provide any guarantee of the scaling of the change rates \textit{between} layers.

We use an iterative procedure to obtain roughly constant parameter change rates $C_{k,i}^2$ across all layers $k$
(as well as all columns $i$ within a layer), given previously-initialized weights. % 
At each iteration we estimate the average change ratio ($\tilde C_{k,i,j}$) per layer.  
We also estimate a global change ratio, as the geometric mean of all layer-wise change ratios.
The geometric mean ensures that the output remains unchanged in completely homogeneous networks.
We then scale the parameters for each layer to be closer to this global change ratio.
We simultaneously undo this scaling in the layer above, such that the function that the entire network computes is unchanged.
This scaling can be undone by inserting an auxiliary scaling layer after each affine layer.
However for homogeneous non-linearities, such as ReLU, Pooling or LRN, this scaling can be undone at in the next affine layer without the need of a special scaling layer.
The between-layer scale adjustment procedure is summarized in \refalg{global_init}.
Adjusting the scale of all layers simultaneously can lead to an oscillatory behavior.
To prevent this we add a small damping factor $\alpha$ (usually $\alpha=0.25$).

\begin{algorithm}[t]
 \begin{algorithmic}
  \State Draw samples $z_0 \in \tilde{D} \subset D$
  \Repeat
   \State Compute the ratio $\tilde C_k = \expect{j}{\tilde C_{k,j}}$
   \State Compute the average ratio $\tilde C = \left( \prod_k C_k \right)^{1/N}$
   \State Compute a scale correction $r_k = \left({\tilde C_k} / {\tilde C}\right)^{\alpha/2}$ with a damping factor $\alpha < 1$
   \State Correct the weights and biases of layer $k$: $b_k \gets r_k b_k$, $W_k \gets r_k W_k$
   \State Undo the scaling $r_k$ in the layer above
  \Until{Convergence (roughly $10$ iterations)}
 \end{algorithmic}
 \caption{Between-layer normalization.}
 \lblalg{global_init}
\end{algorithm}

With a relatively small number of steps (we use $10$),
this procedure results in roughly constant initial change rates of the parameters in all layers of the network, regardless of its depth.
\lblsec{weight_scale}

\subsection{Weight initializations}
\lblsec{kmeans}
Until now, we used a random Gaussian initialization of the weights, but our procedure does not require this.
Hence, we explored two data-driven initializations: a PCA-based initialization and a \kmeans based initialization.
For the PCA-based initialization, we set the weights such that the layer outputs are white and decorrelated.
For each layer $k$ we record the features activations $z_{k-1}$ of each channel $c$ across all spatial locations for all images in $D$.
Then then use the first $M$ principal components of those activations as our weight matrix $W_k$.
For the \kmeans based initialization, we follow \cite{coates2012learning} and apply spherical \kmeans on whitened feature activations.
We use the cluster centers of \kmeans as initial weights for our layers, such that each output unit corresponds to one centroid of \kmeans.
\kmeans usually does a better job than PCA, as it captures the modes of the input data, instead of merely decorrelating it.
We use both \kmeans and PCA on just the convolutional layers of the architecture, as we don't have enough data to estimate the required number of weights for fully connected layers.

In summary, we initialize weights or all filters (\shortrefsec{kmeans}), then normalize those weights such that all activations are equally distributed (\shortrefsec{weight_norm}), and finally rescale each layer such that the gradient ratio is constant across layers (\shortrefsec{global_scale}).
This initialization encures that all weights learn at approximately the same rate, leading to a better convergence and more accurate models, as we will show next.

\section{Evaluation}
\lblsec{eval}
We implement our initialization and all experiments in the open-source deep learning framework Caffe~\citep{caffe}.
To assess how easily a network can be fine-tuned with limited data, we use the classification and detection challenges in PASCAL VOC 2007~\citep{everingham2014pascal}, which contains $5011$ images for training and $4952$ for testing.

\paragraph{Architectures}
Most of our experiments are performed on the $8$ layer \textit{CaffeNet} architecture a small modification of \textit{AlexNet} \citep{krizhevsky2012imagenet}.
We use the default architecture for all comparisons, except for \cite{doersch2015unsupervised} which removed groups in the convolutional layers.
We also show results on the much deeper GoogLeNet~\citep{googlenet} and VGG~\citep{vgg} architectures.

\paragraph{Image classification}
The VOC image classification task is to predict the presence or absence of each of $20$ object classes in an image.
For this task we fine-tune all networks using a sigmoid cross-entropy loss on random crops of each image.
We optimize each network via Stochastic Gradient Descent (SGD) for 80,000 iterations with an initial learning rate of $0.001$ (dropped by $0.5$ every 10,000 iterations), batch size of $10$, and momentum of $0.9$.
The total training takes one hour on a Titan X GPU for \textit{CaffeNet}.
We tried different settings for various methods, but found these setting to work best for all initializations.
At test time we average $10$ random crops of the image to determine the presence or absence of an object.
The CNN estimates the likelihood that each object is present, which we use as a score to compute a precision-recall curve per class.
We evaluate all algorithms using mean average precision (mAP) ~\citep{everingham2014pascal}.

\paragraph{Object detection}
In addition to predicting the presence of absence of an object in a scene, object detection requires the precise localization of each object using a bounding box.
We again evaluate mean average precision \citep{everingham2014pascal}.
We fine-tune all our models using Fast R-CNN~\citep{fastrcnn}.
For a fair comparison we varied the parameters of the fine-tuning for each of the different initializations.
We tried three different learning rates ($0.01$, $0.002$ and $0.001$) dropped by $0.1$ every 50,000 iterations, with a total of 150,000 training iterations.
We used multi-scale training and fine-tuned all layers.
We evaluate all models on single scale.
All other settings were kept at their default values.
Training and evaluation took roughly $8$ hours in a Titan X GPU for \textit{CaffeNet}.
All models are trained from scratch unless otherwise stated.

For both experiments we use $160$ images of the VOC2007 training set for our initialization.
$160$ images are sufficient to robustly estimate activation statistics, as each unit usually sees tens of thousands of activations throughout all spacial locations in an images.
At the same time, this relatively small set of images keeps the computational cost low.

\subsection{Scaling and learning algorithms}
\lblsec{scaling}
\begin{figure}[t]
\vspace{-1em}
 \begin{subfigure}[b]{.47\columnwidth}
   \centering
   \input{figures/scaling/mean.tex}
  \caption{average change rate}
  \lblfig{change_rate_avg}
 \end{subfigure}
 \hspace{0.03\linewidth}
 \begin{subfigure}[b]{.47\columnwidth}
   \centering
   \input{figures/scaling/variation.tex}
  \caption{coefficient of variation}
  \lblfig{change_rate_var}
 \end{subfigure}
 \caption{Visualization of the relative change rate $\tilde C_{k,i,j}$ in \textit{CaffeNet} for various initializations estimated on $100$ images.
 (a) shows the average change rate per layer, a flat curve is better, as all layers learn at the same rate.
 (b) shows the coefficient of variation for the change rate within each layer, lower is better as weights within a layer train more uniformly.}
\vspace{-0.5em}
 \lblfig{change_rate}
\vspace{-0.5em}
\end{figure}
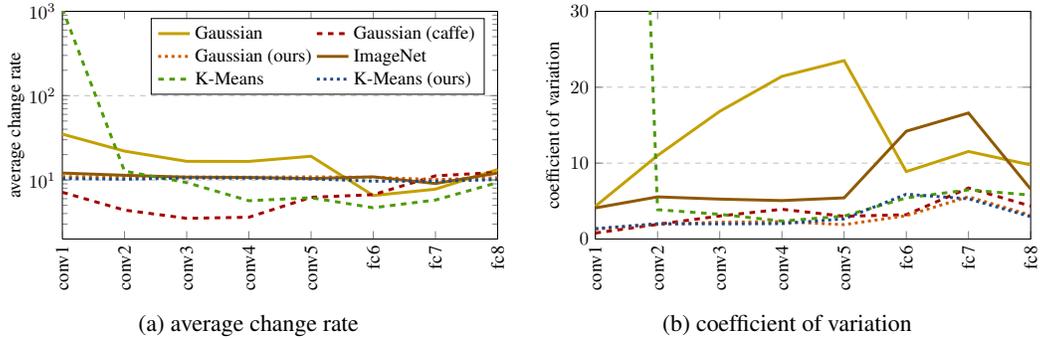
\begin{table}[b]
\vspace{-1em}
\centering
\begin{tabular}{lcccccccc}
 & \multicolumn{2}{c}{SGD} & \multicolumn{2}{c}{SGD + BN} & \multicolumn{2}{c}{ADAM} & \multicolumn{2}{c}{ADAM + BN}\\
Scaling & Gaus. & k-mns. & Gaus. & k-mns. & Gaus. & k-mns. & Gaus. & k-mns.\\
\midrule
no scaling & 50.8\% & 41.2\% & 51.6\% & 49.4\% & 50.9\% & 52.0\% & 55.7\% & 53.8\% \\
\midrule
Within-layer (Ours) & 47.6\% & 41.2\% & - & - & - & - & 53.2\% & 53.1\%\\
Between-layer (Ours) & 52.7\% & 55.7\% & - & - & - & - & 54.5\% & 57.2\%\\
Both (Ours) & 53.3\% & \bf 56.6\% & 56.6\% & \bf 60.0\% & 53.1\% & \bf 56.9\% & 56.9\% & \bf 59.8\%\\
\end{tabular}
\vspace{-0.5em}
\caption{Classification performance of various initializations, training algorithms and with and without batch normalization (BN) on PASCAL VOC2007 for both random Gaussian (Gaus.) and \kmeans (k-mns.) initialized weights.}
\lbltbl{init_method}
\vspace{-1em}
\end{table}

We begin our evaluation by measuring and comparing the relative change rate $\tilde C_{k,i,j}$ of all weights in the network (see \refeq{rel_change}) for different initializations.
We estimate $\tilde C_{k,i,j}$ using $100$ images of the VOC 2007 validation set.
We compare our models to an ImageNet pretrained model, initialized with random Gaussian weights (with standard deviation $\sigma=0.01$), an unscaled \kmeans initialization, as well as the Gaussian initialization in Caffe~\citep{caffe}, for which biases and standard deviations were handpicked per layer.
\reffig{change_rate_avg} visualizes the average change rate per layer. % 
Our initialization, as well as the ImageNet pretrained model, have similar change rates for all layers (i.e., all layers learn at the same rate), while random initializations and \kmeans have a drastically different change rates. % 
\reffig{change_rate_var} measures the coefficient of variation of the change rate for each layer, defined as the standard deviation of the change rate, divided by their mean value.
Our coefficient of variation is low throughout all layers, despite scaling the rate of change of columns of the weight matrix, instead of individual elements.
Note that the low values are mirrored in the hand-tuned Caffe initialization.

Next we explore how those different initializations perform on the VOC 2007 classification task, as shown in \reftbl{init_method}.
We train both a random Gaussian and \kmeans initialization using different initial scalings.
Without scaling the random Gaussian initialization fares quite well, however the \kmeans initialization does poorly, due to the worse initial change rate as shown in \reffig{change_rate}.
Correcting for the within-layer scaling alone does not improve the performance much, as it worsens the between-layer scaling for both initializations.
However in combination with the between-layer adjustment both initializations perform very well.

Both the between-layer and within-layer scaling could potentially be addressed by a stronger second order optimization method, such as ADAM~\citep{kingma2014adam} or batch normalization~\citep{ioffe2015batch}.
In general, ADAM is able to slightly improve on SGD for an unscaled initialization, especially when combined with batch normalization.
Neither batch-norm nor ADAM alone or combined does perform as well as simple SGD with our \kmeans initialization.
More interestingly, our initialization complements those stronger optimization methods and we see an improvement by combining them with our initialization.

\subsection{Weight initialization}
\begin{table}[t]
\vspace{-1em}
\centering
\begin{tabular}{lcc}
Method & Classification & Detection\\
\midrule
Xavier \cite{glorot2010understanding} & 51.1\% & 40.4\%\\
MSRA \cite{he2015delving} & 43.3\% & 37.2\%\\
Random Gaussian (hand tuned) & 53.4\% & 41.3\%\\
\midrule
Ours (Random Gaussian) & 53.3\% & 43.4\%\\
Ours (PCA) & 52.8\% & 43.1\%\\
Ours (\kmeans) & \bf 56.6\% & \bf 45.6\%\\
\end{tabular}
\vspace{-0.5em}
\caption{Comparison of different initialization methods on PASCAL VOC2007 classification and detection.}
\lbltbl{init_cmp}
\vspace{-1em}
\end{table}

Next we compare our Gaussian, PCA and \kmeans based weights, with initializations proposed by \cite{glorot2010understanding} (commonly known as ``xavier''), \cite{he2015delving}, and a carefully chosen Gaussian initialization of \cite{caffe}.
We followed the suggestions of He \etal and used their initialization only for the convolutional layers, while choosing a random Gaussian initialization for the fully connected layers.
We compare all methods on both classification and detection performance in \reftbl{init_cmp}.

The first thing to notice is that both Glorot \& Bengio and He \etal perform worse than a carefully chosen random Gaussian initialization.
One possibility for the drop in performance comes from the additional layers, such as Pooling or LRN used in \textit{CaffeNet}.
Neither Glorot \& Bengio nor He \etal consider those layers but rather focus on linear layers followed by tanh or ReLU non-linearities.

Our initialization on the other hand has no trouble with those additional layers and substantially improves on the random Gaussian initialization.

\subsection{Comparison to unsupervised pre-training}
\begin{table}[b]
\centering
\begin{tabular}{lcccc}
Method & Supervision & Pretraining time & Classification & Detection\\
\midrule
\cite{agrawal2015learning} & egomotion & $10$ hours & 52.9\% & 41.8\%\\
\cite{wang2015unsupervised}\footnotemark & motion & $1$ week & \bf 62.8\% & \bf 47.4\%\\
\cite{doersch2015unsupervised} & unsupervised & $4$ weeks & 55.3\% & 46.6\%\\
\midrule
\cite{krizhevsky2012imagenet} & $1000$ class labels & $3$ days & \bf 78.2\% & \bf 56.8\%\\
\midrule
Ours (\kmeans) & initialization & $54$ seconds & 56.6\% & 45.6\%\\
\midrule
Ours + \cite{agrawal2015learning} & egomotion & $10$ hours & 54.2\% & 43.9\%\\
Ours + \cite{wang2015unsupervised} & motion & $1$ week & 63.1\% & 47.2\%\\
Ours + \cite{doersch2015unsupervised} & unsupervised & $4$ weeks & \bf 65.3\% & \bf 51.1\%\\
\end{tabular}
\vspace{-0.5em}
\caption{Comparison of classification and detection results on the PASCAL VOC2007 test set.}
\lbltbl{classificationRes}
\vspace{-0.5em}
\end{table}

We now compare our simple, properly scaled initializations to the state-of-the-art unsupervised pre-training methods on VOC 2007 classification and detection.
\reftbl{classificationRes} shows a summary of the results, including the amount of pre-training time, as well as the type of supervision used.
\cite{agrawal2015learning} uses egomotion, as measured by a moving car in a city to pre-train a model. 
While this information is not always readily available, it can be read from sensors and is thus ``free.''
We believe egomotion information does not often correlate with the kind of semantic information that is required for classification or detection, and hence the egomotion pretrained model performs worse than our random baseline.
\cite{wang2015unsupervised} supervise their pre-training using relative motion of objects in pre-selected youtube videos, as obtained by a tracker.
Their model is generally quite well scaled and trains well for both classification and detection.
\cite{doersch2015unsupervised} predict the relative arrangement of image patches to pre-train a model.
Their model is trained the longest with $4$ weeks of training.
It does well on detection, but lags behind other methods in classification.

Interestingly our \kmeans initialization is able to keep up with most unsupervised pre-training methods, despite containing very little semantic information.
To analyze what information is actually captured, we sampled 100 random ImageNet images and found nearest neighbors for them from a pool of 50,000 other random ImageNet images, using the high-level feature spaces from different methods.
\reffig{nns} shows the results.  Overall, different unsupervised methods seem to focus on different attributes for matching.  
For example, ours appears to have some texture and material information, whereas the method of~\cite{doersch2015unsupervised} seems to preserve more specific shape information.  

As a final experiment we reinitialize all unsupervised pre-training methods to be properly scaled and compare with our initializations which use no auxiliary training beyond the proposed initializations.
In particular, we take their pretrained network weights and apply the between-layer adjustment described in \refsec{global_scale}.
(We do not perform local scaling as we find that the activations in these models are already scaled reasonably well locally.)
The bottom three rows of \reftbl{classificationRes} give our results for our rescaled versions of these models on the VOC classification and detection tasks.
We find that for two of the three models~\citep{agrawal2015learning,doersch2015unsupervised} this rescaling improves results significantly;
our rescaling of~\cite{wang2015unsupervised} on the other hand does not improve its performance, indicating it was likely relatively well-scaled globally to begin with.
The best-performing method with auxiliary self-supervision using our rescaled features is that of~\cite{doersch2015unsupervised} -- in this case our rescaling improves its results on the classification task by a relative margin of 18\%.
This suggests that our method nicely complements existing unsupervised and self-supervised methods and could facilitate easier future exploration of this rich space of methods.
\footnotetext{an earlier version of this paper reported 58.4\% and 44.0\% for the color model of Wand \& Gupta, this version uses the grayscale model which performs better.}

\subsection{Different architectures}

Finally we compare our initialization across different architectures, again using PASCAL 2007 classification and detection.
We train both the deep architecture of \cite{googlenet} and \cite{vgg} using our \kmeans and Gaussian initializations.
Unlike prior work we are able to train those models without any intermediate losses or stage-wise supervised pre-training.
We simply add a sigmoid cross-entropy loss to the top of both networks.
Unfortunately neither network outperformed \textit{CaffeNet} in the classification tasks.
GoogLeNet achieves a $50.0\%$ and $55.0\%$ mAP for the two initializations respectively, while 16-layer VGG performs as $53.8\%$ and $56.5\%$.
This might have to do with the limited amount of supervised training data available to the model at during training.
The training time was $4$ and $12$ times slower than \textit{CaffeNet}, which made them prohibitively slow for detection.

\subsection{ImageNet training}

Finally, we test our data-dependent initializations on two well-known CNN architectures which have been successfully applied to the ImageNet LSVRC 1000-way classification task: \textit{CaffeNet}~\citep{caffe} and \textit{GoogLeNet}~\citep{googlenet}.
We initialize the 1000-way classification layers to 0 in these experiments (except in our reproductions of the reference models), as we find this improves the initial learning velocity.

\paragraph{CaffeNet}
We train instances of CaffeNet using our initializations, with the architecture and all other hyperparameters set to those used to train the reference model: learning rate 0.01 (dropped by a factor of 0.1 every $10^5$ iterations), momentum 0.9, and batch size 256.
We also train a variant of the architecture with no local response normalization (LRN) layers.

Our CaffeNet training results are presented in \reffig{caffenetcurves}.
Over the first 100,000 iterations (\reffig{caffenetcurves}, middle row), and particularly over the first 10,000 (\reffig{caffenetcurves}, top row),
our initializations reduce the network's classification error on both the training and validation sets
at a much faster rate than the reference initialization.

With the full 320,000 training iterations, all initializations achieve similar accuracy on the training and validation sets; however,
in these experiments the carefully chosen reference initialization pulled non-trivially ahead of our initializations' error
after the second learning rate drop to a rate of $10^{-4}$.
We do not yet know why this occurs, or whether the difference is significant.

Over the first 100,000 iterations, among models initialized using our method, the \kmeans initialization reduces the loss slightly faster than the random initialization.
Interestingly, the model variant without LRN layers seems to learn just as quickly as the directly comparable network with LRNs, suggesting such normalizations may not be necessary given a well-chosen initialization.

\paragraph{GoogLeNet}
We apply our best-performing initialization from the CaffeNet experiments---\kmeans---to a deeper network, GoogLeNet~\citep{googlenet}.
We use the SGD hyperparameters from the Caffe~\citep{caffe} GoogleNet implementation (specifically, the ``quick'' version which is trained for 2.4 million iterations), and also retrain our own instance of the model with the initialization used in the reference model (based on~\cite{glorot2010understanding}).

Due to the depth of the architecture (22 layers, compared to CaffeNet's 8) and the difficulty of propagating gradient signal to the early layers of the network, GoogLeNet includes additional ``auxiliary classifiers'' branching off from intermediate layers of the network to amplify the gradient signal to learn these early layers.
To verify that networks initialized using our proposed method should have no problem backpropagating appropriately scaled gradients through all layers of arbitrarily deep networks, we also train a variant of GoogLeNet which omits the two intermediate loss towers,
otherwise keeping the rest of the architecture fixed.

Our GoogLeNet training results are presented in \reffig{googlenetcurves}.
We plot only the loss of the final classifier for comparability with the single-classifier model.
The models initialized with our method learn much faster than the model using the reference initialization stategy.
Furthermore, the model trained using only a single classifier learns at roughly the same rate as the original three loss tower architecture, and each iteration of training in the single classifier model is slightly faster due to the removal of layers to compute the additional losses.
This result suggests that our initialization could significantly ease exploration of new, deeper CNN architectures,
bypassing the need for architectural tweaks like the intermediate losses used to train GoogLeNet.

\section{Discussion}
Our method is a conceptually simple data-dependent initialization strategy for CNNs which enforces empirically identically distributed activations locally (within a layer), and roughly uniform global scaling of weight gradients across all layers of arbitrarily deep networks.
Our experiments (\refsec{eval}) demonstrate that this rescaling of weights results in substantially improved CNN representations for tasks with limited labeled data (as in the PASCAL VOC classification and detection training sets), improves representations learned by existing self-supervised and unsupervised methods, and substantially accelerates the early stages of CNN training on large-scale datasets (e.g., ImageNet).
We hope that our initializations will facilitate further advancement in unsupervised and self-supervised learning as well as more efficient exploration of deeper and larger CNN architectures.

\section*{Acknowledgements}
The thank Alyosha Efros for his input and encouragement, without his ``Gelato bet'' most of this work would not have been explored.
We thank NVIDIA for their generous GPU donations.

\begin{figure}[t]
 \includegraphics[width=.95\columnwidth]{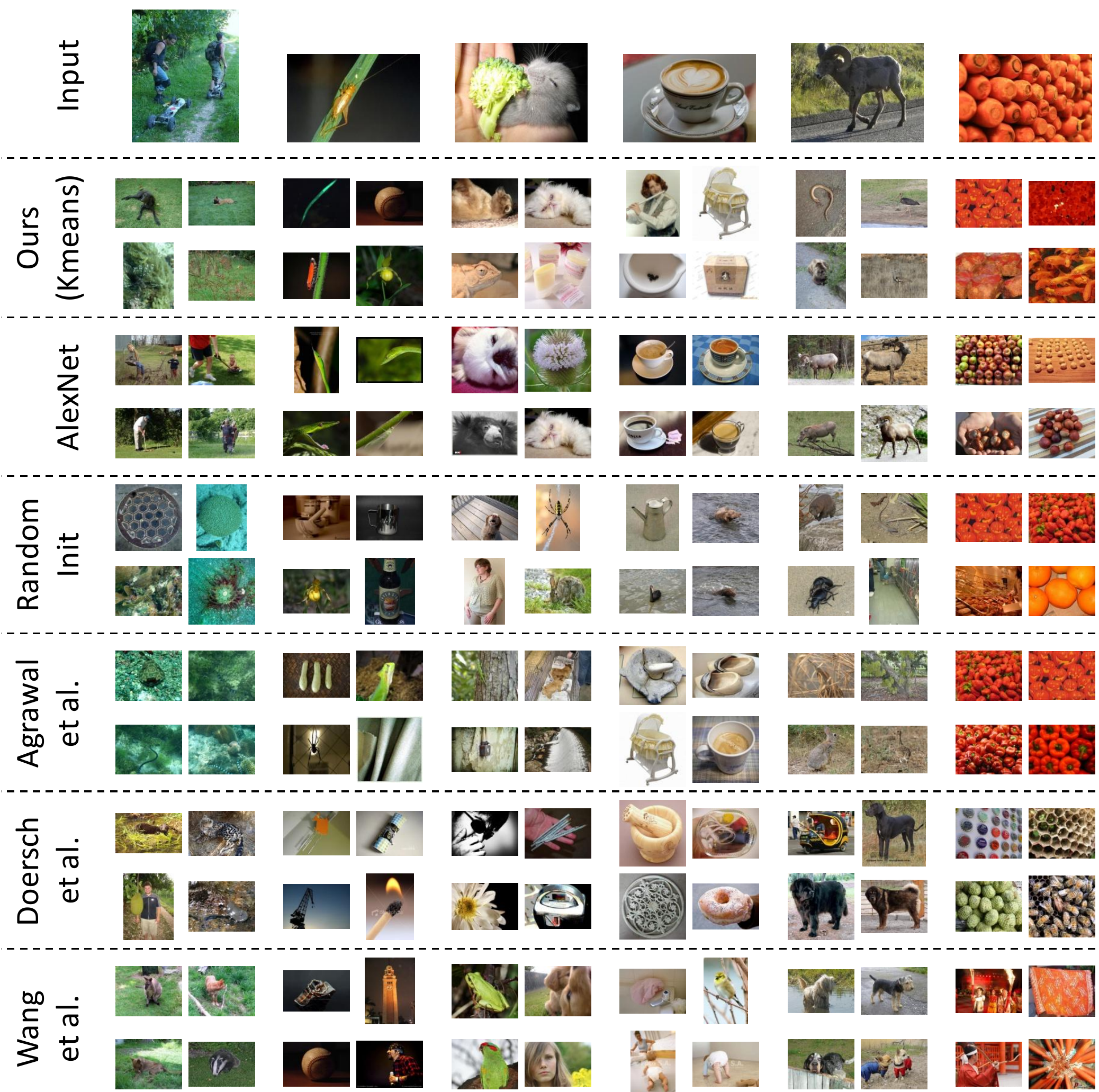}
 \caption{
  Comparison of nearest neighbors for the given input image (top row) in the feature spaces of \textit{CaffeNet}-based CNNs initialized using our method, the fully supervised \textit{CaffeNet}, an untrained \textit{CaffeNet} using Gaussian initialization, and three unsupervised or self-supervised methods from prior work.
  (For \cite{doersch2015unsupervised} we display neighbors in \textit{fc6} feature space; the rest use the \textit{fc7} features.)
  While our initialization is clearly missing the semantics of \textit{CaffeNet}, it does preserve some non-specific texture and shape information, which is often enough for meaningful matches.  
 }
 \lblfig{nns}
\end{figure}

\begin{figure}[t]
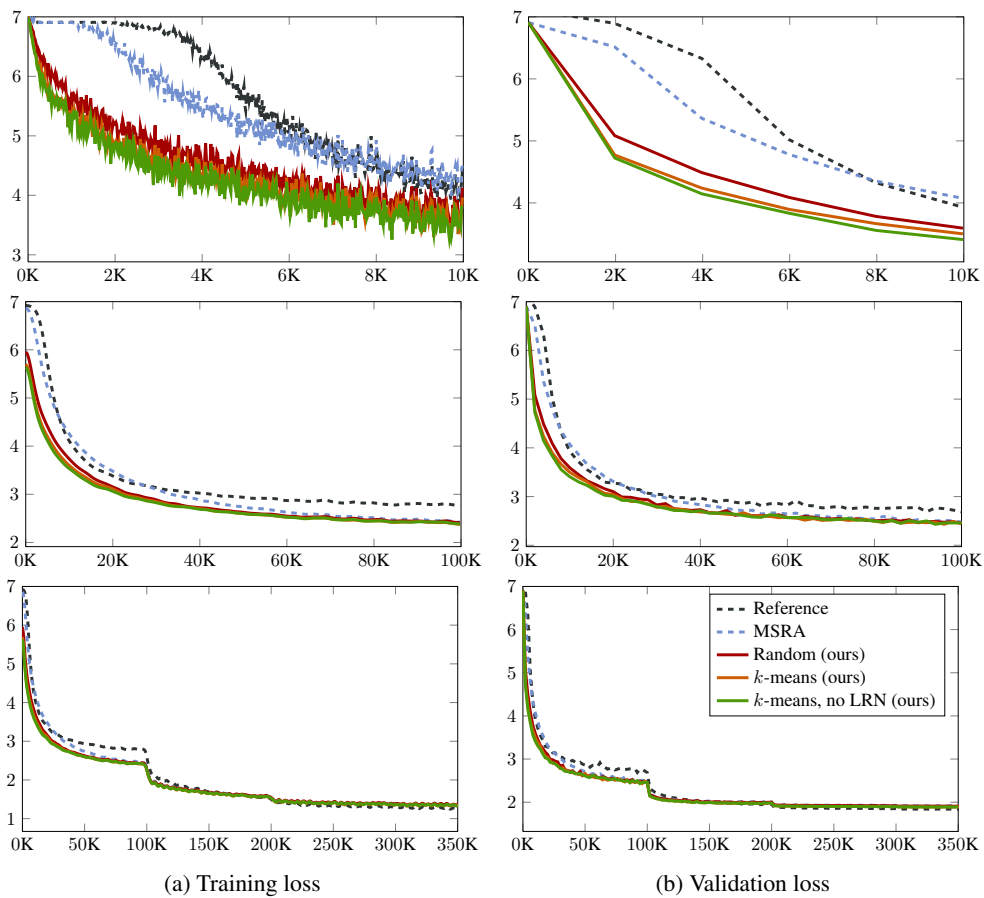

\vspace{-1em}
\centering
 \begin{subfigure}[b]{.47\columnwidth}
   \centering
   \input{figures/imagenet/caffenet_train_loss_10k}%
 \end{subfigure}
 \begin{subfigure}[b]{.47\columnwidth}
   \centering
   \input{figures/imagenet/caffenet_val_loss_10k}%
 \end{subfigure}
 \begin{subfigure}[b]{.47\columnwidth}
   \centering
   \input{figures/imagenet/caffenet_train_loss_100k}%
 \end{subfigure}
 \begin{subfigure}[b]{.47\columnwidth}
   \centering
   \input{figures/imagenet/caffenet_val_loss_100k}%
 \end{subfigure}
 \begin{subfigure}[b]{.47\columnwidth}
   \centering
   \input{figures/imagenet/caffenet_train_loss}%
   \caption{Training loss}
 \end{subfigure}
 \begin{subfigure}[b]{.47\columnwidth}
   \centering
   \input{figures/imagenet/caffenet_val_loss}%
   \caption{Validation loss}
 \end{subfigure}
\vspace{-0.5em}
 \caption{Training and validation loss curves for the CaffeNet architecture trained for the ILSVRC-2012 classification task.
 The training error is unsmoothed in the topmost plot (10K); smoothed over one epoch in the others.
 The validation error is computed over the full validation set every 2000 iterations and is unsmoothed.
 Our initializations (\kmeans, Random) handily outperform both the carefully chosen reference initialization~\citep{caffe} and the MSRA initialization~\citep{he2015delving} over the first 100,000 iterations, but the other initializations catch up after the second learning rate drop at iteration 200,000.
 }
 \lblfig{caffenetcurves}
\vspace{-0.5em}
\end{figure}
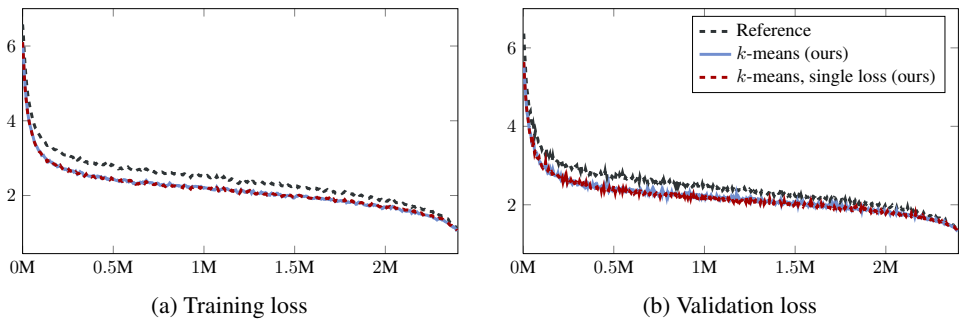
\begin{figure}[t]
\vspace{-0.5em}
 \centering
 \begin{subfigure}[b]{.47\columnwidth}
   \centering
   \input{figures/imagenet/googlenet_train_loss}
   \caption{Training loss}
 \end{subfigure}
 \begin{subfigure}[b]{.47\columnwidth}
   \centering
   \input{figures/imagenet/googlenet_val_loss}
   \caption{Validation loss}
 \end{subfigure}
\vspace{-0.5em}
 \caption{Training and validation loss curves for the GoogLeNet architecture trained for the ILSVRC-2012 classification task.
 The training error plot is again smoothed over roughly the length of an epoch;
 the validation error (computed every 4000 iterations) is unsmoothed.
 Note that our \kmeans initializations outperform the reference initialization, and the single loss model (lacking the auxiliary classifiers) learns at roughly the same rate as the model with auxiliary classifiers.
 The final top-5 validation error are $11.57\%$ for the reference model, $10.85\%$ for our single loss, and $10.69\%$ for our auxiliary loss model.}
 \lblfig{googlenetcurves}
\vspace{-1em}
\end{figure}

\bibliographystyle{iclr2016_conference}
\bibliography{iclr2016_initialization}

\end{document}

%% file: figures/scaling/mean.tex
\begin{tikzpicture}[scale=0.7]
\begin{semilogyaxis}[
    legend cell align=left,
    height=0.9\textwidth,
    width=1.5\textwidth,
    title={},
    xlabel={},
    ylabel={average change rate},
    ylabel absolute, ylabel style={yshift=-1em},
    xmin=0, xmax=7,
    ymin=0, ymax=1000,
    ymajorgrids=true,
    legend pos=north east,
    legend columns=2,
    grid style=dashed,
    xtick={0, 1, 2, 3, 4, 5, 6, 7},
    xticklabels={\rotatebox{90}{conv1}, \rotatebox{90}{conv2}, \rotatebox{90}{conv3}, \rotatebox{90}{conv4}, \rotatebox{90}{conv5}, \rotatebox{90}{fc6}, \rotatebox{90}{fc7}, \rotatebox{90}{fc8}},
]

\addplot[color=ta3butter,ultra thick,solid] coordinates {(0,35.004200)(1,21.994300)(2,16.632500)(3,16.596000)(4,19.116000)(5,6.542880)(6,7.779460)(7,13.315900)}; \addlegendentry{Gaussian}
\addplot[color=ta3scarletred,ultra thick,dashed] coordinates {(0,7.145320)(1,4.395430)(2,3.510380)(3,3.616200)(4,6.232390)(5,6.694680)(6,11.193900)(7,12.313200)}; \addlegendentry{Gaussian (caffe)}
\addplot[color=ta3orange,ultra thick,dotted] coordinates {(0,10.951200)(1,10.876900)(2,10.853700)(3,10.800400)(4,10.888200)(5,10.757800)(6,10.113100)(7,10.634800)}; \addlegendentry{Gaussian (ours)}
\addplot[color=ta3chocolate,ultra thick,solid] coordinates {(0,12.083500)(1,11.369000)(2,10.795900)(3,10.722500)(4,10.400300)(5,10.893500)(6,9.075250)(7,12.080900)}; \addlegendentry{ImageNet}
\addplot[color=ta3chameleon,ultra thick,dashed] coordinates {(0,1076.470000)(1,12.722400)(2,9.312610)(3,5.675140)(4,6.155080)(5,4.686720)(6,5.783210)(7,9.310880)}; \addlegendentry{K-Means}
\addplot[color=ta3skyblue,ultra thick,dotted] coordinates {(0,10.403500)(1,10.208700)(2,10.558800)(3,10.513800)(4,10.302500)(5,9.681300)(6,9.651220)(7,10.106200)}; \addlegendentry{K-Means (ours)}
\end{semilogyaxis}
\end{tikzpicture}

%% file: figures/scaling/variation.tex
\begin{tikzpicture}[scale=0.7]
\begin{axis}[
    legend cell align=left,
    height=0.9\textwidth,
    width=1.5\textwidth,
    title={},
    xlabel={},
    ylabel={coefficient of variation},
    ylabel absolute, ylabel style={yshift=-1.2em},
    xmin=0, xmax=7,
    ymin=0, ymax=30,
    legend pos=north east,
    legend columns=2,
    ymajorgrids=true,
    grid style=dashed,
    xtick={0, 1, 2, 3, 4, 5, 6, 7},
    xticklabels={\rotatebox{90}{conv1}, \rotatebox{90}{conv2}, \rotatebox{90}{conv3}, \rotatebox{90}{conv4}, \rotatebox{90}{conv5}, \rotatebox{90}{fc6}, \rotatebox{90}{fc7}, \rotatebox{90}{fc8}},
]

\addplot[color=ta3butter,ultra thick,solid] coordinates {(0,4.315340)(1,11.004800)(2,16.824000)(3,21.432200)(4,23.507200)(5,8.886080)(6,11.535800)(7,9.763660)};
\addplot[color=ta3scarletred,ultra thick,dashed] coordinates {(0,0.789607)(1,1.929000)(2,3.036550)(3,3.918840)(4,2.995850)(5,3.199120)(6,6.722430)(7,4.341240)};
\addplot[color=ta3orange,ultra thick,dotted] coordinates {(0,1.316830)(1,1.920170)(2,2.225150)(3,2.287510)(4,1.895550)(5,3.074480)(6,5.577490)(7,3.136590)};
\addplot[color=ta3chocolate,ultra thick,solid] coordinates {(0,4.097780)(1,5.550990)(2,5.258870)(3,5.064870)(4,5.421900)(5,14.205400)(6,16.603400)(7,6.576710)};
\addplot[color=ta3chameleon,ultra thick,dashed] coordinates {(0,221.630000)(1,3.866630)(2,3.255190)(3,2.363480)(4,3.021980)(5,5.413960)(6,6.476590)(7,5.788190)};
\addplot[color=ta3skyblue,ultra thick,dotted] coordinates {(0,1.369590)(1,2.017200)(2,1.991220)(3,2.027720)(4,2.651590)(5,5.907260)(6,5.294990)(7,2.946160)};
\end{axis}
\end{tikzpicture}

%% file: figures/imagenet/caffenet_train_loss_10k.tex
\begin{tikzpicture}[scale=0.7]
\begin{axis}[
  legend cell align=left,
  height=0.95\textwidth,
  width=1.5\textwidth,
  xmin=0, xmax=10.0,
  ymax=7,
  legend pos=north east,
  scaled x ticks = false,
  xtick = {0, 2, 4, 6, 8, 10},
  ytick = {3, 4, 5, 6, 7},
  xticklabel = {\pgfmathprintnumber{\tick}K}
]

\addplot[color=ta3gray,ultra thick,dashed] coordinates {(0.000000,7.385940)(0.020000,7.184750)(0.040000,6.982260)(0.060000,6.946820)(0.080000,6.922830)(0.100000,6.920450)(0.120000,6.923320)(0.140000,6.909610)(0.160000,6.909430)(0.180000,6.919610)(0.200000,6.914510)(0.220000,6.904580)(0.240000,6.911220)(0.260000,6.909460)(0.280000,6.910100)(0.300000,6.904690)(0.320000,6.903860)(0.340000,6.908450)(0.360000,6.908120)(0.380000,6.909390)(0.400000,6.908200)(0.420000,6.910150)(0.440000,6.906660)(0.460000,6.905760)(0.480000,6.909290)(0.500000,6.908280)(0.520000,6.908210)(0.540000,6.904740)(0.560000,6.908570)(0.580000,6.909440)(0.600000,6.907410)(0.620000,6.908300)(0.640000,6.907220)(0.660000,6.909730)(0.680000,6.907130)(0.700000,6.906650)(0.720000,6.907500)(0.740000,6.907310)(0.760000,6.913320)(0.780000,6.906180)(0.800000,6.908030)(0.820000,6.908070)(0.840000,6.906960)(0.860000,6.905520)(0.880000,6.907360)(0.900000,6.908010)(0.920000,6.905240)(0.940000,6.904610)(0.960000,6.906590)(0.980000,6.911680)(1.000000,6.907120)(1.020000,6.911090)(1.040000,6.909120)(1.060000,6.908320)(1.080000,6.908070)(1.100000,6.907290)(1.120000,6.908650)(1.140000,6.904050)(1.160000,6.905210)(1.180000,6.903190)(1.200000,6.905870)(1.220000,6.909660)(1.240000,6.908840)(1.260000,6.906730)(1.280000,6.907190)(1.300000,6.909010)(1.320000,6.908150)(1.340000,6.903880)(1.360000,6.906300)(1.380000,6.908160)(1.400000,6.908240)(1.420000,6.907160)(1.440000,6.907290)(1.460000,6.904440)(1.480000,6.908300)(1.500000,6.907730)(1.520000,6.909240)(1.540000,6.909330)(1.560000,6.905270)(1.580000,6.907500)(1.600000,6.910250)(1.620000,6.909300)(1.640000,6.908180)(1.660000,6.906540)(1.680000,6.902700)(1.700000,6.907840)(1.720000,6.912930)(1.740000,6.904620)(1.760000,6.912400)(1.780000,6.908770)(1.800000,6.906240)(1.820000,6.902840)(1.840000,6.909880)(1.860000,6.897360)(1.880000,6.905900)(1.900000,6.898400)(1.920000,6.892770)(1.940000,6.901360)(1.960000,6.900690)(1.980000,6.898240)(2.000000,6.895720)(2.020000,6.902460)(2.040000,6.899130)(2.060000,6.891760)(2.080000,6.865180)(2.100000,6.891530)(2.120000,6.874820)(2.140000,6.883200)(2.160000,6.865020)(2.180000,6.882180)(2.200000,6.882230)(2.220000,6.887070)(2.240000,6.897890)(2.260000,6.833270)(2.280000,6.891570)(2.300000,6.863500)(2.320000,6.866690)(2.340000,6.851530)(2.360000,6.826320)(2.380000,6.831910)(2.400000,6.872760)(2.420000,6.852230)(2.440000,6.857120)(2.460000,6.819960)(2.480000,6.761550)(2.500000,6.830400)(2.520000,6.868780)(2.540000,6.853600)(2.560000,6.825470)(2.580000,6.857820)(2.600000,6.838530)(2.620000,6.850780)(2.640000,6.820870)(2.660000,6.824830)(2.680000,6.829340)(2.700000,6.808520)(2.720000,6.840550)(2.740000,6.855570)(2.760000,6.832860)(2.780000,6.800400)(2.800000,6.813840)(2.820000,6.813210)(2.840000,6.823610)(2.860000,6.809930)(2.880000,6.844190)(2.900000,6.784190)(2.920000,6.790990)(2.940000,6.823720)(2.960000,6.772740)(2.980000,6.758740)(3.000000,6.801070)(3.020000,6.806200)(3.040000,6.838850)(3.060000,6.765520)(3.080000,6.750760)(3.100000,6.736010)(3.120000,6.753780)(3.140000,6.711220)(3.160000,6.724940)(3.180000,6.704730)(3.200000,6.761040)(3.220000,6.714850)(3.240000,6.731540)(3.260000,6.744960)(3.280000,6.740130)(3.300000,6.770100)(3.320000,6.724280)(3.340000,6.776320)(3.360000,6.681080)(3.380000,6.744200)(3.400000,6.697520)(3.420000,6.729390)(3.440000,6.693190)(3.460000,6.693940)(3.480000,6.729120)(3.500000,6.700350)(3.520000,6.655340)(3.540000,6.560890)(3.560000,6.711740)(3.580000,6.634770)(3.600000,6.610640)(3.620000,6.693240)(3.640000,6.554860)(3.660000,6.654680)(3.680000,6.599510)(3.700000,6.618740)(3.720000,6.538090)(3.740000,6.613060)(3.760000,6.557400)(3.780000,6.418470)(3.800000,6.531990)(3.820000,6.451890)(3.840000,6.490700)(3.860000,6.451460)(3.880000,6.493260)(3.900000,6.460670)(3.920000,6.529730)(3.940000,6.482060)(3.960000,6.283110)(3.980000,6.342830)(4.000000,6.377310)(4.020000,6.295060)(4.040000,6.283560)(4.060000,6.356040)(4.080000,6.397570)(4.100000,6.310690)(4.120000,6.233400)(4.140000,6.334220)(4.160000,6.272100)(4.180000,6.407480)(4.200000,6.334790)(4.220000,6.192500)(4.240000,6.210670)(4.260000,6.200000)(4.280000,6.262480)(4.300000,6.167950)(4.320000,6.117580)(4.340000,6.266450)(4.360000,6.227610)(4.380000,6.021180)(4.400000,6.046850)(4.420000,6.051360)(4.440000,6.037150)(4.460000,6.196780)(4.480000,5.952890)(4.500000,5.918250)(4.520000,5.965010)(4.540000,5.958220)(4.560000,5.906010)(4.580000,5.933910)(4.600000,5.987240)(4.620000,5.851170)(4.640000,5.854980)(4.660000,5.867730)(4.680000,5.994850)(4.700000,5.901250)(4.720000,5.791470)(4.740000,5.868250)(4.760000,5.845720)(4.780000,5.827660)(4.800000,5.635280)(4.820000,5.883520)(4.840000,5.789340)(4.860000,5.600740)(4.880000,5.772380)(4.900000,5.594350)(4.920000,5.737000)(4.940000,5.746050)(4.960000,5.825350)(4.980000,5.513570)(5.000000,5.427600)(5.020000,5.700720)(5.040000,5.655890)(5.060000,5.741710)(5.080000,5.592060)(5.100000,5.679640)(5.120000,5.658700)(5.140000,5.529030)(5.160000,5.730840)(5.180000,5.601260)(5.200000,5.538120)(5.220000,5.570930)(5.240000,5.434450)(5.260000,5.529200)(5.280000,5.477960)(5.300000,5.722900)(5.320000,5.457390)(5.340000,5.496650)(5.360000,5.179110)(5.380000,5.595990)(5.400000,5.372380)(5.420000,5.497350)(5.440000,5.219270)(5.460000,5.394090)(5.480000,5.409310)(5.500000,5.245100)(5.520000,5.268590)(5.540000,5.384840)(5.560000,5.360940)(5.580000,5.409930)(5.600000,5.349980)(5.620000,5.436530)(5.640000,5.186070)(5.660000,5.336220)(5.680000,5.361520)(5.700000,5.354550)(5.720000,5.114230)(5.740000,5.219210)(5.760000,5.085870)(5.780000,5.222300)(5.800000,5.038580)(5.820000,5.230440)(5.840000,5.216010)(5.860000,5.173120)(5.880000,5.262310)(5.900000,5.070390)(5.920000,5.125980)(5.940000,5.326980)(5.960000,5.221590)(5.980000,5.043000)(6.000000,5.182940)(6.020000,4.992870)(6.040000,4.976310)(6.060000,5.136430)(6.080000,5.334220)(6.100000,5.067030)(6.120000,5.087280)(6.140000,5.024990)(6.160000,5.211700)(6.180000,4.980490)(6.200000,4.967630)(6.220000,4.995040)(6.240000,5.134040)(6.260000,5.026820)(6.280000,4.981520)(6.300000,5.205980)(6.320000,5.042170)(6.340000,5.047340)(6.360000,4.899460)(6.380000,5.039080)(6.400000,5.096450)(6.420000,5.084300)(6.440000,5.021530)(6.460000,4.929410)(6.480000,5.021370)(6.500000,4.720420)(6.520000,4.883630)(6.540000,4.891340)(6.560000,4.904460)(6.580000,4.925650)(6.600000,4.611910)(6.620000,5.085220)(6.640000,4.901540)(6.660000,4.836160)(6.680000,4.846170)(6.700000,5.000700)(6.720000,4.931710)(6.740000,4.805870)(6.760000,4.819080)(6.780000,4.810570)(6.800000,4.859530)(6.820000,4.767830)(6.840000,4.711670)(6.860000,4.891240)(6.880000,4.804810)(6.900000,4.946840)(6.920000,4.770380)(6.940000,4.847920)(6.960000,4.681370)(6.980000,4.784100)(7.000000,4.605550)(7.020000,4.722530)(7.040000,4.650340)(7.060000,4.899960)(7.080000,4.691000)(7.100000,4.459150)(7.120000,4.956800)(7.140000,4.501320)(7.160000,4.577330)(7.180000,4.527730)(7.200000,4.600910)(7.220000,4.847260)(7.240000,4.716690)(7.260000,4.579330)(7.280000,4.546740)(7.300000,4.841230)(7.320000,4.739520)(7.340000,4.627640)(7.360000,4.553230)(7.380000,4.493130)(7.400000,4.694970)(7.420000,4.525940)(7.440000,4.585040)(7.460000,4.730140)(7.480000,4.622650)(7.500000,4.652400)(7.520000,4.292170)(7.540000,4.641400)(7.560000,4.548230)(7.580000,4.853490)(7.600000,4.530970)(7.620000,4.699450)(7.640000,4.692690)(7.660000,4.570570)(7.680000,4.594660)(7.700000,4.560970)(7.720000,4.428950)(7.740000,4.363400)(7.760000,4.468230)(7.780000,4.441520)(7.800000,4.437340)(7.820000,4.577790)(7.840000,4.714360)(7.860000,4.425060)(7.880000,4.991010)(7.900000,4.647870)(7.920000,4.434090)(7.940000,4.451910)(7.960000,4.537390)(7.980000,4.592740)(8.000000,4.535480)(8.020000,4.760980)(8.040000,4.333680)(8.060000,4.622680)(8.080000,4.395670)(8.100000,4.305980)(8.120000,4.432900)(8.140000,4.391290)(8.160000,4.222930)(8.180000,4.411290)(8.200000,4.334580)(8.220000,4.539510)(8.240000,4.387110)(8.260000,4.386270)(8.280000,4.327190)(8.300000,4.450630)(8.320000,4.220800)(8.340000,4.335860)(8.360000,4.362900)(8.380000,4.379970)(8.400000,4.357230)(8.420000,4.333430)(8.440000,4.366280)(8.460000,4.457880)(8.480000,4.465300)(8.500000,4.370410)(8.520000,4.345750)(8.540000,4.441340)(8.560000,4.484120)(8.580000,4.313310)(8.600000,4.414480)(8.620000,4.571280)(8.640000,4.062230)(8.660000,4.341440)(8.680000,4.193920)(8.700000,4.250690)(8.720000,4.239260)(8.740000,4.260920)(8.760000,4.238430)(8.780000,4.249030)(8.800000,4.414290)(8.820000,4.648170)(8.840000,4.074860)(8.860000,4.265540)(8.880000,4.159220)(8.900000,4.106150)(8.920000,4.502740)(8.940000,4.129680)(8.960000,4.432340)(8.980000,4.189830)(9.000000,4.211070)(9.020000,4.043150)(9.040000,4.048780)(9.060000,4.226440)(9.080000,4.207200)(9.100000,4.152540)(9.120000,4.290370)(9.140000,4.248320)(9.160000,4.206100)(9.180000,4.310190)(9.200000,4.079550)(9.220000,4.275420)(9.240000,4.206440)(9.260000,4.467840)(9.280000,4.506060)(9.300000,4.129030)(9.320000,4.108280)(9.340000,4.096750)(9.360000,3.992990)(9.380000,4.172620)(9.400000,3.951570)(9.420000,4.253370)(9.440000,4.312470)(9.460000,4.363510)(9.480000,4.090050)(9.500000,4.101060)(9.520000,4.266410)(9.540000,4.072830)(9.560000,4.120530)(9.580000,4.211730)(9.600000,4.066460)(9.620000,4.128170)(9.640000,4.202240)(9.660000,4.071430)(9.680000,4.074220)(9.700000,3.994820)(9.720000,3.911730)(9.740000,3.990160)(9.760000,4.175880)(9.780000,4.127030)(9.800000,4.240540)(9.820000,4.369520)(9.840000,4.207720)(9.860000,4.303130)(9.880000,3.888890)(9.900000,4.048270)(9.920000,4.003130)(9.940000,4.120450)(9.960000,4.207290)(9.980000,4.418640)}; 
\addplot[color=taskyblue,ultra thick,dashed] coordinates {(0.000000,6.907760)(0.020000,7.347970)(0.040000,6.958230)(0.060000,6.903800)(0.080000,6.922580)(0.100000,6.958520)(0.120000,6.920880)(0.140000,6.950400)(0.160000,6.905110)(0.180000,6.904310)(0.200000,6.904570)(0.220000,6.907800)(0.240000,6.905030)(0.260000,6.907900)(0.280000,6.904870)(0.300000,6.902650)(0.320000,6.904140)(0.340000,6.925870)(0.360000,6.902630)(0.380000,6.906240)(0.400000,6.909810)(0.420000,6.910500)(0.440000,6.907480)(0.460000,6.906250)(0.480000,6.907080)(0.500000,6.909570)(0.520000,6.908340)(0.540000,6.905940)(0.560000,6.908080)(0.580000,6.910070)(0.600000,6.906130)(0.620000,6.918570)(0.640000,6.916880)(0.660000,6.906650)(0.680000,6.905350)(0.700000,6.909300)(0.720000,6.909020)(0.740000,6.907180)(0.760000,6.904480)(0.780000,6.907680)(0.800000,6.903630)(0.820000,6.914900)(0.840000,6.913650)(0.860000,6.905910)(0.880000,6.905280)(0.900000,6.909430)(0.920000,6.907540)(0.940000,6.905000)(0.960000,6.903860)(0.980000,6.903960)(1.000000,6.909190)(1.020000,6.904700)(1.040000,6.898900)(1.060000,6.907540)(1.080000,6.914370)(1.100000,6.890910)(1.120000,6.863770)(1.140000,6.866070)(1.160000,6.814090)(1.180000,6.868170)(1.200000,6.873940)(1.220000,6.869520)(1.240000,6.868600)(1.260000,6.840710)(1.280000,6.873350)(1.300000,6.819980)(1.320000,6.840830)(1.340000,6.824230)(1.360000,6.798210)(1.380000,6.785920)(1.400000,6.783580)(1.420000,6.857120)(1.440000,6.787380)(1.460000,6.783630)(1.480000,6.811730)(1.500000,6.772110)(1.520000,6.809830)(1.540000,6.759400)(1.560000,6.768520)(1.580000,6.750890)(1.600000,6.818240)(1.620000,6.756660)(1.640000,6.613500)(1.660000,6.720510)(1.680000,6.679540)(1.700000,6.776210)(1.720000,6.663040)(1.740000,6.668450)(1.760000,6.651910)(1.780000,6.688700)(1.800000,6.696490)(1.820000,6.744120)(1.840000,6.657610)(1.860000,6.624480)(1.880000,6.635880)(1.900000,6.485440)(1.920000,6.564340)(1.940000,6.522130)(1.960000,6.639190)(1.980000,6.535470)(2.000000,6.483150)(2.020000,6.554740)(2.040000,6.545640)(2.060000,6.493330)(2.080000,6.419150)(2.100000,6.494070)(2.120000,6.445530)(2.140000,6.416090)(2.160000,6.409160)(2.180000,6.382800)(2.200000,6.380590)(2.220000,6.555800)(2.240000,6.383910)(2.260000,6.336150)(2.280000,6.361700)(2.300000,6.325300)(2.320000,6.279470)(2.340000,6.327740)(2.360000,6.176440)(2.380000,6.127120)(2.400000,6.334390)(2.420000,6.273120)(2.440000,6.331170)(2.460000,6.176290)(2.480000,6.125010)(2.500000,6.213800)(2.520000,6.102000)(2.540000,6.138680)(2.560000,6.054710)(2.580000,6.356840)(2.600000,6.139030)(2.620000,6.159380)(2.640000,6.229170)(2.660000,6.019750)(2.680000,6.069830)(2.700000,6.097600)(2.720000,6.103640)(2.740000,6.111790)(2.760000,6.045350)(2.780000,6.047890)(2.800000,5.924910)(2.820000,6.046740)(2.840000,5.961980)(2.860000,6.119580)(2.880000,5.807050)(2.900000,6.018810)(2.920000,6.108770)(2.940000,5.977420)(2.960000,5.899160)(2.980000,5.797840)(3.000000,5.959410)(3.020000,5.898290)(3.040000,5.927600)(3.060000,6.082850)(3.080000,6.018200)(3.100000,5.964780)(3.120000,5.843220)(3.140000,5.564380)(3.160000,5.846380)(3.180000,5.944230)(3.200000,5.969640)(3.220000,5.807130)(3.240000,5.859900)(3.260000,5.656790)(3.280000,5.715050)(3.300000,5.782450)(3.320000,5.575190)(3.340000,5.802470)(3.360000,5.795960)(3.380000,5.800780)(3.400000,5.735400)(3.420000,5.824450)(3.440000,5.710380)(3.460000,5.777620)(3.480000,5.559960)(3.500000,5.740530)(3.520000,5.556370)(3.540000,5.532180)(3.560000,5.731570)(3.580000,5.619060)(3.600000,5.583880)(3.620000,5.617480)(3.640000,5.591540)(3.660000,5.748400)(3.680000,5.553640)(3.700000,5.644580)(3.720000,5.549850)(3.740000,5.709700)(3.760000,5.681650)(3.780000,5.462090)(3.800000,5.612400)(3.820000,5.490670)(3.840000,5.365090)(3.860000,5.680270)(3.880000,5.555000)(3.900000,5.509910)(3.920000,5.766210)(3.940000,5.562100)(3.960000,5.444980)(3.980000,5.371960)(4.000000,5.413500)(4.020000,5.290830)(4.040000,5.434290)(4.060000,5.484660)(4.080000,5.616540)(4.100000,5.627560)(4.120000,5.418780)(4.140000,5.384730)(4.160000,5.390580)(4.180000,5.430610)(4.200000,5.540920)(4.220000,5.415990)(4.240000,5.527190)(4.260000,5.496020)(4.280000,5.362230)(4.300000,5.500110)(4.320000,5.363120)(4.340000,5.519050)(4.360000,5.510470)(4.380000,5.300630)(4.400000,5.217080)(4.420000,5.351160)(4.440000,5.236790)(4.460000,5.476840)(4.480000,5.280080)(4.500000,5.228130)(4.520000,5.309970)(4.540000,5.328640)(4.560000,5.318560)(4.580000,5.359280)(4.600000,5.242340)(4.620000,5.290440)(4.640000,5.361030)(4.660000,5.153330)(4.680000,5.379410)(4.700000,5.380760)(4.720000,5.386250)(4.740000,5.339890)(4.760000,5.299150)(4.780000,5.308640)(4.800000,4.962440)(4.820000,5.351450)(4.840000,5.353390)(4.860000,5.012910)(4.880000,5.251860)(4.900000,5.031520)(4.920000,5.265840)(4.940000,5.283530)(4.960000,5.280340)(4.980000,4.954040)(5.000000,4.989280)(5.020000,5.118380)(5.040000,5.269730)(5.060000,5.343170)(5.080000,5.207490)(5.100000,5.283370)(5.120000,5.203170)(5.140000,5.174090)(5.160000,5.211850)(5.180000,5.115550)(5.200000,5.161150)(5.220000,4.932970)(5.240000,4.955270)(5.260000,5.184100)(5.280000,5.144910)(5.300000,5.274180)(5.320000,4.844930)(5.340000,5.091560)(5.360000,4.877850)(5.380000,5.281400)(5.400000,4.958250)(5.420000,5.152750)(5.440000,4.914180)(5.460000,5.102230)(5.480000,5.195410)(5.500000,4.944310)(5.520000,4.970930)(5.540000,5.221920)(5.560000,4.953300)(5.580000,5.057240)(5.600000,4.950670)(5.620000,5.257530)(5.640000,4.988060)(5.660000,5.233510)(5.680000,5.113490)(5.700000,5.035130)(5.720000,4.957480)(5.740000,4.912540)(5.760000,4.971070)(5.780000,4.966190)(5.800000,4.762040)(5.820000,5.152100)(5.840000,5.018400)(5.860000,4.834920)(5.880000,5.026080)(5.900000,4.774520)(5.920000,4.922740)(5.940000,4.987390)(5.960000,4.851260)(5.980000,4.902940)(6.000000,4.973610)(6.020000,4.723700)(6.040000,4.946940)(6.060000,4.884950)(6.080000,5.002230)(6.100000,4.915430)(6.120000,4.796180)(6.140000,4.869930)(6.160000,5.090840)(6.180000,4.818390)(6.200000,4.649680)(6.220000,5.002630)(6.240000,4.856350)(6.260000,4.852950)(6.280000,4.842060)(6.300000,5.017650)(6.320000,5.137590)(6.340000,4.887740)(6.360000,4.769310)(6.380000,4.971420)(6.400000,4.845030)(6.420000,4.866330)(6.440000,5.011380)(6.460000,4.665750)(6.480000,4.819860)(6.500000,4.620600)(6.520000,4.799950)(6.540000,4.742180)(6.560000,4.676490)(6.580000,4.795590)(6.600000,4.663660)(6.620000,4.829210)(6.640000,4.884080)(6.660000,4.642330)(6.680000,4.736070)(6.700000,4.827510)(6.720000,4.945070)(6.740000,4.812220)(6.760000,4.537980)(6.780000,4.814730)(6.800000,4.566550)(6.820000,4.538380)(6.840000,4.535120)(6.860000,4.794330)(6.880000,4.723630)(6.900000,4.846560)(6.920000,4.794120)(6.940000,4.898700)(6.960000,4.707340)(6.980000,4.689650)(7.000000,4.572100)(7.020000,4.623250)(7.040000,4.715810)(7.060000,4.885580)(7.080000,4.652180)(7.100000,4.469910)(7.120000,5.014770)(7.140000,4.543710)(7.160000,4.573690)(7.180000,4.454080)(7.200000,4.521150)(7.220000,4.925770)(7.240000,4.646290)(7.260000,4.524110)(7.280000,4.599260)(7.300000,4.816310)(7.320000,4.758990)(7.340000,4.397970)(7.360000,4.664790)(7.380000,4.588180)(7.400000,4.715710)(7.420000,4.479020)(7.440000,4.749880)(7.460000,4.818380)(7.480000,4.562900)(7.500000,4.655970)(7.520000,4.389860)(7.540000,4.784700)(7.560000,4.594910)(7.580000,4.639670)(7.600000,4.667000)(7.620000,4.702740)(7.640000,4.832540)(7.660000,4.514250)(7.680000,4.648420)(7.700000,4.582720)(7.720000,4.507270)(7.740000,4.457110)(7.760000,4.560730)(7.780000,4.490490)(7.800000,4.566050)(7.820000,4.767280)(7.840000,4.669010)(7.860000,4.417430)(7.880000,4.797820)(7.900000,4.629900)(7.920000,4.502200)(7.940000,4.559200)(7.960000,4.550980)(7.980000,4.669130)(8.000000,4.928580)(8.020000,4.623350)(8.040000,4.361520)(8.060000,4.473350)(8.080000,4.541700)(8.100000,4.349270)(8.120000,4.545900)(8.140000,4.377880)(8.160000,4.497430)(8.180000,4.620180)(8.200000,4.418010)(8.220000,4.580800)(8.240000,4.424790)(8.260000,4.298510)(8.280000,4.393360)(8.300000,4.482930)(8.320000,4.186430)(8.340000,4.353700)(8.360000,4.311750)(8.380000,4.396420)(8.400000,4.339070)(8.420000,4.407940)(8.440000,4.485960)(8.460000,4.554010)(8.480000,4.508820)(8.500000,4.544860)(8.520000,4.539130)(8.540000,4.548910)(8.560000,4.555440)(8.580000,4.401680)(8.600000,4.593830)(8.620000,4.689280)(8.640000,4.203180)(8.660000,4.460190)(8.680000,4.324000)(8.700000,4.448690)(8.720000,4.302410)(8.740000,4.590270)(8.760000,4.335840)(8.780000,4.366690)(8.800000,4.545850)(8.820000,4.628500)(8.840000,4.335030)(8.860000,4.395740)(8.880000,4.378950)(8.900000,4.359740)(8.920000,4.589180)(8.940000,4.401000)(8.960000,4.474830)(8.980000,4.511110)(9.000000,4.358450)(9.020000,4.245680)(9.040000,4.111150)(9.060000,4.237420)(9.080000,4.493430)(9.100000,4.309250)(9.120000,4.310010)(9.140000,4.445300)(9.160000,4.445090)(9.180000,4.454790)(9.200000,4.193470)(9.220000,4.302120)(9.240000,4.333850)(9.260000,4.725330)(9.280000,4.755170)(9.300000,4.194820)(9.320000,4.100400)(9.340000,4.387640)(9.360000,4.222520)(9.380000,4.228480)(9.400000,4.083390)(9.420000,4.461630)(9.440000,4.301740)(9.460000,4.343090)(9.480000,4.145710)(9.500000,4.394320)(9.520000,4.570620)(9.540000,4.355100)(9.560000,4.296920)(9.580000,4.240970)(9.600000,4.215720)(9.620000,4.396660)(9.640000,4.380550)(9.660000,4.210100)(9.680000,4.304020)(9.700000,4.332070)(9.720000,3.992750)(9.740000,4.257260)(9.760000,4.272090)(9.780000,4.228890)(9.800000,4.422490)(9.820000,4.502890)(9.840000,4.426810)(9.860000,4.484000)(9.880000,4.070010)(9.900000,4.121010)(9.920000,4.049580)(9.940000,4.040880)(9.960000,4.444480)(9.980000,4.495730)}; 
\addplot[color=ta3scarletred,ultra thick,solid] coordinates {(0.000000,6.907760)(0.020000,6.959670)(0.040000,6.919330)(0.060000,6.865620)(0.080000,6.837370)(0.100000,6.773890)(0.120000,6.758020)(0.140000,6.656780)(0.160000,6.625700)(0.180000,6.567780)(0.200000,6.498690)(0.220000,6.394970)(0.240000,6.448970)(0.260000,6.409130)(0.280000,6.272410)(0.300000,6.336520)(0.320000,6.323920)(0.340000,6.240670)(0.360000,6.317750)(0.380000,6.275100)(0.400000,6.143210)(0.420000,6.145610)(0.440000,6.060270)(0.460000,6.220110)(0.480000,6.055810)(0.500000,6.019760)(0.520000,5.942650)(0.540000,6.073620)(0.560000,6.017970)(0.580000,6.012480)(0.600000,5.971500)(0.620000,6.036030)(0.640000,5.904500)(0.660000,5.980200)(0.680000,5.885920)(0.700000,5.828830)(0.720000,5.870200)(0.740000,5.896780)(0.760000,5.852050)(0.780000,5.896040)(0.800000,5.792840)(0.820000,5.835390)(0.840000,5.808250)(0.860000,5.872850)(0.880000,5.691470)(0.900000,5.671860)(0.920000,5.523240)(0.940000,5.653080)(0.960000,5.439910)(0.980000,5.565440)(1.000000,5.405150)(1.020000,5.712230)(1.040000,5.720420)(1.060000,5.685940)(1.080000,5.588340)(1.100000,5.572250)(1.120000,5.652430)(1.140000,5.519230)(1.160000,5.392050)(1.180000,5.578050)(1.200000,5.316660)(1.220000,5.647850)(1.240000,5.467760)(1.260000,5.607340)(1.280000,5.386230)(1.300000,5.569010)(1.320000,5.447850)(1.340000,5.681860)(1.360000,5.362810)(1.380000,5.408130)(1.400000,5.458180)(1.420000,5.486380)(1.440000,5.627810)(1.460000,5.409880)(1.480000,5.352250)(1.500000,5.431760)(1.520000,5.430210)(1.540000,5.300060)(1.560000,5.485200)(1.580000,5.224670)(1.600000,5.355770)(1.620000,5.263950)(1.640000,5.164960)(1.660000,5.198940)(1.680000,5.360470)(1.700000,5.428350)(1.720000,5.442690)(1.740000,5.217220)(1.760000,5.363160)(1.780000,5.086260)(1.800000,5.225550)(1.820000,5.140590)(1.840000,5.241190)(1.860000,5.177430)(1.880000,5.337590)(1.900000,4.939220)(1.920000,5.062340)(1.940000,5.136980)(1.960000,5.229740)(1.980000,5.190440)(2.000000,5.200220)(2.020000,5.225940)(2.040000,5.173800)(2.060000,4.945120)(2.080000,5.027070)(2.100000,5.070250)(2.120000,5.085850)(2.140000,5.101980)(2.160000,5.206690)(2.180000,5.009200)(2.200000,5.168110)(2.220000,5.086840)(2.240000,5.102450)(2.260000,4.965960)(2.280000,4.846180)(2.300000,5.236410)(2.320000,5.127360)(2.340000,5.157180)(2.360000,4.820770)(2.380000,4.848010)(2.400000,4.997550)(2.420000,5.152030)(2.440000,5.142030)(2.460000,5.049840)(2.480000,5.029580)(2.500000,4.898620)(2.520000,4.927820)(2.540000,4.869800)(2.560000,4.846030)(2.580000,5.155550)(2.600000,5.075640)(2.620000,4.873140)(2.640000,5.113980)(2.660000,4.956400)(2.680000,4.902330)(2.700000,5.046540)(2.720000,4.971720)(2.740000,4.953370)(2.760000,4.918330)(2.780000,4.777320)(2.800000,4.831030)(2.820000,4.953270)(2.840000,4.904520)(2.860000,4.848590)(2.880000,4.444080)(2.900000,4.935060)(2.920000,5.175800)(2.940000,4.696180)(2.960000,4.714020)(2.980000,4.630650)(3.000000,4.924860)(3.020000,4.771840)(3.040000,4.626900)(3.060000,4.939900)(3.080000,4.679290)(3.100000,4.916730)(3.120000,4.731320)(3.140000,4.642410)(3.160000,4.854470)(3.180000,4.790920)(3.200000,4.871040)(3.220000,4.760080)(3.240000,4.767210)(3.260000,4.846730)(3.280000,4.576880)(3.300000,4.753180)(3.320000,4.574000)(3.340000,4.665290)(3.360000,4.794550)(3.380000,4.792940)(3.400000,4.532940)(3.420000,4.767390)(3.440000,4.808910)(3.460000,5.010910)(3.480000,4.341760)(3.500000,4.577080)(3.520000,4.646030)(3.540000,4.544520)(3.560000,4.749090)(3.580000,4.719450)(3.600000,4.687380)(3.620000,4.685370)(3.640000,4.671910)(3.660000,4.831300)(3.680000,4.372150)(3.700000,4.770040)(3.720000,4.638100)(3.740000,4.695000)(3.760000,4.519450)(3.780000,4.504550)(3.800000,4.586420)(3.820000,4.407480)(3.840000,4.380060)(3.860000,4.854490)(3.880000,4.695830)(3.900000,4.368930)(3.920000,4.835270)(3.940000,4.597960)(3.960000,4.433240)(3.980000,4.581700)(4.000000,4.476310)(4.020000,4.451970)(4.040000,4.411370)(4.060000,4.629610)(4.080000,4.874960)(4.100000,4.582450)(4.120000,4.556110)(4.140000,4.471060)(4.160000,4.439680)(4.180000,4.492140)(4.200000,4.863300)(4.220000,4.592810)(4.240000,4.810100)(4.260000,4.597380)(4.280000,4.533960)(4.300000,4.584240)(4.320000,4.538800)(4.340000,4.729410)(4.360000,4.692370)(4.380000,4.416300)(4.400000,4.183580)(4.420000,4.598420)(4.440000,4.375610)(4.460000,4.602850)(4.480000,4.474080)(4.500000,4.396570)(4.520000,4.285250)(4.540000,4.464390)(4.560000,4.443230)(4.580000,4.467310)(4.600000,4.647790)(4.620000,4.502340)(4.640000,4.472620)(4.660000,4.414920)(4.680000,4.685240)(4.700000,4.660980)(4.720000,4.470380)(4.740000,4.475730)(4.760000,4.370950)(4.780000,4.403620)(4.800000,4.359140)(4.820000,4.626360)(4.840000,4.683850)(4.860000,4.180590)(4.880000,4.400700)(4.900000,4.241080)(4.920000,4.436760)(4.940000,4.348340)(4.960000,4.475340)(4.980000,4.132400)(5.000000,4.194580)(5.020000,4.393580)(5.040000,4.271850)(5.060000,4.370860)(5.080000,4.444360)(5.100000,4.407830)(5.120000,4.473940)(5.140000,4.326020)(5.160000,4.563960)(5.180000,4.337160)(5.200000,4.503340)(5.220000,4.334350)(5.240000,4.367060)(5.260000,4.371550)(5.280000,4.443030)(5.300000,4.475860)(5.320000,4.261560)(5.340000,4.441240)(5.360000,4.039070)(5.380000,4.413280)(5.400000,4.293480)(5.420000,4.526040)(5.440000,4.354040)(5.460000,4.479540)(5.480000,4.448900)(5.500000,4.222160)(5.520000,4.355750)(5.540000,4.365640)(5.560000,4.201140)(5.580000,4.297650)(5.600000,4.372500)(5.620000,4.331330)(5.640000,4.242270)(5.660000,4.361020)(5.680000,4.344310)(5.700000,4.475520)(5.720000,4.144560)(5.740000,4.348540)(5.760000,4.193080)(5.780000,4.198210)(5.800000,4.087410)(5.820000,4.308690)(5.840000,4.146400)(5.860000,4.224240)(5.880000,4.312660)(5.900000,4.155820)(5.920000,4.157920)(5.940000,4.269170)(5.960000,4.236460)(5.980000,4.191120)(6.000000,4.306510)(6.020000,3.897150)(6.040000,3.999170)(6.060000,4.169190)(6.080000,4.264260)(6.100000,4.101550)(6.120000,4.194790)(6.140000,4.288370)(6.160000,4.340220)(6.180000,3.958560)(6.200000,4.079090)(6.220000,4.235520)(6.240000,4.194200)(6.260000,4.044360)(6.280000,4.231040)(6.300000,4.375690)(6.320000,4.283770)(6.340000,4.336180)(6.360000,4.079650)(6.380000,4.221410)(6.400000,4.253880)(6.420000,4.224610)(6.440000,4.336990)(6.460000,3.850500)(6.480000,4.122070)(6.500000,4.012980)(6.520000,4.096290)(6.540000,4.059060)(6.560000,4.081760)(6.580000,4.136850)(6.600000,3.904280)(6.620000,4.291900)(6.640000,4.025310)(6.660000,4.021630)(6.680000,4.133420)(6.700000,4.084150)(6.720000,4.324530)(6.740000,4.157900)(6.760000,3.953370)(6.780000,4.012240)(6.800000,4.070440)(6.820000,4.087780)(6.840000,3.818270)(6.860000,4.213520)(6.880000,4.100030)(6.900000,4.057890)(6.920000,3.996580)(6.940000,4.198050)(6.960000,3.978830)(6.980000,4.049370)(7.000000,4.009460)(7.020000,3.884250)(7.040000,3.900480)(7.060000,4.221390)(7.080000,4.194970)(7.100000,3.860100)(7.120000,4.329970)(7.140000,4.010170)(7.160000,4.016300)(7.180000,3.938940)(7.200000,3.782040)(7.220000,4.208050)(7.240000,4.121380)(7.260000,3.760140)(7.280000,3.987080)(7.300000,4.149820)(7.320000,3.957210)(7.340000,3.873710)(7.360000,3.984980)(7.380000,4.026920)(7.400000,4.092460)(7.420000,3.969710)(7.440000,3.961550)(7.460000,4.087590)(7.480000,3.777050)(7.500000,4.135190)(7.520000,3.673280)(7.540000,4.039620)(7.560000,3.920500)(7.580000,4.128160)(7.600000,3.964170)(7.620000,4.184640)(7.640000,4.140100)(7.660000,4.007950)(7.680000,4.061160)(7.700000,3.938770)(7.720000,3.895090)(7.740000,3.692090)(7.760000,3.952080)(7.780000,3.798030)(7.800000,3.843680)(7.820000,4.135380)(7.840000,4.043550)(7.860000,3.818210)(7.880000,4.136720)(7.900000,4.016320)(7.920000,3.937740)(7.940000,3.921210)(7.960000,3.992170)(7.980000,4.061300)(8.000000,4.121700)(8.020000,4.116100)(8.040000,3.854220)(8.060000,4.121750)(8.080000,4.056860)(8.100000,3.783210)(8.120000,3.829310)(8.140000,3.848110)(8.160000,3.831700)(8.180000,4.040340)(8.200000,3.879870)(8.220000,4.010820)(8.240000,3.865990)(8.260000,3.876930)(8.280000,3.867010)(8.300000,3.811990)(8.320000,3.678040)(8.340000,3.944560)(8.360000,3.751800)(8.380000,3.853020)(8.400000,3.822760)(8.420000,3.900590)(8.440000,3.956860)(8.460000,4.001860)(8.480000,3.955590)(8.500000,3.919860)(8.520000,3.906510)(8.540000,3.964070)(8.560000,3.756760)(8.580000,3.816530)(8.600000,4.008960)(8.620000,4.026470)(8.640000,3.785710)(8.660000,3.915230)(8.680000,3.905430)(8.700000,3.806580)(8.720000,3.776470)(8.740000,3.824040)(8.760000,3.734730)(8.780000,3.761070)(8.800000,4.030030)(8.820000,4.147750)(8.840000,3.751430)(8.860000,3.928540)(8.880000,3.764340)(8.900000,3.755760)(8.920000,4.123030)(8.940000,3.630080)(8.960000,3.871120)(8.980000,3.873210)(9.000000,3.666820)(9.020000,3.692980)(9.040000,3.596180)(9.060000,3.628460)(9.080000,3.844270)(9.100000,3.668920)(9.120000,3.787260)(9.140000,3.993360)(9.160000,3.946540)(9.180000,3.929590)(9.200000,3.788940)(9.220000,3.876980)(9.240000,3.750130)(9.260000,4.051610)(9.280000,3.961810)(9.300000,3.811380)(9.320000,3.556110)(9.340000,3.951700)(9.360000,3.718260)(9.380000,3.669330)(9.400000,3.547820)(9.420000,3.818170)(9.440000,3.947590)(9.460000,3.941500)(9.480000,3.700640)(9.500000,3.881050)(9.520000,3.911600)(9.540000,3.914610)(9.560000,3.840030)(9.580000,3.887320)(9.600000,3.766960)(9.620000,3.853660)(9.640000,3.745830)(9.660000,3.661670)(9.680000,3.479280)(9.700000,3.470930)(9.720000,3.608680)(9.740000,3.613530)(9.760000,3.838550)(9.780000,3.676800)(9.800000,3.982520)(9.820000,3.869720)(9.840000,3.770490)(9.860000,3.920510)(9.880000,3.530720)(9.900000,3.695790)(9.920000,3.676400)(9.940000,3.686990)(9.960000,3.906940)(9.980000,4.125460)}; 
\addplot[color=ta3orange,ultra thick,solid] coordinates {(0.000000,6.907760)(0.020000,6.952760)(0.040000,6.898700)(0.060000,6.773880)(0.080000,6.716340)(0.100000,6.658190)(0.120000,6.541950)(0.140000,6.471600)(0.160000,6.386700)(0.180000,6.371540)(0.200000,6.259810)(0.220000,6.205670)(0.240000,6.225590)(0.260000,6.048040)(0.280000,5.998740)(0.300000,6.126190)(0.320000,6.052080)(0.340000,5.933960)(0.360000,5.964600)(0.380000,6.035850)(0.400000,5.869390)(0.420000,5.787270)(0.440000,5.716380)(0.460000,5.824070)(0.480000,5.732360)(0.500000,5.734720)(0.520000,5.681220)(0.540000,5.666090)(0.560000,5.715940)(0.580000,5.682900)(0.600000,5.629920)(0.620000,5.590150)(0.640000,5.607890)(0.660000,5.668230)(0.680000,5.653900)(0.700000,5.421360)(0.720000,5.602640)(0.740000,5.598050)(0.760000,5.528940)(0.780000,5.597640)(0.800000,5.550250)(0.820000,5.576940)(0.840000,5.553220)(0.860000,5.608300)(0.880000,5.438140)(0.900000,5.405010)(0.920000,5.151810)(0.940000,5.427120)(0.960000,5.251400)(0.980000,5.407840)(1.000000,5.239120)(1.020000,5.427250)(1.040000,5.391210)(1.060000,5.458430)(1.080000,5.363000)(1.100000,5.372530)(1.120000,5.380520)(1.140000,5.204110)(1.160000,5.107870)(1.180000,5.220250)(1.200000,5.273060)(1.220000,5.147230)(1.240000,5.208270)(1.260000,5.352530)(1.280000,5.232880)(1.300000,5.343770)(1.320000,5.037680)(1.340000,5.338050)(1.360000,5.138180)(1.380000,5.102030)(1.400000,5.118880)(1.420000,5.349840)(1.440000,5.326280)(1.460000,5.196420)(1.480000,4.986100)(1.500000,5.175890)(1.520000,5.245550)(1.540000,4.995140)(1.560000,5.260840)(1.580000,4.968860)(1.600000,5.127270)(1.620000,5.003070)(1.640000,4.869430)(1.660000,4.985050)(1.680000,5.042630)(1.700000,5.195900)(1.720000,5.135240)(1.740000,4.965500)(1.760000,4.995800)(1.780000,4.760330)(1.800000,4.893310)(1.820000,4.814480)(1.840000,4.955920)(1.860000,4.970940)(1.880000,5.023220)(1.900000,4.710890)(1.920000,4.753380)(1.940000,4.926120)(1.960000,4.882580)(1.980000,4.825250)(2.000000,4.892370)(2.020000,4.920310)(2.040000,4.879280)(2.060000,4.819130)(2.080000,4.756660)(2.100000,4.889590)(2.120000,4.810500)(2.140000,4.833140)(2.160000,4.810910)(2.180000,4.789150)(2.200000,4.966360)(2.220000,4.787260)(2.240000,4.867300)(2.260000,4.729330)(2.280000,4.686650)(2.300000,4.891150)(2.320000,4.883460)(2.340000,4.771540)(2.360000,4.609370)(2.380000,4.657490)(2.400000,4.806060)(2.420000,4.898310)(2.440000,4.932890)(2.460000,4.629130)(2.480000,4.751020)(2.500000,4.630080)(2.520000,4.624890)(2.540000,4.644800)(2.560000,4.662270)(2.580000,4.946950)(2.600000,4.771450)(2.620000,4.614190)(2.640000,4.856030)(2.660000,4.578870)(2.680000,4.515150)(2.700000,4.744320)(2.720000,4.652670)(2.740000,4.760360)(2.760000,4.466350)(2.780000,4.635110)(2.800000,4.641960)(2.820000,4.585090)(2.840000,4.643450)(2.860000,4.587930)(2.880000,4.189650)(2.900000,4.754290)(2.920000,4.846500)(2.940000,4.427070)(2.960000,4.527760)(2.980000,4.396750)(3.000000,4.788150)(3.020000,4.557630)(3.040000,4.334570)(3.060000,4.701530)(3.080000,4.448630)(3.100000,4.513780)(3.120000,4.461910)(3.140000,4.480770)(3.160000,4.660730)(3.180000,4.453870)(3.200000,4.579190)(3.220000,4.586380)(3.240000,4.535560)(3.260000,4.554660)(3.280000,4.305050)(3.300000,4.464120)(3.320000,4.236050)(3.340000,4.367240)(3.360000,4.514710)(3.380000,4.668970)(3.400000,4.333110)(3.420000,4.414920)(3.440000,4.470420)(3.460000,4.798630)(3.480000,4.131670)(3.500000,4.436560)(3.520000,4.388640)(3.540000,4.234780)(3.560000,4.553450)(3.580000,4.441900)(3.600000,4.438660)(3.620000,4.360310)(3.640000,4.524290)(3.660000,4.600400)(3.680000,4.314070)(3.700000,4.538400)(3.720000,4.381830)(3.740000,4.375020)(3.760000,4.271040)(3.780000,4.346510)(3.800000,4.235860)(3.820000,4.265840)(3.840000,4.110640)(3.860000,4.641780)(3.880000,4.332990)(3.900000,4.173790)(3.920000,4.602910)(3.940000,4.484130)(3.960000,4.246280)(3.980000,4.380320)(4.000000,4.197500)(4.020000,4.205370)(4.040000,4.421480)(4.060000,4.336940)(4.080000,4.560050)(4.100000,4.274960)(4.120000,4.293440)(4.140000,4.237440)(4.160000,4.282030)(4.180000,4.376540)(4.200000,4.577410)(4.220000,4.315280)(4.240000,4.430180)(4.260000,4.449740)(4.280000,4.314440)(4.300000,4.394660)(4.320000,4.262610)(4.340000,4.397820)(4.360000,4.432250)(4.380000,4.191060)(4.400000,4.051430)(4.420000,4.277040)(4.440000,4.114290)(4.460000,4.520850)(4.480000,4.276120)(4.500000,4.153060)(4.520000,4.076780)(4.540000,4.250670)(4.560000,4.197440)(4.580000,4.221870)(4.600000,4.232280)(4.620000,4.338070)(4.640000,4.317370)(4.660000,4.260250)(4.680000,4.467160)(4.700000,4.332550)(4.720000,4.393260)(4.740000,4.287320)(4.760000,4.179360)(4.780000,4.301410)(4.800000,4.074270)(4.820000,4.381850)(4.840000,4.441970)(4.860000,3.986550)(4.880000,4.307640)(4.900000,4.097720)(4.920000,4.056340)(4.940000,4.138950)(4.960000,4.213780)(4.980000,4.099960)(5.000000,4.044280)(5.020000,4.204370)(5.040000,4.088250)(5.060000,4.267950)(5.080000,4.114530)(5.100000,4.298470)(5.120000,4.257500)(5.140000,4.109520)(5.160000,4.336300)(5.180000,4.102410)(5.200000,4.241150)(5.220000,4.085290)(5.240000,4.100420)(5.260000,4.253290)(5.280000,4.221890)(5.300000,4.263440)(5.320000,4.144930)(5.340000,4.236350)(5.360000,3.936420)(5.380000,4.299440)(5.400000,4.121470)(5.420000,4.228680)(5.440000,4.144740)(5.460000,4.138150)(5.480000,4.191520)(5.500000,4.055300)(5.520000,4.044730)(5.540000,4.198170)(5.560000,4.157600)(5.580000,4.100480)(5.600000,4.145430)(5.620000,4.223590)(5.640000,4.068080)(5.660000,4.146010)(5.680000,4.246270)(5.700000,4.286300)(5.720000,4.020440)(5.740000,4.019980)(5.760000,4.045160)(5.780000,3.913710)(5.800000,3.767620)(5.820000,3.957850)(5.840000,3.875140)(5.860000,3.950060)(5.880000,3.983180)(5.900000,3.855490)(5.920000,3.979610)(5.940000,4.295450)(5.960000,4.067440)(5.980000,4.046140)(6.000000,3.975470)(6.020000,3.865330)(6.040000,3.805070)(6.060000,4.205110)(6.080000,4.006690)(6.100000,3.944430)(6.120000,3.895320)(6.140000,4.135090)(6.160000,4.273800)(6.180000,3.907150)(6.200000,3.927670)(6.220000,3.964510)(6.240000,3.931930)(6.260000,3.949350)(6.280000,4.174060)(6.300000,4.365280)(6.320000,4.120940)(6.340000,4.075520)(6.360000,3.908970)(6.380000,4.143520)(6.400000,4.102580)(6.420000,4.173500)(6.440000,4.063960)(6.460000,3.607780)(6.480000,3.953980)(6.500000,3.762870)(6.520000,4.039270)(6.540000,4.001080)(6.560000,3.782040)(6.580000,4.010080)(6.600000,3.737910)(6.620000,4.109330)(6.640000,4.010260)(6.660000,3.815580)(6.680000,3.813240)(6.700000,3.925960)(6.720000,4.223610)(6.740000,3.948870)(6.760000,3.790200)(6.780000,3.895100)(6.800000,3.875310)(6.820000,3.845560)(6.840000,3.699930)(6.860000,3.991880)(6.880000,3.818890)(6.900000,3.938080)(6.920000,3.732190)(6.940000,4.007530)(6.960000,3.994080)(6.980000,3.880420)(7.000000,3.805110)(7.020000,3.509850)(7.040000,3.738270)(7.060000,4.044730)(7.080000,3.923020)(7.100000,3.789040)(7.120000,4.194020)(7.140000,3.789790)(7.160000,3.937600)(7.180000,3.737360)(7.200000,3.668900)(7.220000,4.101670)(7.240000,3.914300)(7.260000,3.690890)(7.280000,3.988030)(7.300000,4.125520)(7.320000,3.841640)(7.340000,3.756910)(7.360000,3.905540)(7.380000,3.804970)(7.400000,3.910600)(7.420000,3.832880)(7.440000,3.767520)(7.460000,3.903140)(7.480000,3.707750)(7.500000,3.923410)(7.520000,3.634710)(7.540000,3.912640)(7.560000,3.842390)(7.580000,4.042070)(7.600000,3.758490)(7.620000,4.121590)(7.640000,3.959740)(7.660000,3.837070)(7.680000,3.828690)(7.700000,3.764010)(7.720000,3.711090)(7.740000,3.601060)(7.760000,3.784640)(7.780000,3.590200)(7.800000,3.841750)(7.820000,3.924600)(7.840000,3.993610)(7.860000,3.745160)(7.880000,4.059710)(7.900000,3.805080)(7.920000,3.772910)(7.940000,3.891560)(7.960000,3.829660)(7.980000,3.998680)(8.000000,4.044250)(8.020000,3.937010)(8.040000,3.696830)(8.060000,3.888290)(8.080000,3.861600)(8.100000,3.633390)(8.120000,3.679570)(8.140000,3.700420)(8.160000,3.529420)(8.180000,3.957890)(8.200000,3.603460)(8.220000,3.745030)(8.240000,3.744110)(8.260000,3.680280)(8.280000,3.761700)(8.300000,3.674640)(8.320000,3.428300)(8.340000,3.696760)(8.360000,3.504690)(8.380000,3.754700)(8.400000,3.687770)(8.420000,3.655010)(8.440000,3.711160)(8.460000,3.858150)(8.480000,3.775010)(8.500000,3.874810)(8.520000,3.710010)(8.540000,3.903390)(8.560000,3.703680)(8.580000,3.726970)(8.600000,3.745700)(8.620000,3.944430)(8.640000,3.637340)(8.660000,3.822190)(8.680000,3.688350)(8.700000,3.561390)(8.720000,3.653840)(8.740000,3.571300)(8.760000,3.601330)(8.780000,3.713310)(8.800000,3.932300)(8.820000,3.955830)(8.840000,3.422770)(8.860000,3.592340)(8.880000,3.717620)(8.900000,3.443760)(8.920000,3.992060)(8.940000,3.541940)(8.960000,3.809840)(8.980000,3.853560)(9.000000,3.600700)(9.020000,3.538770)(9.040000,3.414710)(9.060000,3.472000)(9.080000,3.706480)(9.100000,3.386290)(9.120000,3.586630)(9.140000,3.770440)(9.160000,3.714340)(9.180000,3.689160)(9.200000,3.588990)(9.220000,3.761100)(9.240000,3.702140)(9.260000,3.926680)(9.280000,3.907540)(9.300000,3.722290)(9.320000,3.298720)(9.340000,3.777320)(9.360000,3.517340)(9.380000,3.595700)(9.400000,3.488260)(9.420000,3.818210)(9.440000,3.831910)(9.460000,3.788430)(9.480000,3.563120)(9.500000,3.731280)(9.520000,3.791450)(9.540000,3.701050)(9.560000,3.641480)(9.580000,3.658000)(9.600000,3.677010)(9.620000,3.770660)(9.640000,3.592420)(9.660000,3.458330)(9.680000,3.365410)(9.700000,3.525420)(9.720000,3.342030)(9.740000,3.509790)(9.760000,3.780740)(9.780000,3.659720)(9.800000,3.863180)(9.820000,3.855160)(9.840000,3.565250)(9.860000,3.753690)(9.880000,3.348030)(9.900000,3.509230)(9.920000,3.605450)(9.940000,3.506970)(9.960000,3.759260)(9.980000,3.965180)}; 
\addplot[color=ta3chameleon,ultra thick,solid] coordinates {(0.000000,6.907760)(0.020000,6.945200)(0.040000,6.895520)(0.060000,6.774210)(0.080000,6.675780)(0.100000,6.619770)(0.120000,6.511840)(0.140000,6.445040)(0.160000,6.375360)(0.180000,6.358270)(0.200000,6.196750)(0.220000,6.135400)(0.240000,6.150290)(0.260000,6.035320)(0.280000,5.914200)(0.300000,6.047930)(0.320000,6.033810)(0.340000,5.840400)(0.360000,5.938060)(0.380000,5.892700)(0.400000,5.820010)(0.420000,5.739280)(0.440000,5.739740)(0.460000,5.767070)(0.480000,5.577850)(0.500000,5.738210)(0.520000,5.627750)(0.540000,5.498500)(0.560000,5.664330)(0.580000,5.571320)(0.600000,5.583110)(0.620000,5.505990)(0.640000,5.531920)(0.660000,5.643770)(0.680000,5.567450)(0.700000,5.450790)(0.720000,5.469170)(0.740000,5.645050)(0.760000,5.515240)(0.780000,5.527170)(0.800000,5.551260)(0.820000,5.445920)(0.840000,5.501790)(0.860000,5.567840)(0.880000,5.456870)(0.900000,5.384280)(0.920000,5.144770)(0.940000,5.378110)(0.960000,5.131770)(0.980000,5.182780)(1.000000,5.097720)(1.020000,5.364540)(1.040000,5.390290)(1.060000,5.295740)(1.080000,5.304870)(1.100000,5.400490)(1.120000,5.397470)(1.140000,5.043110)(1.160000,5.095220)(1.180000,5.001990)(1.200000,5.091700)(1.220000,5.032460)(1.240000,5.162100)(1.260000,5.233720)(1.280000,5.007600)(1.300000,5.104230)(1.320000,5.034450)(1.340000,5.238870)(1.360000,4.975180)(1.380000,5.037930)(1.400000,5.093840)(1.420000,5.188300)(1.440000,5.308910)(1.460000,5.168800)(1.480000,5.120210)(1.500000,5.048090)(1.520000,5.055700)(1.540000,4.866090)(1.560000,5.110830)(1.580000,5.023800)(1.600000,5.050530)(1.620000,4.898570)(1.640000,4.759640)(1.660000,4.929750)(1.680000,4.993950)(1.700000,5.057430)(1.720000,5.084160)(1.740000,4.872710)(1.760000,4.889040)(1.780000,4.734470)(1.800000,4.845940)(1.820000,4.836570)(1.840000,4.947250)(1.860000,4.872180)(1.880000,4.919830)(1.900000,4.530350)(1.920000,4.716670)(1.940000,4.854190)(1.960000,4.852290)(1.980000,4.766620)(2.000000,4.875980)(2.020000,4.888580)(2.040000,4.875540)(2.060000,4.727380)(2.080000,4.688610)(2.100000,4.878440)(2.120000,4.670750)(2.140000,4.900710)(2.160000,4.782590)(2.180000,4.674380)(2.200000,4.738440)(2.220000,4.707970)(2.240000,4.806600)(2.260000,4.538140)(2.280000,4.585810)(2.300000,4.843360)(2.320000,4.765280)(2.340000,4.935700)(2.360000,4.523030)(2.380000,4.571810)(2.400000,4.691590)(2.420000,4.908430)(2.440000,4.836790)(2.460000,4.628300)(2.480000,4.527140)(2.500000,4.495130)(2.520000,4.618360)(2.540000,4.492620)(2.560000,4.580420)(2.580000,4.841240)(2.600000,4.686170)(2.620000,4.458740)(2.640000,4.785810)(2.660000,4.407710)(2.680000,4.392400)(2.700000,4.657690)(2.720000,4.619010)(2.740000,4.562600)(2.760000,4.377300)(2.780000,4.556530)(2.800000,4.453080)(2.820000,4.483590)(2.840000,4.458320)(2.860000,4.466260)(2.880000,4.171850)(2.900000,4.553890)(2.920000,4.669040)(2.940000,4.401880)(2.960000,4.392920)(2.980000,4.205670)(3.000000,4.657860)(3.020000,4.350960)(3.040000,4.254880)(3.060000,4.624260)(3.080000,4.383450)(3.100000,4.463290)(3.120000,4.527180)(3.140000,4.348850)(3.160000,4.426450)(3.180000,4.487420)(3.200000,4.403400)(3.220000,4.551050)(3.240000,4.505620)(3.260000,4.453300)(3.280000,4.217760)(3.300000,4.409480)(3.320000,4.140680)(3.340000,4.271470)(3.360000,4.433940)(3.380000,4.567280)(3.400000,4.336640)(3.420000,4.497560)(3.440000,4.287420)(3.460000,4.617410)(3.480000,3.988450)(3.500000,4.432840)(3.520000,4.215450)(3.540000,4.139580)(3.560000,4.525720)(3.580000,4.174080)(3.600000,4.265600)(3.620000,4.385210)(3.640000,4.536900)(3.660000,4.517110)(3.680000,4.284360)(3.700000,4.366000)(3.720000,4.273530)(3.740000,4.408830)(3.760000,4.229090)(3.780000,4.187550)(3.800000,4.263440)(3.820000,4.110070)(3.840000,4.016650)(3.860000,4.492430)(3.880000,4.459140)(3.900000,4.182620)(3.920000,4.527320)(3.940000,4.280740)(3.960000,4.105350)(3.980000,4.339930)(4.000000,4.124260)(4.020000,4.182750)(4.040000,4.297150)(4.060000,4.297630)(4.080000,4.484740)(4.100000,4.220250)(4.120000,4.270060)(4.140000,4.115990)(4.160000,4.194580)(4.180000,4.282420)(4.200000,4.542050)(4.220000,4.164410)(4.240000,4.377040)(4.260000,4.356870)(4.280000,4.268540)(4.300000,4.418820)(4.320000,4.187820)(4.340000,4.375330)(4.360000,4.254090)(4.380000,4.195520)(4.400000,3.995950)(4.420000,4.255100)(4.440000,4.021000)(4.460000,4.353830)(4.480000,4.114570)(4.500000,4.099270)(4.520000,3.939900)(4.540000,4.241030)(4.560000,4.194470)(4.580000,4.289780)(4.600000,4.215370)(4.620000,4.196750)(4.640000,4.120750)(4.660000,4.051030)(4.680000,4.461840)(4.700000,4.276980)(4.720000,4.315760)(4.740000,4.211510)(4.760000,4.132550)(4.780000,4.229500)(4.800000,3.963660)(4.820000,4.157080)(4.840000,4.422610)(4.860000,3.833590)(4.880000,4.055930)(4.900000,3.936360)(4.920000,3.925830)(4.940000,4.135180)(4.960000,4.133060)(4.980000,3.928880)(5.000000,3.902890)(5.020000,4.099950)(5.040000,4.062950)(5.060000,4.006540)(5.080000,4.061650)(5.100000,4.274200)(5.120000,4.228680)(5.140000,4.084720)(5.160000,4.351410)(5.180000,3.967270)(5.200000,4.024370)(5.220000,4.007560)(5.240000,4.010910)(5.260000,4.105440)(5.280000,4.202450)(5.300000,4.186880)(5.320000,4.033600)(5.340000,4.052810)(5.360000,3.817720)(5.380000,4.056260)(5.400000,4.033260)(5.420000,4.163000)(5.440000,4.062270)(5.460000,4.228840)(5.480000,4.216020)(5.500000,3.943060)(5.520000,3.996020)(5.540000,4.043720)(5.560000,4.028720)(5.580000,3.994200)(5.600000,4.070080)(5.620000,4.169080)(5.640000,3.947600)(5.660000,4.089290)(5.680000,4.116240)(5.700000,4.207430)(5.720000,3.974600)(5.740000,3.998690)(5.760000,3.979080)(5.780000,3.816350)(5.800000,3.747500)(5.820000,3.950560)(5.840000,3.875470)(5.860000,3.860550)(5.880000,3.911830)(5.900000,3.792270)(5.920000,3.832230)(5.940000,4.223100)(5.960000,4.064360)(5.980000,3.991640)(6.000000,3.801250)(6.020000,3.645140)(6.040000,3.740430)(6.060000,4.054770)(6.080000,4.062710)(6.100000,3.916300)(6.120000,3.856240)(6.140000,3.999150)(6.160000,4.102770)(6.180000,3.703200)(6.200000,3.719460)(6.220000,3.765600)(6.240000,3.788630)(6.260000,3.914630)(6.280000,3.850700)(6.300000,4.127770)(6.320000,3.989120)(6.340000,3.950060)(6.360000,3.778190)(6.380000,4.058370)(6.400000,3.951580)(6.420000,4.089190)(6.440000,4.008460)(6.460000,3.549790)(6.480000,3.874030)(6.500000,3.610450)(6.520000,3.879120)(6.540000,3.839840)(6.560000,3.815190)(6.580000,3.909950)(6.600000,3.614350)(6.620000,3.881250)(6.640000,3.789830)(6.660000,3.869020)(6.680000,3.727890)(6.700000,3.892900)(6.720000,4.085290)(6.740000,3.871820)(6.760000,3.788820)(6.780000,3.912760)(6.800000,3.812960)(6.820000,3.785780)(6.840000,3.463750)(6.860000,3.909250)(6.880000,3.796450)(6.900000,3.910330)(6.920000,3.758130)(6.940000,3.910670)(6.960000,3.638520)(6.980000,3.872320)(7.000000,3.568210)(7.020000,3.591730)(7.040000,3.646370)(7.060000,3.836280)(7.080000,3.855250)(7.100000,3.592200)(7.120000,4.200070)(7.140000,3.636590)(7.160000,3.669360)(7.180000,3.543920)(7.200000,3.515320)(7.220000,4.093120)(7.240000,3.839840)(7.260000,3.654990)(7.280000,3.800490)(7.300000,3.980320)(7.320000,3.758960)(7.340000,3.576070)(7.360000,3.720910)(7.380000,3.747800)(7.400000,3.691470)(7.420000,3.710000)(7.440000,3.682920)(7.460000,3.963230)(7.480000,3.587440)(7.500000,3.918000)(7.520000,3.458270)(7.540000,3.906800)(7.560000,3.826860)(7.580000,3.974010)(7.600000,3.673850)(7.620000,3.993130)(7.640000,4.065140)(7.660000,3.838090)(7.680000,3.760550)(7.700000,3.718750)(7.720000,3.723650)(7.740000,3.674530)(7.760000,3.707760)(7.780000,3.612000)(7.800000,3.671110)(7.820000,3.852170)(7.840000,3.815550)(7.860000,3.644550)(7.880000,3.819450)(7.900000,3.780820)(7.920000,3.726030)(7.940000,3.830110)(7.960000,3.746370)(7.980000,3.836230)(8.000000,3.965800)(8.020000,3.733710)(8.040000,3.717470)(8.060000,3.868280)(8.080000,3.715120)(8.100000,3.508430)(8.120000,3.740880)(8.140000,3.574650)(8.160000,3.495600)(8.180000,3.919850)(8.200000,3.609200)(8.220000,3.810000)(8.240000,3.575970)(8.260000,3.587690)(8.280000,3.619440)(8.300000,3.573970)(8.320000,3.410110)(8.340000,3.618470)(8.360000,3.256230)(8.380000,3.630030)(8.400000,3.680100)(8.420000,3.624840)(8.440000,3.681740)(8.460000,3.748820)(8.480000,3.723810)(8.500000,3.660050)(8.520000,3.698080)(8.540000,3.850470)(8.560000,3.656610)(8.580000,3.645820)(8.600000,3.796970)(8.620000,3.825750)(8.640000,3.562680)(8.660000,3.804730)(8.680000,3.548160)(8.700000,3.634080)(8.720000,3.531810)(8.740000,3.597330)(8.760000,3.575830)(8.780000,3.582730)(8.800000,3.759230)(8.820000,3.911250)(8.840000,3.357430)(8.860000,3.673490)(8.880000,3.550810)(8.900000,3.451170)(8.920000,3.904310)(8.940000,3.580490)(8.960000,3.741650)(8.980000,3.687260)(9.000000,3.446440)(9.020000,3.405680)(9.040000,3.391500)(9.060000,3.471350)(9.080000,3.609090)(9.100000,3.475220)(9.120000,3.447520)(9.140000,3.573100)(9.160000,3.600100)(9.180000,3.733770)(9.200000,3.569240)(9.220000,3.533260)(9.240000,3.541800)(9.260000,3.802570)(9.280000,3.802750)(9.300000,3.441660)(9.320000,3.340920)(9.340000,3.549240)(9.360000,3.472800)(9.380000,3.417550)(9.400000,3.484160)(9.420000,3.751390)(9.440000,3.690850)(9.460000,3.644790)(9.480000,3.364210)(9.500000,3.609840)(9.520000,3.643810)(9.540000,3.776260)(9.560000,3.366870)(9.580000,3.587560)(9.600000,3.535470)(9.620000,3.568440)(9.640000,3.520060)(9.660000,3.383940)(9.680000,3.265340)(9.700000,3.370660)(9.720000,3.289480)(9.740000,3.406190)(9.760000,3.573640)(9.780000,3.470710)(9.800000,3.760380)(9.820000,3.692650)(9.840000,3.484210)(9.860000,3.615970)(9.880000,3.480680)(9.900000,3.533490)(9.920000,3.543550)(9.940000,3.479530)(9.960000,3.803510)(9.980000,3.820080)}; 
\end{axis}
\end{tikzpicture}

%% file: figures/imagenet/caffenet_val_loss_10k.tex
\begin{tikzpicture}[scale=0.7]
\begin{axis}[
  legend cell align=left,
  height=0.95\textwidth,
  width=1.5\textwidth,
  xmin=0, xmax=10.0,
  ymax=7,
  legend pos=north east,
  scaled x ticks = false,
  xtick = {0, 2, 4, 6, 8, 10},
  ytick = {3, 4, 5, 6, 7},
  xticklabel = {\pgfmathprintnumber{\tick}K}
]

\addplot[color=ta3gray,ultra thick,dashed] coordinates {(0.000000,7.127340)(1.980000,6.894290)(3.980000,6.329610)(5.980000,5.021850)(7.980000,4.324910)(9.980000,3.929770)}; 
\addplot[color=taskyblue,ultra thick,dashed] coordinates {(0.000000,6.907720)(1.980000,6.514940)(3.980000,5.364330)(5.980000,4.782840)(7.980000,4.352270)(9.980000,4.073630)}; 
\addplot[color=ta3scarletred,ultra thick,solid] coordinates {(0.000000,6.907720)(1.980000,5.084470)(3.980000,4.488700)(5.980000,4.089560)(7.980000,3.782350)(9.980000,3.592700)}; 
\addplot[color=ta3orange,ultra thick,solid] coordinates {(0.000000,6.907720)(1.980000,4.772280)(3.980000,4.238680)(5.980000,3.896280)(7.980000,3.666120)(9.980000,3.500990)}; 
\addplot[color=ta3chameleon,ultra thick,solid] coordinates {(0.000000,6.907720)(1.980000,4.724270)(3.980000,4.146360)(5.980000,3.834350)(7.980000,3.556910)(9.980000,3.408780)}; 
\end{axis}
\end{tikzpicture}

%% file: figures/imagenet/caffenet_val_loss_100k.tex
\begin{tikzpicture}[scale=0.7]
\begin{axis}[
  legend cell align=left,
  height=0.95\textwidth,
  width=1.5\textwidth,
  xmin=0, xmax=100.0,
  ymax=7,
  legend pos=north east,
  scaled x ticks = false,
  xtick = {0, 20, 40, 60, 80, 100},
  ytick = {2, 3, 4, 5, 6, 7},
  xticklabel = {\pgfmathprintnumber{\tick}K}
]

\addplot[color=ta3gray,ultra thick,dashed] coordinates {(0.000000,7.127340)(1.980000,6.894290)(3.980000,6.329610)(5.980000,5.021850)(7.980000,4.324910)(9.980000,3.929770)(11.980000,3.718170)(13.980000,3.551870)(15.980000,3.439560)(17.980000,3.294330)(19.980000,3.276960)(21.980000,3.256120)(23.980000,3.142680)(25.980000,3.143960)(27.980000,3.066930)(29.980000,3.048090)(31.980000,3.034460)(33.980000,2.973800)(35.980000,2.979530)(37.980000,2.944740)(39.980000,2.968630)(41.980000,2.930090)(43.980000,2.893680)(45.980000,2.924000)(47.980000,2.882600)(49.980000,2.860490)(51.980000,2.885970)(53.980000,2.810390)(55.980000,2.853380)(57.980000,2.855330)(59.980000,2.824880)(61.980000,2.915170)(63.980000,2.817940)(65.980000,2.809640)(67.980000,2.766350)(69.980000,2.781450)(71.980000,2.799610)(73.980000,2.747600)(75.980000,2.764440)(77.980000,2.784950)(79.980000,2.759300)(81.980000,2.756910)(83.980000,2.764810)(85.980000,2.772160)(87.980000,2.769310)(89.980000,2.778090)(91.980000,2.688350)(93.980000,2.766760)(95.980000,2.752070)(97.980000,2.723780)(99.980000,2.687040)}; 
\addplot[color=taskyblue,ultra thick,dashed] coordinates {(0.000000,6.907720)(1.980000,6.514940)(3.980000,5.364330)(5.980000,4.782840)(7.980000,4.352270)(9.980000,4.073630)(11.980000,3.867110)(13.980000,3.662120)(15.980000,3.551380)(17.980000,3.413480)(19.980000,3.310410)(21.980000,3.236560)(23.980000,3.176520)(25.980000,3.125000)(27.980000,3.041650)(29.980000,3.004150)(31.980000,2.975340)(33.980000,2.906900)(35.980000,2.880040)(37.980000,2.859960)(39.980000,2.831580)(41.980000,2.809260)(43.980000,2.787650)(45.980000,2.755500)(47.980000,2.729630)(49.980000,2.708980)(51.980000,2.718910)(53.980000,2.641860)(55.980000,2.664150)(57.980000,2.661750)(59.980000,2.643530)(61.980000,2.659730)(63.980000,2.611590)(65.980000,2.610850)(67.980000,2.604890)(69.980000,2.570470)(71.980000,2.597230)(73.980000,2.560310)(75.980000,2.563690)(77.980000,2.541330)(79.980000,2.541860)(81.980000,2.589730)(83.980000,2.562640)(85.980000,2.523240)(87.980000,2.501290)(89.980000,2.524270)(91.980000,2.467800)(93.980000,2.501890)(95.980000,2.509550)(97.980000,2.503010)(99.980000,2.476170)}; 
\addplot[color=ta3scarletred,ultra thick,solid] coordinates {(0.000000,6.907720)(1.980000,5.084470)(3.980000,4.488700)(5.980000,4.089560)(7.980000,3.782350)(9.980000,3.592700)(11.980000,3.453540)(13.980000,3.323030)(15.980000,3.245140)(17.980000,3.160880)(19.980000,3.101280)(21.980000,2.990060)(23.980000,2.984600)(25.980000,2.950570)(27.980000,2.934110)(29.980000,2.826410)(31.980000,2.855490)(33.980000,2.747400)(35.980000,2.735930)(37.980000,2.715780)(39.980000,2.729740)(41.980000,2.673630)(43.980000,2.659050)(45.980000,2.711940)(47.980000,2.615060)(49.980000,2.639340)(51.980000,2.606040)(53.980000,2.608990)(55.980000,2.598350)(57.980000,2.626660)(59.980000,2.571460)(61.980000,2.563390)(63.980000,2.591580)(65.980000,2.577600)(67.980000,2.543240)(69.980000,2.540510)(71.980000,2.551380)(73.980000,2.517250)(75.980000,2.547010)(77.980000,2.511080)(79.980000,2.502880)(81.980000,2.508900)(83.980000,2.520810)(85.980000,2.497000)(87.980000,2.525100)(89.980000,2.462530)(91.980000,2.482850)(93.980000,2.487050)(95.980000,2.473160)(97.980000,2.485570)(99.980000,2.451840)}; 
\addplot[color=ta3orange,ultra thick,solid] coordinates {(0.000000,6.907720)(1.980000,4.772280)(3.980000,4.238680)(5.980000,3.896280)(7.980000,3.666120)(9.980000,3.500990)(11.980000,3.392090)(13.980000,3.303110)(15.980000,3.169220)(17.980000,3.065780)(19.980000,3.042280)(21.980000,2.962430)(23.980000,2.900240)(25.980000,2.900080)(27.980000,2.849690)(29.980000,2.781420)(31.980000,2.800800)(33.980000,2.745510)(35.980000,2.734680)(37.980000,2.696880)(39.980000,2.683910)(41.980000,2.670350)(43.980000,2.654220)(45.980000,2.637490)(47.980000,2.626770)(49.980000,2.665740)(51.980000,2.600540)(53.980000,2.606890)(55.980000,2.592250)(57.980000,2.566440)(59.980000,2.569710)(61.980000,2.561400)(63.980000,2.516350)(65.980000,2.545520)(67.980000,2.510470)(69.980000,2.521750)(71.980000,2.532290)(73.980000,2.514040)(75.980000,2.518520)(77.980000,2.534910)(79.980000,2.494140)(81.980000,2.468790)(83.980000,2.505290)(85.980000,2.455660)(87.980000,2.469580)(89.980000,2.474480)(91.980000,2.461890)(93.980000,2.473310)(95.980000,2.431450)(97.980000,2.487250)(99.980000,2.443090)}; 
\addplot[color=ta3chameleon,ultra thick,solid] coordinates {(0.000000,6.907720)(1.980000,4.724270)(3.980000,4.146360)(5.980000,3.834350)(7.980000,3.556910)(9.980000,3.408780)(11.980000,3.303110)(13.980000,3.227990)(15.980000,3.121070)(17.980000,3.028460)(19.980000,3.006640)(21.980000,2.921760)(23.980000,2.917500)(25.980000,2.884920)(27.980000,2.838100)(29.980000,2.792470)(31.980000,2.757510)(33.980000,2.713940)(35.980000,2.711440)(37.980000,2.687400)(39.980000,2.709600)(41.980000,2.657280)(43.980000,2.660320)(45.980000,2.661490)(47.980000,2.636570)(49.980000,2.608850)(51.980000,2.581030)(53.980000,2.571900)(55.980000,2.626790)(57.980000,2.607740)(59.980000,2.565820)(61.980000,2.570260)(63.980000,2.567280)(65.980000,2.562860)(67.980000,2.534390)(69.980000,2.521920)(71.980000,2.533350)(73.980000,2.548900)(75.980000,2.514500)(77.980000,2.518160)(79.980000,2.496970)(81.980000,2.478420)(83.980000,2.508480)(85.980000,2.451680)(87.980000,2.506040)(89.980000,2.470450)(91.980000,2.500320)(93.980000,2.429370)(95.980000,2.459880)(97.980000,2.458800)(99.980000,2.457260)}; 
\end{axis}
\end{tikzpicture}

%% file: figures/imagenet/caffenet_val_loss.tex
\begin{tikzpicture}[scale=0.7]
\begin{axis}[
  legend cell align=left,
  height=0.95\textwidth,
  width=1.5\textwidth,
  xmin=0, xmax=350.0,
  ymax=7,
  legend pos=north east,
  scaled x ticks = false,
  xtick = {0, 50, 100, 150, 200, 250, 300, 350},
  ytick = {1, 2, 3, 4, 5, 6, 7},
  xticklabel = {\pgfmathprintnumber{\tick}K}
]

\addplot[color=ta3gray,ultra thick,dashed] coordinates {(0.000000,7.127340)(1.980000,6.894290)(3.980000,6.329610)(5.980000,5.021850)(7.980000,4.324910)(9.980000,3.929770)(11.980000,3.718170)(13.980000,3.551870)(15.980000,3.439560)(17.980000,3.294330)(19.980000,3.276960)(21.980000,3.256120)(23.980000,3.142680)(25.980000,3.143960)(27.980000,3.066930)(29.980000,3.048090)(31.980000,3.034460)(33.980000,2.973800)(35.980000,2.979530)(37.980000,2.944740)(39.980000,2.968630)(41.980000,2.930090)(43.980000,2.893680)(45.980000,2.924000)(47.980000,2.882600)(49.980000,2.860490)(51.980000,2.885970)(53.980000,2.810390)(55.980000,2.853380)(57.980000,2.855330)(59.980000,2.824880)(61.980000,2.915170)(63.980000,2.817940)(65.980000,2.809640)(67.980000,2.766350)(69.980000,2.781450)(71.980000,2.799610)(73.980000,2.747600)(75.980000,2.764440)(77.980000,2.784950)(79.980000,2.759300)(81.980000,2.756910)(83.980000,2.764810)(85.980000,2.772160)(87.980000,2.769310)(89.980000,2.778090)(91.980000,2.688350)(93.980000,2.766760)(95.980000,2.752070)(97.980000,2.723780)(99.980000,2.687040)(101.980000,2.321640)(103.980000,2.274680)(105.980000,2.234130)(107.980000,2.202650)(109.980000,2.183050)(111.980000,2.166570)(113.980000,2.149260)(115.980000,2.137530)(117.980000,2.118860)(119.980000,2.107610)(121.980000,2.100170)(123.980000,2.089020)(125.980000,2.083140)(127.980000,2.069280)(129.980000,2.062440)(131.980000,2.054890)(133.980000,2.040330)(135.980000,2.034220)(137.980000,2.030370)(139.980000,2.024010)(141.980000,2.034250)(143.980000,2.012030)(145.980000,2.009770)(147.980000,2.010270)(149.980000,2.007750)(151.980000,1.993830)(153.980000,1.996940)(155.980000,1.983800)(157.980000,1.990610)(159.980000,1.983030)(161.980000,1.988200)(163.980000,1.977220)(165.980000,1.978340)(167.980000,1.973370)(169.980000,1.975300)(171.980000,1.968900)(173.980000,1.974750)(175.980000,1.956860)(177.980000,1.962020)(179.980000,1.955950)(181.980000,1.956820)(183.980000,1.960640)(185.980000,1.955140)(187.980000,1.955590)(189.980000,1.948760)(191.980000,1.943390)(193.980000,1.965690)(195.980000,1.951950)(197.980000,1.952100)(199.980000,1.957120)(201.980000,1.882490)(203.980000,1.883900)(205.980000,1.880420)(207.980000,1.879110)(209.980000,1.878190)(211.980000,1.872810)(213.980000,1.874670)(215.980000,1.869860)(217.980000,1.870910)(219.980000,1.871720)(221.980000,1.868880)(223.980000,1.871300)(225.980000,1.862300)(227.980000,1.866150)(229.980000,1.868070)(231.980000,1.863580)(233.980000,1.863290)(235.980000,1.860060)(237.980000,1.865350)(239.980000,1.864950)(241.980000,1.861310)(243.980000,1.863500)(245.980000,1.856920)(247.980000,1.861570)(249.980000,1.863040)(251.980000,1.857190)(253.980000,1.859940)(255.980000,1.857000)(257.980000,1.862100)(259.980000,1.858820)(261.980000,1.858110)(263.980000,1.859870)(265.980000,1.855340)(267.980000,1.857700)(269.980000,1.856550)(271.980000,1.856460)(273.980000,1.858350)(275.980000,1.853610)(277.980000,1.853450)(279.980000,1.854120)(281.980000,1.854690)(283.980000,1.855340)(285.980000,1.847800)(287.980000,1.849680)(289.980000,1.852090)(291.980000,1.849690)(293.980000,1.850470)(295.980000,1.848260)(297.980000,1.850430)(299.980000,1.851190)(301.980000,1.842390)(303.980000,1.842130)(305.980000,1.840580)(307.980000,1.841400)(309.980000,1.841600)(311.980000,1.840040)(313.980000,1.841400)(315.980000,1.839770)(317.980000,1.840250)(319.980000,1.841380)(321.980000,1.839610)(323.980000,1.841130)(325.980000,1.839790)(327.980000,1.839840)(329.980000,1.840400)(331.980000,1.839470)(333.980000,1.841070)(335.980000,1.840310)(337.980000,1.840830)(339.980000,1.840000)(341.980000,1.839780)(343.980000,1.840450)(345.980000,1.839220)(347.980000,1.839380)(349.980000,1.839930)(351.980000,1.838890)(353.980000,1.840400)(355.980000,1.839610)(357.980000,1.839250)(359.980000,1.840140)}; 
\addlegendentry{Reference}
\addplot[color=taskyblue,ultra thick,dashed] coordinates {(0.000000,6.907720)(1.980000,6.514940)(3.980000,5.364330)(5.980000,4.782840)(7.980000,4.352270)(9.980000,4.073630)(11.980000,3.867110)(13.980000,3.662120)(15.980000,3.551380)(17.980000,3.413480)(19.980000,3.310410)(21.980000,3.236560)(23.980000,3.176520)(25.980000,3.125000)(27.980000,3.041650)(29.980000,3.004150)(31.980000,2.975340)(33.980000,2.906900)(35.980000,2.880040)(37.980000,2.859960)(39.980000,2.831580)(41.980000,2.809260)(43.980000,2.787650)(45.980000,2.755500)(47.980000,2.729630)(49.980000,2.708980)(51.980000,2.718910)(53.980000,2.641860)(55.980000,2.664150)(57.980000,2.661750)(59.980000,2.643530)(61.980000,2.659730)(63.980000,2.611590)(65.980000,2.610850)(67.980000,2.604890)(69.980000,2.570470)(71.980000,2.597230)(73.980000,2.560310)(75.980000,2.563690)(77.980000,2.541330)(79.980000,2.541860)(81.980000,2.589730)(83.980000,2.562640)(85.980000,2.523240)(87.980000,2.501290)(89.980000,2.524270)(91.980000,2.467800)(93.980000,2.501890)(95.980000,2.509550)(97.980000,2.503010)(99.980000,2.476170)(101.980000,2.182880)(103.980000,2.154600)(105.980000,2.134340)(107.980000,2.116690)(109.980000,2.107840)(111.980000,2.088720)(113.980000,2.087880)(115.980000,2.074840)(117.980000,2.063780)(119.980000,2.072970)(121.980000,2.060910)(123.980000,2.060300)(125.980000,2.046160)(127.980000,2.043700)(129.980000,2.058480)(131.980000,2.043440)(133.980000,2.038710)(135.980000,2.030130)(137.980000,2.023120)(139.980000,2.040440)(141.980000,2.023740)(143.980000,2.013400)(145.980000,2.016690)(147.980000,2.024720)(149.980000,2.027000)(151.980000,2.020090)(153.980000,2.017760)(155.980000,2.016000)(157.980000,2.007980)(159.980000,2.022490)(161.980000,2.024170)(163.980000,2.002060)(165.980000,1.995510)(167.980000,2.008470)(169.980000,2.012300)(171.980000,2.009930)(173.980000,2.009480)(175.980000,2.011810)(177.980000,2.001920)(179.980000,2.003390)(181.980000,1.996380)(183.980000,2.015100)(185.980000,2.003970)(187.980000,1.999430)(189.980000,1.998430)(191.980000,1.996140)(193.980000,2.011190)(195.980000,2.005930)(197.980000,2.000920)(199.980000,2.004190)(201.980000,1.945410)(203.980000,1.943250)(205.980000,1.939350)(207.980000,1.938400)(209.980000,1.940770)(211.980000,1.935500)(213.980000,1.935610)(215.980000,1.932010)(217.980000,1.935280)(219.980000,1.937090)(221.980000,1.933280)(223.980000,1.933360)(225.980000,1.930590)(227.980000,1.933730)(229.980000,1.936380)(231.980000,1.931160)(233.980000,1.931410)(235.980000,1.928650)(237.980000,1.931740)(239.980000,1.932740)(241.980000,1.928560)(243.980000,1.933240)(245.980000,1.927660)(247.980000,1.930250)(249.980000,1.935060)(251.980000,1.926830)(253.980000,1.930610)(255.980000,1.927220)(257.980000,1.931290)(259.980000,1.930560)(261.980000,1.927960)(263.980000,1.929550)(265.980000,1.927130)(267.980000,1.929580)(269.980000,1.930840)(271.980000,1.929830)(273.980000,1.925120)(275.980000,1.924690)(277.980000,1.924380)(279.980000,1.932000)(281.980000,1.926800)(283.980000,1.928210)(285.980000,1.919970)(287.980000,1.925360)(289.980000,1.929080)(291.980000,1.924030)(293.980000,1.927790)(295.980000,1.919960)(297.980000,1.922330)(299.980000,1.930050)(301.980000,1.919690)(303.980000,1.920020)(305.980000,1.917780)(307.980000,1.919880)(309.980000,1.920590)(311.980000,1.917980)(313.980000,1.918760)(315.980000,1.917560)(317.980000,1.918660)(319.980000,1.919770)(321.980000,1.917670)(323.980000,1.918130)(325.980000,1.916770)(327.980000,1.918290)(329.980000,1.919910)(331.980000,1.917840)(333.980000,1.918320)(335.980000,1.916920)(337.980000,1.918100)(339.980000,1.919200)(341.980000,1.917690)(343.980000,1.918470)(345.980000,1.917440)(347.980000,1.917740)(349.980000,1.918570)(351.980000,1.917330)(353.980000,1.917850)(355.980000,1.916770)(357.980000,1.917450)(359.980000,1.918250)(361.980000,1.917400)(363.980000,1.917770)(365.980000,1.916760)(367.980000,1.917600)(369.980000,1.918680)(371.980000,1.917990)(373.980000,1.917880)(375.980000,1.916740)(377.980000,1.916610)(379.980000,1.918440)(381.980000,1.916610)(383.980000,1.918150)}; 
\addlegendentry{MSRA}
\addplot[color=ta3scarletred,ultra thick,solid] coordinates {(0.000000,6.907720)(1.980000,5.084470)(3.980000,4.488700)(5.980000,4.089560)(7.980000,3.782350)(9.980000,3.592700)(11.980000,3.453540)(13.980000,3.323030)(15.980000,3.245140)(17.980000,3.160880)(19.980000,3.101280)(21.980000,2.990060)(23.980000,2.984600)(25.980000,2.950570)(27.980000,2.934110)(29.980000,2.826410)(31.980000,2.855490)(33.980000,2.747400)(35.980000,2.735930)(37.980000,2.715780)(39.980000,2.729740)(41.980000,2.673630)(43.980000,2.659050)(45.980000,2.711940)(47.980000,2.615060)(49.980000,2.639340)(51.980000,2.606040)(53.980000,2.608990)(55.980000,2.598350)(57.980000,2.626660)(59.980000,2.571460)(61.980000,2.563390)(63.980000,2.591580)(65.980000,2.577600)(67.980000,2.543240)(69.980000,2.540510)(71.980000,2.551380)(73.980000,2.517250)(75.980000,2.547010)(77.980000,2.511080)(79.980000,2.502880)(81.980000,2.508900)(83.980000,2.520810)(85.980000,2.497000)(87.980000,2.525100)(89.980000,2.462530)(91.980000,2.482850)(93.980000,2.487050)(95.980000,2.473160)(97.980000,2.485570)(99.980000,2.451840)(101.980000,2.173180)(103.980000,2.150160)(105.980000,2.128810)(107.980000,2.107220)(109.980000,2.092870)(111.980000,2.077560)(113.980000,2.080270)(115.980000,2.065050)(117.980000,2.056940)(119.980000,2.055910)(121.980000,2.047710)(123.980000,2.047950)(125.980000,2.039590)(127.980000,2.029770)(129.980000,2.063290)(131.980000,2.031620)(133.980000,2.033840)(135.980000,2.026940)(137.980000,2.016680)(139.980000,2.040240)(141.980000,2.022070)(143.980000,2.004860)(145.980000,2.010990)(147.980000,2.012810)(149.980000,2.009990)(151.980000,2.002860)(153.980000,2.013320)(155.980000,2.006600)(157.980000,2.009210)(159.980000,1.993950)(161.980000,2.009650)(163.980000,2.001200)(165.980000,2.004120)(167.980000,2.001700)(169.980000,1.995350)(171.980000,1.990180)(173.980000,1.997670)(175.980000,2.003060)(177.980000,1.997260)(179.980000,1.993580)(181.980000,1.989590)(183.980000,1.991450)(185.980000,2.005680)(187.980000,1.990320)(189.980000,1.992720)(191.980000,1.993120)(193.980000,1.999700)(195.980000,1.995790)(197.980000,1.999020)(199.980000,2.003490)(201.980000,1.937420)(203.980000,1.936300)(205.980000,1.931610)(207.980000,1.929600)(209.980000,1.936120)(211.980000,1.927750)(213.980000,1.931250)(215.980000,1.924630)(217.980000,1.925820)(219.980000,1.931910)(221.980000,1.923920)(223.980000,1.928710)(225.980000,1.921130)(227.980000,1.926050)(229.980000,1.928190)(231.980000,1.921700)(233.980000,1.924640)(235.980000,1.921960)(237.980000,1.923340)(239.980000,1.927080)(241.980000,1.922390)(243.980000,1.923750)(245.980000,1.919780)(247.980000,1.922370)(249.980000,1.924190)(251.980000,1.921610)(253.980000,1.926070)(255.980000,1.918270)(257.980000,1.924420)(259.980000,1.923450)(261.980000,1.920560)(263.980000,1.924100)(265.980000,1.919210)(267.980000,1.921180)(269.980000,1.920850)(271.980000,1.920790)(273.980000,1.920220)(275.980000,1.917960)(277.980000,1.918490)(279.980000,1.924610)(281.980000,1.920410)(283.980000,1.925110)(285.980000,1.915110)(287.980000,1.917060)(289.980000,1.919670)(291.980000,1.919560)(293.980000,1.919980)(295.980000,1.912340)(297.980000,1.917420)(299.980000,1.919140)(301.980000,1.908870)(303.980000,1.910780)(305.980000,1.908150)(307.980000,1.909750)(309.980000,1.911720)(311.980000,1.908370)(313.980000,1.910660)(315.980000,1.907720)(317.980000,1.909550)(319.980000,1.911750)(321.980000,1.908280)(323.980000,1.910200)(325.980000,1.907590)(327.980000,1.909570)(329.980000,1.910360)(331.980000,1.907100)(333.980000,1.909840)(335.980000,1.907590)(337.980000,1.909620)(339.980000,1.910310)(341.980000,1.908380)(343.980000,1.910160)(345.980000,1.906900)(347.980000,1.908970)(349.980000,1.910680)(351.980000,1.908300)(353.980000,1.909750)(355.980000,1.907970)(357.980000,1.909230)(359.980000,1.909690)(361.980000,1.908210)(363.980000,1.909610)(365.980000,1.907740)(367.980000,1.909000)(369.980000,1.909510)(371.980000,1.907670)}; 
\addlegendentry{Random (ours)}
\addplot[color=ta3orange,ultra thick,solid] coordinates {(0.000000,6.907720)(1.980000,4.772280)(3.980000,4.238680)(5.980000,3.896280)(7.980000,3.666120)(9.980000,3.500990)(11.980000,3.392090)(13.980000,3.303110)(15.980000,3.169220)(17.980000,3.065780)(19.980000,3.042280)(21.980000,2.962430)(23.980000,2.900240)(25.980000,2.900080)(27.980000,2.849690)(29.980000,2.781420)(31.980000,2.800800)(33.980000,2.745510)(35.980000,2.734680)(37.980000,2.696880)(39.980000,2.683910)(41.980000,2.670350)(43.980000,2.654220)(45.980000,2.637490)(47.980000,2.626770)(49.980000,2.665740)(51.980000,2.600540)(53.980000,2.606890)(55.980000,2.592250)(57.980000,2.566440)(59.980000,2.569710)(61.980000,2.561400)(63.980000,2.516350)(65.980000,2.545520)(67.980000,2.510470)(69.980000,2.521750)(71.980000,2.532290)(73.980000,2.514040)(75.980000,2.518520)(77.980000,2.534910)(79.980000,2.494140)(81.980000,2.468790)(83.980000,2.505290)(85.980000,2.455660)(87.980000,2.469580)(89.980000,2.474480)(91.980000,2.461890)(93.980000,2.473310)(95.980000,2.431450)(97.980000,2.487250)(99.980000,2.443090)(101.980000,2.146000)(103.980000,2.113890)(105.980000,2.094630)(107.980000,2.078050)(109.980000,2.068880)(111.980000,2.051380)(113.980000,2.052590)(115.980000,2.039760)(117.980000,2.030570)(119.980000,2.036730)(121.980000,2.025690)(123.980000,2.029070)(125.980000,2.019030)(127.980000,2.008950)(129.980000,2.028250)(131.980000,2.005410)(133.980000,2.006130)(135.980000,1.994410)(137.980000,1.996230)(139.980000,2.021140)(141.980000,1.992770)(143.980000,1.988020)(145.980000,1.984340)(147.980000,1.992210)(149.980000,1.995720)(151.980000,1.987110)(153.980000,1.988830)(155.980000,1.981710)(157.980000,1.979940)(159.980000,1.974000)(161.980000,1.982270)(163.980000,1.984890)(165.980000,1.976070)(167.980000,1.972650)(169.980000,1.979880)(171.980000,1.971680)(173.980000,1.988260)(175.980000,1.982850)(177.980000,1.978960)(179.980000,1.976650)(181.980000,1.973170)(183.980000,1.984900)(185.980000,1.975010)(187.980000,1.971340)(189.980000,1.966880)(191.980000,1.965300)(193.980000,1.977530)(195.980000,1.967380)(197.980000,1.985040)(199.980000,1.976520)(201.980000,1.919000)(203.980000,1.916360)(205.980000,1.912570)(207.980000,1.909580)(209.980000,1.914640)(211.980000,1.908030)(213.980000,1.912810)(215.980000,1.904750)(217.980000,1.906810)(219.980000,1.910900)(221.980000,1.907970)(223.980000,1.907970)(225.980000,1.903080)(227.980000,1.907260)(229.980000,1.907410)(231.980000,1.903700)(233.980000,1.906340)(235.980000,1.901400)(237.980000,1.903520)(239.980000,1.905700)(241.980000,1.903020)(243.980000,1.908200)(245.980000,1.899770)(247.980000,1.900720)(249.980000,1.905770)(251.980000,1.900400)(253.980000,1.906080)(255.980000,1.898430)(257.980000,1.901730)(259.980000,1.903570)(261.980000,1.902200)(263.980000,1.902590)(265.980000,1.901640)(267.980000,1.900730)(269.980000,1.900660)(271.980000,1.902960)(273.980000,1.898780)(275.980000,1.896890)(277.980000,1.895400)(279.980000,1.903380)(281.980000,1.900320)(283.980000,1.902890)(285.980000,1.893930)(287.980000,1.896160)(289.980000,1.899180)(291.980000,1.898810)(293.980000,1.902230)(295.980000,1.892750)(297.980000,1.895450)(299.980000,1.899680)(301.980000,1.891040)(303.980000,1.892030)(305.980000,1.889360)(307.980000,1.890990)(309.980000,1.892780)(311.980000,1.889030)(313.980000,1.890850)(315.980000,1.887750)(317.980000,1.890050)(319.980000,1.892520)(321.980000,1.889140)(323.980000,1.890430)(325.980000,1.888460)(327.980000,1.889850)(329.980000,1.892060)(331.980000,1.888890)(333.980000,1.890610)(335.980000,1.888170)(337.980000,1.890360)(339.980000,1.891290)(341.980000,1.889010)(343.980000,1.890310)(345.980000,1.887330)(347.980000,1.889760)(349.980000,1.891460)(351.980000,1.888660)(353.980000,1.890760)(355.980000,1.888710)(357.980000,1.889120)(359.980000,1.891300)(361.980000,1.888800)(363.980000,1.889970)(365.980000,1.888080)(367.980000,1.889590)(369.980000,1.890620)(371.980000,1.888170)}; 
\addlegendentry{\kmeans (ours)}
\addplot[color=ta3chameleon,ultra thick,solid] coordinates {(0.000000,6.907720)(1.980000,4.724270)(3.980000,4.146360)(5.980000,3.834350)(7.980000,3.556910)(9.980000,3.408780)(11.980000,3.303110)(13.980000,3.227990)(15.980000,3.121070)(17.980000,3.028460)(19.980000,3.006640)(21.980000,2.921760)(23.980000,2.917500)(25.980000,2.884920)(27.980000,2.838100)(29.980000,2.792470)(31.980000,2.757510)(33.980000,2.713940)(35.980000,2.711440)(37.980000,2.687400)(39.980000,2.709600)(41.980000,2.657280)(43.980000,2.660320)(45.980000,2.661490)(47.980000,2.636570)(49.980000,2.608850)(51.980000,2.581030)(53.980000,2.571900)(55.980000,2.626790)(57.980000,2.607740)(59.980000,2.565820)(61.980000,2.570260)(63.980000,2.567280)(65.980000,2.562860)(67.980000,2.534390)(69.980000,2.521920)(71.980000,2.533350)(73.980000,2.548900)(75.980000,2.514500)(77.980000,2.518160)(79.980000,2.496970)(81.980000,2.478420)(83.980000,2.508480)(85.980000,2.451680)(87.980000,2.506040)(89.980000,2.470450)(91.980000,2.500320)(93.980000,2.429370)(95.980000,2.459880)(97.980000,2.458800)(99.980000,2.457260)(101.980000,2.149430)(103.980000,2.117120)(105.980000,2.095510)(107.980000,2.079160)(109.980000,2.066310)(111.980000,2.048400)(113.980000,2.049570)(115.980000,2.037190)(117.980000,2.035780)(119.980000,2.037290)(121.980000,2.020330)(123.980000,2.028830)(125.980000,2.018180)(127.980000,2.013380)(129.980000,2.028770)(131.980000,2.008000)(133.980000,2.012980)(135.980000,1.994220)(137.980000,1.997610)(139.980000,2.013570)(141.980000,1.988530)(143.980000,1.994440)(145.980000,1.985600)(147.980000,1.990980)(149.980000,1.994480)(151.980000,1.990690)(153.980000,1.982450)(155.980000,1.981660)(157.980000,1.980380)(159.980000,1.986340)(161.980000,1.981940)(163.980000,1.984330)(165.980000,1.978650)(167.980000,1.976630)(169.980000,1.984770)(171.980000,1.970520)(173.980000,1.982970)(175.980000,1.971530)(177.980000,1.970260)(179.980000,1.986150)(181.980000,1.973040)(183.980000,1.966980)(185.980000,1.980640)(187.980000,1.969950)(189.980000,1.977970)(191.980000,1.962660)(193.980000,1.991370)(195.980000,1.976620)(197.980000,1.977050)(199.980000,1.980580)(201.980000,1.917520)(203.980000,1.915660)(205.980000,1.908850)(207.980000,1.910190)(209.980000,1.911010)(211.980000,1.907820)(213.980000,1.910040)(215.980000,1.903960)(217.980000,1.904450)(219.980000,1.910660)(221.980000,1.905360)(223.980000,1.907140)(225.980000,1.901190)(227.980000,1.905380)(229.980000,1.905560)(231.980000,1.903380)(233.980000,1.906200)(235.980000,1.901160)(237.980000,1.902530)(239.980000,1.903480)(241.980000,1.901370)(243.980000,1.906370)(245.980000,1.902590)(247.980000,1.900210)(249.980000,1.902560)(251.980000,1.900090)(253.980000,1.900820)(255.980000,1.899870)(257.980000,1.902030)(259.980000,1.903610)(261.980000,1.899180)(263.980000,1.902750)(265.980000,1.899230)(267.980000,1.901280)(269.980000,1.902400)(271.980000,1.901850)(273.980000,1.899440)(275.980000,1.894520)(277.980000,1.897570)(279.980000,1.904370)(281.980000,1.898960)(283.980000,1.899390)(285.980000,1.891390)(287.980000,1.897380)(289.980000,1.903370)(291.980000,1.898860)(293.980000,1.898520)(295.980000,1.890950)(297.980000,1.897180)(299.980000,1.901830)(301.980000,1.889470)(303.980000,1.891040)(305.980000,1.888070)(307.980000,1.890390)(309.980000,1.892600)(311.980000,1.888370)(313.980000,1.890580)(315.980000,1.887860)(317.980000,1.890060)(319.980000,1.891320)(321.980000,1.888290)(323.980000,1.890150)(325.980000,1.887770)(327.980000,1.889720)(329.980000,1.891240)(331.980000,1.887500)(333.980000,1.889870)(335.980000,1.887470)(337.980000,1.889840)(339.980000,1.890510)(341.980000,1.888570)(343.980000,1.890050)(345.980000,1.887160)(347.980000,1.889500)(349.980000,1.890780)}; 
\addlegendentry{\kmeans, no LRN (ours)}
\end{axis}
\end{tikzpicture}

%% file: figures/imagenet/googlenet_train_loss.tex
\begin{tikzpicture}[scale=0.7]
\begin{axis}[
  legend cell align=left,
  height=0.95\textwidth,
  width=1.5\textwidth,
  xmin=0, xmax=2.4,
  xtick={0, 0.5, 1, 1.5, 2},
  ymax=7,
  legend pos=north east,
  scaled x ticks = false,
  xticklabel = {\pgfmathprintnumber{\tick}M}
]

\addplot[color=ta3gray,ultra thick,dashed] coordinates {(0.000000,6.578551)(0.004800,6.255088)(0.009600,5.802617)(0.014400,5.439716)(0.019200,5.146286)(0.024000,4.907068)(0.028800,4.699059)(0.033600,4.549220)(0.038400,4.420564)(0.043200,4.298716)(0.048000,4.202760)(0.052800,4.097090)(0.057600,4.010900)(0.062400,3.924010)(0.067200,3.853209)(0.072000,3.789961)(0.076800,3.762890)(0.081600,3.723008)(0.086400,3.670761)(0.091200,3.630230)(0.096000,3.592868)(0.100800,3.548572)(0.105600,3.497318)(0.110400,3.498229)(0.115200,3.486041)(0.120000,3.450256)(0.124800,3.424645)(0.129600,3.371392)(0.134400,3.340652)(0.139200,3.337372)(0.144000,3.345666)(0.148800,3.350907)(0.153600,3.330690)(0.158400,3.323082)(0.163200,3.310307)(0.168000,3.263030)(0.172800,3.234281)(0.177600,3.223389)(0.182400,3.226089)(0.187200,3.217894)(0.192000,3.176388)(0.196800,3.165997)(0.201600,3.171859)(0.206400,3.179559)(0.211200,3.153706)(0.216000,3.134288)(0.220800,3.110945)(0.225600,3.094875)(0.230400,3.084357)(0.235200,3.084413)(0.240000,3.084495)(0.244800,3.074950)(0.249600,3.077416)(0.254400,3.105589)(0.259200,3.064395)(0.264000,3.045635)(0.268800,3.045347)(0.273600,3.032436)(0.278400,3.024683)(0.283200,2.978014)(0.288000,3.001434)(0.292800,3.036336)(0.297600,2.989486)(0.302400,2.946808)(0.307200,2.956613)(0.312000,2.982796)(0.316800,3.002507)(0.321600,2.963376)(0.326400,2.967861)(0.331200,2.985795)(0.336000,2.946986)(0.340800,2.906311)(0.345600,2.896061)(0.350400,2.908543)(0.355200,2.903450)(0.360000,2.863277)(0.364800,2.874361)(0.369600,2.889028)(0.374400,2.889076)(0.379200,2.871059)(0.384000,2.881287)(0.388800,2.875433)(0.393600,2.872271)(0.398400,2.843727)(0.403200,2.832709)(0.408000,2.860521)(0.412800,2.888986)(0.417600,2.868326)(0.422400,2.839125)(0.427200,2.845321)(0.432000,2.849688)(0.436800,2.834930)(0.441600,2.820894)(0.446400,2.820297)(0.451200,2.845009)(0.456000,2.856043)(0.460800,2.814697)(0.465600,2.782831)(0.470400,2.794974)(0.475200,2.789735)(0.480000,2.761243)(0.484800,2.793807)(0.489600,2.831955)(0.494400,2.821254)(0.499200,2.809870)(0.504000,2.789618)(0.508800,2.781145)(0.513600,2.765148)(0.518400,2.765512)(0.523200,2.749760)(0.528000,2.712980)(0.532800,2.718144)(0.537600,2.732754)(0.542400,2.739035)(0.547200,2.712803)(0.552000,2.718713)(0.556800,2.744245)(0.561600,2.770112)(0.566400,2.764602)(0.571200,2.756217)(0.576000,2.747701)(0.580800,2.763285)(0.585600,2.736316)(0.590400,2.723081)(0.595200,2.704746)(0.600000,2.694875)(0.604800,2.682544)(0.609600,2.685837)(0.614400,2.706921)(0.619200,2.670830)(0.624000,2.661971)(0.628800,2.676579)(0.633600,2.684102)(0.638400,2.694543)(0.643200,2.701701)(0.648000,2.689980)(0.652800,2.677958)(0.657600,2.694023)(0.662400,2.689982)(0.667200,2.634571)(0.672000,2.653236)(0.676800,2.708743)(0.681600,2.715114)(0.686400,2.702603)(0.691200,2.675597)(0.696000,2.644777)(0.700800,2.635119)(0.705600,2.669333)(0.710400,2.669030)(0.715200,2.650734)(0.720000,2.639963)(0.724800,2.643718)(0.729600,2.612836)(0.734400,2.601831)(0.739200,2.633705)(0.744000,2.660917)(0.748800,2.656177)(0.753600,2.637161)(0.758400,2.626729)(0.763200,2.625862)(0.768000,2.630513)(0.772800,2.618845)(0.777600,2.605800)(0.782400,2.585846)(0.787200,2.587047)(0.792000,2.591759)(0.796800,2.595760)(0.801600,2.598858)(0.806400,2.586685)(0.811200,2.630788)(0.816000,2.614129)(0.820800,2.614068)(0.825600,2.593095)(0.830400,2.586980)(0.835200,2.573361)(0.840000,2.588648)(0.844800,2.584769)(0.849600,2.576961)(0.854400,2.569600)(0.859200,2.579192)(0.864000,2.575977)(0.868800,2.548583)(0.873600,2.555814)(0.878400,2.562458)(0.883200,2.550549)(0.888000,2.576740)(0.892800,2.582095)(0.897600,2.568425)(0.902400,2.548759)(0.907200,2.519463)(0.912000,2.531561)(0.916800,2.545020)(0.921600,2.519123)(0.926400,2.486468)(0.931200,2.518794)(0.936000,2.541456)(0.940800,2.540085)(0.945600,2.538017)(0.950400,2.531513)(0.955200,2.525120)(0.960000,2.563222)(0.964800,2.584924)(0.969600,2.554286)(0.974400,2.530393)(0.979200,2.524270)(0.984000,2.535694)(0.988800,2.518591)(0.993600,2.507928)(0.998400,2.485051)(1.003200,2.502541)(1.008000,2.511372)(1.012800,2.526023)(1.017600,2.538644)(1.022400,2.522642)(1.027200,2.493903)(1.032000,2.478692)(1.036800,2.471261)(1.041600,2.487922)(1.046400,2.455830)(1.051200,2.446798)(1.056000,2.426819)(1.060800,2.452722)(1.065600,2.479513)(1.070400,2.519833)(1.075200,2.470790)(1.080000,2.407162)(1.084800,2.411776)(1.089600,2.420742)(1.094400,2.440931)(1.099200,2.470930)(1.104000,2.476715)(1.108800,2.505335)(1.113600,2.478965)(1.118400,2.417702)(1.123200,2.386628)(1.128000,2.429462)(1.132800,2.415269)(1.137600,2.404648)(1.142400,2.408040)(1.147200,2.422656)(1.152000,2.448482)(1.156800,2.439259)(1.161600,2.421584)(1.166400,2.419886)(1.171200,2.415606)(1.176000,2.399202)(1.180800,2.373262)(1.185600,2.333888)(1.190400,2.320675)(1.195200,2.332618)(1.200000,2.370683)(1.204800,2.417899)(1.209600,2.400448)(1.214400,2.405060)(1.219200,2.368869)(1.224000,2.356098)(1.228800,2.375746)(1.233600,2.365468)(1.238400,2.351450)(1.243200,2.383280)(1.248000,2.394256)(1.252800,2.416733)(1.257600,2.378545)(1.262400,2.344555)(1.267200,2.382973)(1.272000,2.380557)(1.276800,2.346591)(1.281600,2.322280)(1.286400,2.317611)(1.291200,2.313345)(1.296000,2.333648)(1.300800,2.335029)(1.305600,2.343337)(1.310400,2.347648)(1.315200,2.374484)(1.320000,2.356206)(1.324800,2.353362)(1.329600,2.325725)(1.334400,2.309102)(1.339200,2.309726)(1.344000,2.303863)(1.348800,2.291937)(1.353600,2.307092)(1.358400,2.300943)(1.363200,2.274428)(1.368000,2.309185)(1.372800,2.328981)(1.377600,2.306140)(1.382400,2.309119)(1.387200,2.287026)(1.392000,2.285270)(1.396800,2.306981)(1.401600,2.322538)(1.406400,2.317151)(1.411200,2.314241)(1.416000,2.328104)(1.420800,2.301214)(1.425600,2.265597)(1.430400,2.256813)(1.435200,2.260646)(1.440000,2.255174)(1.444800,2.245878)(1.449600,2.210386)(1.454400,2.219594)(1.459200,2.259015)(1.464000,2.282757)(1.468800,2.265924)(1.473600,2.260691)(1.478400,2.239361)(1.483200,2.236740)(1.488000,2.246114)(1.492800,2.220472)(1.497600,2.205900)(1.502400,2.225158)(1.507200,2.255139)(1.512000,2.255425)(1.516800,2.217892)(1.521600,2.239150)(1.526400,2.228787)(1.531200,2.220506)(1.536000,2.210627)(1.540800,2.210185)(1.545600,2.209825)(1.550400,2.194648)(1.555200,2.191674)(1.560000,2.206097)(1.564800,2.213421)(1.569600,2.201402)(1.574400,2.184247)(1.579200,2.188067)(1.584000,2.172767)(1.588800,2.158583)(1.593600,2.161691)(1.598400,2.149689)(1.603200,2.163408)(1.608000,2.179143)(1.612800,2.199473)(1.617600,2.196547)(1.622400,2.165393)(1.627200,2.178441)(1.632000,2.170361)(1.636800,2.171372)(1.641600,2.135537)(1.646400,2.134988)(1.651200,2.175260)(1.656000,2.157914)(1.660800,2.108435)(1.665600,2.114881)(1.670400,2.132377)(1.675200,2.145370)(1.680000,2.158388)(1.684800,2.134827)(1.689600,2.140867)(1.694400,2.118047)(1.699200,2.083432)(1.704000,2.084568)(1.708800,2.091738)(1.713600,2.110761)(1.718400,2.103043)(1.723200,2.062679)(1.728000,2.063159)(1.732800,2.060060)(1.737600,2.055139)(1.742400,2.068454)(1.747200,2.070813)(1.752000,2.077327)(1.756800,2.079715)(1.761600,2.054022)(1.766400,2.066890)(1.771200,2.080810)(1.776000,2.107391)(1.780800,2.096101)(1.785600,2.057023)(1.790400,2.052735)(1.795200,2.045942)(1.800000,2.042579)(1.804800,2.054054)(1.809600,2.046462)(1.814400,2.038248)(1.819200,2.044225)(1.824000,2.010241)(1.828800,2.009996)(1.833600,1.988629)(1.838400,1.984933)(1.843200,2.001891)(1.848000,2.042655)(1.852800,2.037506)(1.857600,2.001721)(1.862400,2.001896)(1.867200,2.013408)(1.872000,2.001318)(1.876800,1.970919)(1.881600,1.949718)(1.886400,1.949341)(1.891200,1.952621)(1.896000,1.941093)(1.900800,1.937683)(1.905600,1.922159)(1.910400,1.922259)(1.915200,1.950396)(1.920000,1.970216)(1.924800,1.963818)(1.929600,1.943591)(1.934400,1.931352)(1.939200,1.928705)(1.944000,1.947512)(1.948800,1.939405)(1.953600,1.938284)(1.958400,1.909875)(1.963200,1.879661)(1.968000,1.865850)(1.972800,1.890869)(1.977600,1.897893)(1.982400,1.854694)(1.987200,1.858760)(1.992000,1.890436)(1.996800,1.867277)(2.001600,1.852756)(2.006400,1.846082)(2.011200,1.847582)(2.016000,1.873817)(2.020800,1.880639)(2.025600,1.861557)(2.030400,1.825296)(2.035200,1.865283)(2.040000,1.890348)(2.044800,1.887804)(2.049600,1.857923)(2.054400,1.831359)(2.059200,1.808189)(2.064000,1.807714)(2.068800,1.809582)(2.073600,1.814076)(2.078400,1.806377)(2.083200,1.779330)(2.088000,1.776104)(2.092800,1.765524)(2.097600,1.793205)(2.102400,1.778975)(2.107200,1.760071)(2.112000,1.761680)(2.116800,1.751476)(2.121600,1.741043)(2.126400,1.750210)(2.131200,1.754140)(2.136000,1.746784)(2.140800,1.717338)(2.145600,1.686276)(2.150400,1.713283)(2.155200,1.700959)(2.160000,1.705491)(2.164800,1.686669)(2.169600,1.706780)(2.174400,1.712205)(2.179200,1.683584)(2.184000,1.660271)(2.188800,1.644124)(2.193600,1.657737)(2.198400,1.638580)(2.203200,1.636378)(2.208000,1.614901)(2.212800,1.614305)(2.217600,1.629081)(2.222400,1.612211)(2.227200,1.608418)(2.232000,1.591378)(2.236800,1.596024)(2.241600,1.570162)(2.246400,1.568667)(2.251200,1.573560)(2.256000,1.566348)(2.260800,1.572383)(2.265600,1.534821)(2.270400,1.526397)(2.275200,1.519039)(2.280000,1.479727)(2.284800,1.437448)(2.289600,1.452716)(2.294400,1.485571)(2.299200,1.477818)(2.304000,1.471471)(2.308800,1.463300)(2.313600,1.448878)(2.318400,1.455061)(2.323200,1.444561)(2.328000,1.422283)(2.332800,1.404941)(2.337600,1.390260)(2.342400,1.384374)(2.347200,1.362419)(2.352000,1.314536)(2.356800,1.282216)(2.361600,1.271529)(2.366400,1.274905)(2.371200,1.243530)(2.376000,1.231321)(2.380800,1.208624)(2.385600,1.172849)(2.390400,1.142943)(2.395200,1.117768)}; 
\addplot[color=taskyblue,ultra thick,solid] coordinates {(0.000000,6.108533)(0.004800,5.707867)(0.009600,5.220636)(0.014400,4.875886)(0.019200,4.609128)(0.024000,4.388885)(0.028800,4.203531)(0.033600,4.047036)(0.038400,3.932318)(0.043200,3.832614)(0.048000,3.748099)(0.052800,3.649839)(0.057600,3.578952)(0.062400,3.481315)(0.067200,3.411697)(0.072000,3.355431)(0.076800,3.324352)(0.081600,3.279986)(0.086400,3.224129)(0.091200,3.198708)(0.096000,3.180254)(0.100800,3.131483)(0.105600,3.068445)(0.110400,3.072110)(0.115200,3.077422)(0.120000,3.031780)(0.124800,3.005957)(0.129600,2.984850)(0.134400,2.915251)(0.139200,2.894199)(0.144000,2.919571)(0.148800,2.935589)(0.153600,2.899993)(0.158400,2.889968)(0.163200,2.888882)(0.168000,2.842650)(0.172800,2.826241)(0.177600,2.814369)(0.182400,2.832455)(0.187200,2.808413)(0.192000,2.768196)(0.196800,2.774106)(0.201600,2.797971)(0.206400,2.811222)(0.211200,2.778755)(0.216000,2.753568)(0.220800,2.733849)(0.225600,2.714344)(0.230400,2.697959)(0.235200,2.711186)(0.240000,2.707136)(0.244800,2.680084)(0.249600,2.675842)(0.254400,2.701766)(0.259200,2.655182)(0.264000,2.653862)(0.268800,2.658922)(0.273600,2.660749)(0.278400,2.657152)(0.283200,2.622312)(0.288000,2.632148)(0.292800,2.654012)(0.297600,2.604780)(0.302400,2.563278)(0.307200,2.576705)(0.312000,2.607400)(0.316800,2.637639)(0.321600,2.617236)(0.326400,2.616216)(0.331200,2.600584)(0.336000,2.537509)(0.340800,2.518750)(0.345600,2.541113)(0.350400,2.561815)(0.355200,2.549897)(0.360000,2.528917)(0.364800,2.518726)(0.369600,2.516337)(0.374400,2.516744)(0.379200,2.508017)(0.384000,2.521432)(0.388800,2.524959)(0.393600,2.532984)(0.398400,2.498172)(0.403200,2.469881)(0.408000,2.484138)(0.412800,2.520745)(0.417600,2.515347)(0.422400,2.492344)(0.427200,2.500596)(0.432000,2.506132)(0.436800,2.460128)(0.441600,2.473548)(0.446400,2.477442)(0.451200,2.488655)(0.456000,2.487849)(0.460800,2.454906)(0.465600,2.435798)(0.470400,2.445988)(0.475200,2.428613)(0.480000,2.417850)(0.484800,2.445886)(0.489600,2.469480)(0.494400,2.456107)(0.499200,2.450494)(0.504000,2.423713)(0.508800,2.421468)(0.513600,2.407250)(0.518400,2.400369)(0.523200,2.409959)(0.528000,2.385517)(0.532800,2.375986)(0.537600,2.398933)(0.542400,2.392707)(0.547200,2.372422)(0.552000,2.392824)(0.556800,2.425016)(0.561600,2.423954)(0.566400,2.419224)(0.571200,2.408291)(0.576000,2.372290)(0.580800,2.381711)(0.585600,2.362564)(0.590400,2.376988)(0.595200,2.375920)(0.600000,2.389526)(0.604800,2.354135)(0.609600,2.326247)(0.614400,2.338755)(0.619200,2.309663)(0.624000,2.311636)(0.628800,2.320614)(0.633600,2.332907)(0.638400,2.329937)(0.643200,2.334189)(0.648000,2.354857)(0.652800,2.335059)(0.657600,2.342104)(0.662400,2.359491)(0.667200,2.321627)(0.672000,2.321600)(0.676800,2.377942)(0.681600,2.406307)(0.686400,2.379616)(0.691200,2.350163)(0.696000,2.345930)(0.700800,2.336596)(0.705600,2.347011)(0.710400,2.343975)(0.715200,2.341505)(0.720000,2.315632)(0.724800,2.315630)(0.729600,2.302579)(0.734400,2.306282)(0.739200,2.308219)(0.744000,2.330888)(0.748800,2.322334)(0.753600,2.287428)(0.758400,2.271560)(0.763200,2.278257)(0.768000,2.294120)(0.772800,2.290167)(0.777600,2.298790)(0.782400,2.260086)(0.787200,2.248036)(0.792000,2.261063)(0.796800,2.258797)(0.801600,2.282277)(0.806400,2.292571)(0.811200,2.332995)(0.816000,2.311055)(0.820800,2.286114)(0.825600,2.264195)(0.830400,2.277517)(0.835200,2.254954)(0.840000,2.249817)(0.844800,2.259619)(0.849600,2.269365)(0.854400,2.268188)(0.859200,2.253665)(0.864000,2.263949)(0.868800,2.251805)(0.873600,2.243542)(0.878400,2.248705)(0.883200,2.229185)(0.888000,2.246962)(0.892800,2.278184)(0.897600,2.273002)(0.902400,2.255396)(0.907200,2.224968)(0.912000,2.217892)(0.916800,2.234715)(0.921600,2.221930)(0.926400,2.192371)(0.931200,2.224301)(0.936000,2.244365)(0.940800,2.231370)(0.945600,2.235552)(0.950400,2.244886)(0.955200,2.245341)(0.960000,2.239480)(0.964800,2.240405)(0.969600,2.231929)(0.974400,2.223373)(0.979200,2.219585)(0.984000,2.240340)(0.988800,2.220896)(0.993600,2.214619)(0.998400,2.197047)(1.003200,2.212977)(1.008000,2.203560)(1.012800,2.201427)(1.017600,2.199104)(1.022400,2.188570)(1.027200,2.181662)(1.032000,2.187388)(1.036800,2.170158)(1.041600,2.180856)(1.046400,2.155406)(1.051200,2.159318)(1.056000,2.147571)(1.060800,2.148924)(1.065600,2.180739)(1.070400,2.229982)(1.075200,2.193860)(1.080000,2.130207)(1.084800,2.119301)(1.089600,2.153602)(1.094400,2.160692)(1.099200,2.176195)(1.104000,2.169544)(1.108800,2.197019)(1.113600,2.192870)(1.118400,2.176454)(1.123200,2.168481)(1.128000,2.184216)(1.132800,2.172166)(1.137600,2.148514)(1.142400,2.142301)(1.147200,2.140351)(1.152000,2.140972)(1.156800,2.121675)(1.161600,2.124793)(1.166400,2.130595)(1.171200,2.142292)(1.176000,2.147046)(1.180800,2.095125)(1.185600,2.069352)(1.190400,2.084522)(1.195200,2.093360)(1.200000,2.127409)(1.204800,2.157627)(1.209600,2.122177)(1.214400,2.132451)(1.219200,2.121553)(1.224000,2.089357)(1.228800,2.083349)(1.233600,2.088832)(1.238400,2.085479)(1.243200,2.124494)(1.248000,2.134349)(1.252800,2.132877)(1.257600,2.108535)(1.262400,2.060333)(1.267200,2.099731)(1.272000,2.123348)(1.276800,2.085571)(1.281600,2.052698)(1.286400,2.043567)(1.291200,2.040739)(1.296000,2.059707)(1.300800,2.057129)(1.305600,2.050522)(1.310400,2.066992)(1.315200,2.090894)(1.320000,2.060855)(1.324800,2.073322)(1.329600,2.058685)(1.334400,2.061620)(1.339200,2.078382)(1.344000,2.066095)(1.348800,2.013419)(1.353600,2.023451)(1.358400,2.012199)(1.363200,2.006940)(1.368000,2.045599)(1.372800,2.064378)(1.377600,2.045873)(1.382400,2.035753)(1.387200,2.003621)(1.392000,2.005717)(1.396800,2.034401)(1.401600,2.060906)(1.406400,2.059170)(1.411200,2.023934)(1.416000,2.027227)(1.420800,2.029984)(1.425600,1.996465)(1.430400,1.966451)(1.435200,1.978090)(1.440000,1.990757)(1.444800,2.000641)(1.449600,1.970312)(1.454400,1.986371)(1.459200,2.003364)(1.464000,2.002730)(1.468800,1.995186)(1.473600,2.012786)(1.478400,2.003240)(1.483200,1.971337)(1.488000,1.947519)(1.492800,1.931444)(1.497600,1.934980)(1.502400,1.958646)(1.507200,1.978494)(1.512000,1.998001)(1.516800,1.983180)(1.521600,2.003596)(1.526400,2.007099)(1.531200,2.000553)(1.536000,1.984685)(1.540800,1.970889)(1.545600,1.982710)(1.550400,1.974715)(1.555200,1.971358)(1.560000,1.983636)(1.564800,1.983327)(1.569600,1.976797)(1.574400,1.947413)(1.579200,1.931278)(1.584000,1.926363)(1.588800,1.922173)(1.593600,1.920096)(1.598400,1.916144)(1.603200,1.921973)(1.608000,1.921317)(1.612800,1.941303)(1.617600,1.951357)(1.622400,1.925149)(1.627200,1.937239)(1.632000,1.928762)(1.636800,1.929956)(1.641600,1.910662)(1.646400,1.902247)(1.651200,1.922713)(1.656000,1.916777)(1.660800,1.898286)(1.665600,1.896552)(1.670400,1.900873)(1.675200,1.908405)(1.680000,1.910780)(1.684800,1.897349)(1.689600,1.917028)(1.694400,1.909629)(1.699200,1.878126)(1.704000,1.884717)(1.708800,1.886532)(1.713600,1.893244)(1.718400,1.876644)(1.723200,1.840792)(1.728000,1.849364)(1.732800,1.865137)(1.737600,1.843316)(1.742400,1.852334)(1.747200,1.853059)(1.752000,1.864613)(1.756800,1.864199)(1.761600,1.844950)(1.766400,1.845310)(1.771200,1.857825)(1.776000,1.879814)(1.780800,1.850583)(1.785600,1.806815)(1.790400,1.822951)(1.795200,1.830357)(1.800000,1.815088)(1.804800,1.811542)(1.809600,1.821060)(1.814400,1.846942)(1.819200,1.845793)(1.824000,1.822150)(1.828800,1.820906)(1.833600,1.788536)(1.838400,1.763354)(1.843200,1.770260)(1.848000,1.804671)(1.852800,1.822733)(1.857600,1.817322)(1.862400,1.810213)(1.867200,1.803314)(1.872000,1.795228)(1.876800,1.776549)(1.881600,1.753264)(1.886400,1.746617)(1.891200,1.739848)(1.896000,1.736508)(1.900800,1.750405)(1.905600,1.728321)(1.910400,1.718368)(1.915200,1.724935)(1.920000,1.751060)(1.924800,1.768310)(1.929600,1.754108)(1.934400,1.734893)(1.939200,1.757058)(1.944000,1.750217)(1.948800,1.731258)(1.953600,1.739036)(1.958400,1.717338)(1.963200,1.708852)(1.968000,1.701408)(1.972800,1.701385)(1.977600,1.681196)(1.982400,1.648256)(1.987200,1.656686)(1.992000,1.694651)(1.996800,1.695603)(2.001600,1.674317)(2.006400,1.672825)(2.011200,1.684721)(2.016000,1.699749)(2.020800,1.709284)(2.025600,1.690617)(2.030400,1.654919)(2.035200,1.679880)(2.040000,1.708902)(2.044800,1.702879)(2.049600,1.681405)(2.054400,1.657620)(2.059200,1.636692)(2.064000,1.644676)(2.068800,1.645053)(2.073600,1.644486)(2.078400,1.636626)(2.083200,1.613448)(2.088000,1.601208)(2.092800,1.579292)(2.097600,1.606370)(2.102400,1.610362)(2.107200,1.590424)(2.112000,1.580903)(2.116800,1.575655)(2.121600,1.579772)(2.126400,1.582308)(2.131200,1.583495)(2.136000,1.596917)(2.140800,1.576891)(2.145600,1.524991)(2.150400,1.543838)(2.155200,1.553786)(2.160000,1.560672)(2.164800,1.527878)(2.169600,1.542655)(2.174400,1.560941)(2.179200,1.536284)(2.184000,1.510571)(2.188800,1.512243)(2.193600,1.515244)(2.198400,1.498590)(2.203200,1.500764)(2.208000,1.490395)(2.212800,1.487867)(2.217600,1.481637)(2.222400,1.476637)(2.227200,1.473493)(2.232000,1.471536)(2.236800,1.458237)(2.241600,1.449350)(2.246400,1.446705)(2.251200,1.445922)(2.256000,1.445242)(2.260800,1.423646)(2.265600,1.398573)(2.270400,1.392332)(2.275200,1.383602)(2.280000,1.364990)(2.284800,1.345804)(2.289600,1.363745)(2.294400,1.382586)(2.299200,1.372972)(2.304000,1.364982)(2.308800,1.368261)(2.313600,1.341637)(2.318400,1.339710)(2.323200,1.337069)(2.328000,1.306602)(2.332800,1.296397)(2.337600,1.281103)(2.342400,1.267119)(2.347200,1.252659)(2.352000,1.238249)(2.356800,1.210055)(2.361600,1.185683)(2.366400,1.179903)(2.371200,1.156570)(2.376000,1.165617)(2.380800,1.146296)(2.385600,1.120767)(2.390400,1.090293)(2.395200,1.041080)}; 
\addplot[color=ta3scarletred,ultra thick,dashed] coordinates {(0.000000,6.103691)(0.004800,5.693313)(0.009600,5.192724)(0.014400,4.849443)(0.019200,4.592167)(0.024000,4.370978)(0.028800,4.179181)(0.033600,4.036182)(0.038400,3.924475)(0.043200,3.829025)(0.048000,3.739044)(0.052800,3.639048)(0.057600,3.559961)(0.062400,3.463108)(0.067200,3.405317)(0.072000,3.340285)(0.076800,3.309359)(0.081600,3.267154)(0.086400,3.213765)(0.091200,3.175258)(0.096000,3.158334)(0.100800,3.136828)(0.105600,3.096637)(0.110400,3.068730)(0.115200,3.068753)(0.120000,3.035011)(0.124800,3.000623)(0.129600,2.969962)(0.134400,2.908437)(0.139200,2.915299)(0.144000,2.938828)(0.148800,2.951711)(0.153600,2.919873)(0.158400,2.888826)(0.163200,2.908113)(0.168000,2.864857)(0.172800,2.845668)(0.177600,2.832436)(0.182400,2.816131)(0.187200,2.793455)(0.192000,2.745236)(0.196800,2.750545)(0.201600,2.803091)(0.206400,2.813473)(0.211200,2.768261)(0.216000,2.747170)(0.220800,2.707591)(0.225600,2.685590)(0.230400,2.671964)(0.235200,2.676388)(0.240000,2.690809)(0.244800,2.680636)(0.249600,2.656366)(0.254400,2.694594)(0.259200,2.672549)(0.264000,2.651583)(0.268800,2.655854)(0.273600,2.657693)(0.278400,2.646295)(0.283200,2.595461)(0.288000,2.604961)(0.292800,2.630215)(0.297600,2.612274)(0.302400,2.579940)(0.307200,2.587246)(0.312000,2.589163)(0.316800,2.612493)(0.321600,2.596944)(0.326400,2.609111)(0.331200,2.605885)(0.336000,2.546200)(0.340800,2.518276)(0.345600,2.541297)(0.350400,2.570818)(0.355200,2.525247)(0.360000,2.490806)(0.364800,2.513943)(0.369600,2.511019)(0.374400,2.508197)(0.379200,2.494369)(0.384000,2.512908)(0.388800,2.519442)(0.393600,2.509694)(0.398400,2.487433)(0.403200,2.477905)(0.408000,2.494099)(0.412800,2.521326)(0.417600,2.510153)(0.422400,2.469119)(0.427200,2.474554)(0.432000,2.486796)(0.436800,2.461262)(0.441600,2.448955)(0.446400,2.456432)(0.451200,2.467180)(0.456000,2.463636)(0.460800,2.450528)(0.465600,2.441695)(0.470400,2.449648)(0.475200,2.438371)(0.480000,2.418758)(0.484800,2.434026)(0.489600,2.469720)(0.494400,2.452460)(0.499200,2.428768)(0.504000,2.431145)(0.508800,2.437378)(0.513600,2.404441)(0.518400,2.382709)(0.523200,2.384168)(0.528000,2.389954)(0.532800,2.396717)(0.537600,2.404668)(0.542400,2.416531)(0.547200,2.378788)(0.552000,2.376161)(0.556800,2.402587)(0.561600,2.414552)(0.566400,2.403440)(0.571200,2.400578)(0.576000,2.390377)(0.580800,2.393526)(0.585600,2.387766)(0.590400,2.405931)(0.595200,2.404258)(0.600000,2.380751)(0.604800,2.337774)(0.609600,2.323264)(0.614400,2.358373)(0.619200,2.324280)(0.624000,2.300454)(0.628800,2.306142)(0.633600,2.315141)(0.638400,2.306331)(0.643200,2.323000)(0.648000,2.317548)(0.652800,2.323438)(0.657600,2.335484)(0.662400,2.369389)(0.667200,2.332517)(0.672000,2.317042)(0.676800,2.362686)(0.681600,2.373118)(0.686400,2.352659)(0.691200,2.330070)(0.696000,2.317840)(0.700800,2.316133)(0.705600,2.345603)(0.710400,2.333676)(0.715200,2.319003)(0.720000,2.293005)(0.724800,2.290578)(0.729600,2.291870)(0.734400,2.316459)(0.739200,2.317068)(0.744000,2.321921)(0.748800,2.296000)(0.753600,2.281339)(0.758400,2.280886)(0.763200,2.300244)(0.768000,2.293584)(0.772800,2.267615)(0.777600,2.277769)(0.782400,2.251129)(0.787200,2.239375)(0.792000,2.267548)(0.796800,2.265574)(0.801600,2.278799)(0.806400,2.280102)(0.811200,2.300397)(0.816000,2.282445)(0.820800,2.267057)(0.825600,2.246024)(0.830400,2.254521)(0.835200,2.255128)(0.840000,2.257706)(0.844800,2.242767)(0.849600,2.241478)(0.854400,2.248515)(0.859200,2.251621)(0.864000,2.244061)(0.868800,2.227150)(0.873600,2.220370)(0.878400,2.218681)(0.883200,2.207019)(0.888000,2.258296)(0.892800,2.273659)(0.897600,2.274469)(0.902400,2.253050)(0.907200,2.218976)(0.912000,2.218249)(0.916800,2.216136)(0.921600,2.193195)(0.926400,2.174061)(0.931200,2.216837)(0.936000,2.228824)(0.940800,2.216064)(0.945600,2.233319)(0.950400,2.231550)(0.955200,2.213203)(0.960000,2.222179)(0.964800,2.237122)(0.969600,2.231259)(0.974400,2.207327)(0.979200,2.193574)(0.984000,2.216422)(0.988800,2.206099)(0.993600,2.195647)(0.998400,2.189724)(1.003200,2.208606)(1.008000,2.195996)(1.012800,2.205772)(1.017600,2.213118)(1.022400,2.197173)(1.027200,2.191763)(1.032000,2.195486)(1.036800,2.176182)(1.041600,2.194895)(1.046400,2.176012)(1.051200,2.161521)(1.056000,2.141702)(1.060800,2.162738)(1.065600,2.186732)(1.070400,2.211229)(1.075200,2.188366)(1.080000,2.122387)(1.084800,2.095630)(1.089600,2.137570)(1.094400,2.172781)(1.099200,2.181857)(1.104000,2.161386)(1.108800,2.198477)(1.113600,2.177775)(1.118400,2.130245)(1.123200,2.132040)(1.128000,2.154537)(1.132800,2.126641)(1.137600,2.089019)(1.142400,2.095066)(1.147200,2.123757)(1.152000,2.139178)(1.156800,2.115113)(1.161600,2.111066)(1.166400,2.134879)(1.171200,2.138132)(1.176000,2.118820)(1.180800,2.092744)(1.185600,2.081470)(1.190400,2.082107)(1.195200,2.054505)(1.200000,2.085417)(1.204800,2.140104)(1.209600,2.115262)(1.214400,2.123431)(1.219200,2.082691)(1.224000,2.061099)(1.228800,2.090143)(1.233600,2.096521)(1.238400,2.091389)(1.243200,2.113619)(1.248000,2.114168)(1.252800,2.132773)(1.257600,2.108626)(1.262400,2.075754)(1.267200,2.102239)(1.272000,2.101844)(1.276800,2.081706)(1.281600,2.045642)(1.286400,2.038055)(1.291200,2.041041)(1.296000,2.066226)(1.300800,2.075615)(1.305600,2.085046)(1.310400,2.059504)(1.315200,2.073594)(1.320000,2.059249)(1.324800,2.066786)(1.329600,2.036676)(1.334400,2.032756)(1.339200,2.042512)(1.344000,2.044015)(1.348800,2.011295)(1.353600,2.016980)(1.358400,2.028734)(1.363200,2.024092)(1.368000,2.062190)(1.372800,2.074229)(1.377600,2.050070)(1.382400,2.047805)(1.387200,2.031568)(1.392000,2.029688)(1.396800,2.025990)(1.401600,2.038243)(1.406400,2.057880)(1.411200,2.050548)(1.416000,2.065730)(1.420800,2.052335)(1.425600,2.008958)(1.430400,1.974459)(1.435200,1.985950)(1.440000,2.001783)(1.444800,1.998922)(1.449600,1.966058)(1.454400,1.989868)(1.459200,2.018669)(1.464000,2.037274)(1.468800,2.009522)(1.473600,2.006079)(1.478400,2.005457)(1.483200,1.984939)(1.488000,1.971360)(1.492800,1.951809)(1.497600,1.947523)(1.502400,1.955040)(1.507200,1.987174)(1.512000,1.989987)(1.516800,1.973722)(1.521600,2.003482)(1.526400,2.006074)(1.531200,1.997420)(1.536000,1.992676)(1.540800,1.969014)(1.545600,1.956751)(1.550400,1.960608)(1.555200,1.948132)(1.560000,1.960758)(1.564800,1.982674)(1.569600,1.974122)(1.574400,1.942677)(1.579200,1.934108)(1.584000,1.931601)(1.588800,1.927659)(1.593600,1.927209)(1.598400,1.914670)(1.603200,1.932345)(1.608000,1.946927)(1.612800,1.978436)(1.617600,1.996988)(1.622400,1.946868)(1.627200,1.937787)(1.632000,1.924077)(1.636800,1.930408)(1.641600,1.922409)(1.646400,1.909848)(1.651200,1.928847)(1.656000,1.929326)(1.660800,1.892819)(1.665600,1.895267)(1.670400,1.907290)(1.675200,1.929265)(1.680000,1.924555)(1.684800,1.918289)(1.689600,1.932948)(1.694400,1.907111)(1.699200,1.857636)(1.704000,1.859956)(1.708800,1.895795)(1.713600,1.907140)(1.718400,1.866168)(1.723200,1.842766)(1.728000,1.860078)(1.732800,1.872420)(1.737600,1.867069)(1.742400,1.869076)(1.747200,1.879747)(1.752000,1.877794)(1.756800,1.872822)(1.761600,1.855477)(1.766400,1.858996)(1.771200,1.855745)(1.776000,1.887486)(1.780800,1.857648)(1.785600,1.810338)(1.790400,1.842686)(1.795200,1.862428)(1.800000,1.847039)(1.804800,1.840926)(1.809600,1.844413)(1.814400,1.860428)(1.819200,1.852852)(1.824000,1.821674)(1.828800,1.822021)(1.833600,1.821729)(1.838400,1.795111)(1.843200,1.777133)(1.848000,1.811962)(1.852800,1.812841)(1.857600,1.798425)(1.862400,1.802388)(1.867200,1.813078)(1.872000,1.808414)(1.876800,1.795334)(1.881600,1.780089)(1.886400,1.770056)(1.891200,1.737474)(1.896000,1.745702)(1.900800,1.771170)(1.905600,1.755401)(1.910400,1.744010)(1.915200,1.730743)(1.920000,1.745482)(1.924800,1.775223)(1.929600,1.750604)(1.934400,1.736585)(1.939200,1.758636)(1.944000,1.755437)(1.948800,1.752740)(1.953600,1.766698)(1.958400,1.729299)(1.963200,1.715453)(1.968000,1.695832)(1.972800,1.700681)(1.977600,1.687927)(1.982400,1.650559)(1.987200,1.665503)(1.992000,1.698227)(1.996800,1.710427)(2.001600,1.701903)(2.006400,1.674133)(2.011200,1.691425)(2.016000,1.688054)(2.020800,1.687555)(2.025600,1.703625)(2.030400,1.676550)(2.035200,1.716066)(2.040000,1.735717)(2.044800,1.714383)(2.049600,1.689647)(2.054400,1.672834)(2.059200,1.651653)(2.064000,1.660331)(2.068800,1.658923)(2.073600,1.658285)(2.078400,1.639304)(2.083200,1.619081)(2.088000,1.619311)(2.092800,1.614321)(2.097600,1.635510)(2.102400,1.628935)(2.107200,1.610870)(2.112000,1.609005)(2.116800,1.613167)(2.121600,1.613842)(2.126400,1.607137)(2.131200,1.596165)(2.136000,1.607671)(2.140800,1.592015)(2.145600,1.539516)(2.150400,1.566626)(2.155200,1.591124)(2.160000,1.590627)(2.164800,1.559614)(2.169600,1.581588)(2.174400,1.582979)(2.179200,1.556603)(2.184000,1.530191)(2.188800,1.521498)(2.193600,1.532614)(2.198400,1.534089)(2.203200,1.511724)(2.208000,1.478934)(2.212800,1.486824)(2.217600,1.503572)(2.222400,1.497048)(2.227200,1.498578)(2.232000,1.490283)(2.236800,1.477700)(2.241600,1.445601)(2.246400,1.452662)(2.251200,1.463547)(2.256000,1.465324)(2.260800,1.451368)(2.265600,1.424552)(2.270400,1.420824)(2.275200,1.413236)(2.280000,1.390413)(2.284800,1.356907)(2.289600,1.369141)(2.294400,1.394361)(2.299200,1.383897)(2.304000,1.388526)(2.308800,1.390177)(2.313600,1.373780)(2.318400,1.358453)(2.323200,1.335860)(2.328000,1.329051)(2.332800,1.314489)(2.337600,1.295166)(2.342400,1.290506)(2.347200,1.262429)(2.352000,1.229748)(2.356800,1.206588)(2.361600,1.202284)(2.366400,1.199872)(2.371200,1.177029)(2.376000,1.169871)(2.380800,1.152712)(2.385600,1.126044)(2.390400,1.105226)(2.395200,1.062987)}; 
\end{axis}
\end{tikzpicture}

%% file: figures/imagenet/googlenet_val_loss.tex
\begin{tikzpicture}[scale=0.7]
\begin{axis}[
  legend cell align=left,
  height=0.95\textwidth,
  width=1.5\textwidth,
  xmin=0, xmax=2.4,
  xtick={0, 0.5, 1, 1.5, 2},
  ymax=7,
  legend pos=north east,
  scaled x ticks = true,
  xticklabel = {\pgfmathprintnumber{\tick}M}
]

\addplot[color=ta3gray,ultra thick,dashed] coordinates {(0.003960,6.364770)(0.007960,5.940980)(0.011960,5.635990)(0.015960,5.263830)(0.019960,4.985810)(0.023960,4.860850)(0.027960,4.735140)(0.031960,4.521310)(0.035960,4.349350)(0.039960,4.383320)(0.043960,4.169050)(0.047960,4.094170)(0.051960,3.962490)(0.055960,4.050420)(0.059960,3.920760)(0.063960,3.776130)(0.067960,3.879100)(0.071960,3.785860)(0.075960,3.731650)(0.079960,3.700120)(0.083960,3.615260)(0.087960,3.636360)(0.091960,3.541480)(0.095960,3.498980)(0.099960,3.480130)(0.103960,3.403390)(0.107960,3.443480)(0.111960,3.397110)(0.115960,3.334570)(0.119960,3.329600)(0.123960,3.262050)(0.127960,3.433700)(0.131960,3.275750)(0.135960,3.303130)(0.139960,3.328890)(0.143960,3.240120)(0.147960,3.292070)(0.151960,3.166230)(0.155960,3.178280)(0.159960,3.123490)(0.163960,3.236590)(0.167960,3.126160)(0.171960,3.128400)(0.175960,3.116480)(0.179960,3.141390)(0.183960,3.123100)(0.187960,3.117240)(0.191960,3.098200)(0.195960,3.037600)(0.199960,3.099400)(0.203960,3.122470)(0.207960,3.128900)(0.211960,3.047610)(0.215960,3.056560)(0.219960,2.956770)(0.223960,2.989040)(0.227960,3.048680)(0.231960,2.992680)(0.235960,3.011620)(0.239960,2.988190)(0.243960,2.936100)(0.247960,2.919840)(0.251960,2.932330)(0.255960,2.984970)(0.259960,2.992620)(0.263960,2.951220)(0.267960,2.910610)(0.271960,2.943960)(0.275960,2.956910)(0.279960,2.899940)(0.283960,2.909450)(0.287960,2.981670)(0.291960,2.988360)(0.295960,3.007950)(0.299960,2.861800)(0.303960,2.841930)(0.307960,2.908140)(0.311960,2.913270)(0.315960,2.852790)(0.319960,2.994690)(0.323960,2.866620)(0.327960,2.835090)(0.331960,2.832480)(0.335960,2.867710)(0.339960,2.822900)(0.343960,2.917940)(0.347960,2.848930)(0.351960,2.801860)(0.355960,2.832750)(0.359960,2.840170)(0.363960,2.775990)(0.367960,2.804890)(0.371960,2.787590)(0.375960,2.856740)(0.379960,2.792830)(0.383960,2.794150)(0.387960,2.757020)(0.391960,2.782710)(0.395960,2.767240)(0.399960,2.741500)(0.403960,2.763080)(0.407960,2.805890)(0.411960,2.817780)(0.415960,2.754680)(0.419960,2.765690)(0.423960,2.780020)(0.427960,2.824050)(0.431960,2.776970)(0.435960,2.724240)(0.439960,2.722720)(0.443960,2.751410)(0.447960,2.756250)(0.451960,2.854500)(0.455960,2.718100)(0.459960,2.795380)(0.463960,2.673550)(0.467960,2.736610)(0.471960,2.692090)(0.475960,2.729280)(0.479960,2.710810)(0.483960,2.713920)(0.487960,2.730110)(0.491960,2.694820)(0.495960,2.748790)(0.499960,2.741510)(0.503960,2.701330)(0.507960,2.647160)(0.511960,2.773010)(0.515960,2.756140)(0.519960,2.664340)(0.523960,2.708250)(0.527960,2.710470)(0.531960,2.778420)(0.535960,2.733750)(0.539960,2.672720)(0.543960,2.747490)(0.547960,2.818290)(0.551960,2.740640)(0.555960,2.624130)(0.559960,2.628650)(0.563960,2.708700)(0.567960,2.684150)(0.571960,2.662190)(0.575960,2.685770)(0.579960,2.651110)(0.583960,2.632410)(0.587960,2.657920)(0.591960,2.656120)(0.595960,2.672690)(0.599960,2.667140)(0.603960,2.716670)(0.607960,2.651320)(0.611960,2.643780)(0.615960,2.689770)(0.619960,2.609080)(0.623960,2.628280)(0.627960,2.657130)(0.631960,2.709610)(0.635960,2.660730)(0.639960,2.578820)(0.643960,2.704800)(0.647960,2.635970)(0.651960,2.609890)(0.655960,2.626510)(0.659960,2.640100)(0.663960,2.655290)(0.667960,2.606950)(0.671960,2.656170)(0.675960,2.591160)(0.679960,2.583230)(0.683960,2.644900)(0.687960,2.531230)(0.691960,2.604690)(0.695960,2.662780)(0.699960,2.642920)(0.703960,2.655620)(0.707960,2.647960)(0.711960,2.631290)(0.715960,2.584040)(0.719960,2.605540)(0.723960,2.686420)(0.727960,2.577610)(0.731960,2.595190)(0.735960,2.576970)(0.739960,2.554150)(0.743960,2.591690)(0.747960,2.585310)(0.751960,2.509840)(0.755960,2.558140)(0.759960,2.542550)(0.763960,2.578950)(0.767960,2.611260)(0.771960,2.538020)(0.775960,2.630770)(0.779960,2.560330)(0.783960,2.574560)(0.787960,2.583800)(0.791960,2.575720)(0.795960,2.555340)(0.799960,2.516580)(0.803960,2.590960)(0.807960,2.618000)(0.811960,2.543990)(0.815960,2.527650)(0.819960,2.541850)(0.823960,2.545340)(0.827960,2.500600)(0.831960,2.564360)(0.835960,2.536020)(0.839960,2.475730)(0.843960,2.605220)(0.847960,2.516910)(0.851960,2.607250)(0.855960,2.472750)(0.859960,2.529730)(0.863960,2.539960)(0.867960,2.478480)(0.871960,2.495220)(0.875960,2.520780)(0.879960,2.472800)(0.883960,2.523330)(0.887960,2.503790)(0.891960,2.477380)(0.895960,2.596480)(0.899960,2.493660)(0.903960,2.478490)(0.907960,2.508890)(0.911960,2.544540)(0.915960,2.471540)(0.919960,2.488070)(0.923960,2.480500)(0.927960,2.485320)(0.931960,2.459490)(0.935960,2.480930)(0.939960,2.481740)(0.943960,2.529750)(0.947960,2.507420)(0.951960,2.447550)(0.955960,2.514900)(0.959960,2.528390)(0.963960,2.473610)(0.967960,2.474120)(0.971960,2.518520)(0.975960,2.502820)(0.979960,2.471840)(0.983960,2.460760)(0.987960,2.529320)(0.991960,2.451790)(0.995960,2.480330)(0.999960,2.425440)(1.003960,2.425900)(1.007960,2.468980)(1.011960,2.489870)(1.015960,2.499260)(1.019960,2.466480)(1.023960,2.474510)(1.027960,2.464490)(1.031960,2.504460)(1.035960,2.442130)(1.039960,2.497540)(1.043960,2.431290)(1.047960,2.488150)(1.051960,2.448630)(1.055960,2.487970)(1.059960,2.428660)(1.063960,2.423830)(1.067960,2.449130)(1.071960,2.416100)(1.075960,2.425400)(1.079960,2.420720)(1.083960,2.457330)(1.087960,2.410750)(1.091960,2.419370)(1.095960,2.434460)(1.099960,2.367350)(1.103960,2.389030)(1.107960,2.471340)(1.111960,2.405870)(1.115960,2.367250)(1.119960,2.397940)(1.123960,2.373940)(1.127960,2.421710)(1.131960,2.387420)(1.135960,2.409480)(1.139960,2.366540)(1.143960,2.376720)(1.147960,2.376540)(1.151960,2.425000)(1.155960,2.368660)(1.159960,2.402000)(1.163960,2.404250)(1.167960,2.374070)(1.171960,2.444150)(1.175960,2.382880)(1.179960,2.381200)(1.183960,2.378300)(1.187960,2.362110)(1.191960,2.378490)(1.195960,2.389840)(1.199960,2.389320)(1.203960,2.343280)(1.207960,2.379870)(1.211960,2.412810)(1.215960,2.407540)(1.219960,2.436570)(1.223960,2.394600)(1.227960,2.378560)(1.231960,2.401160)(1.235960,2.352860)(1.239960,2.377340)(1.243960,2.367680)(1.247960,2.395200)(1.251960,2.372080)(1.255960,2.329980)(1.259960,2.350320)(1.263960,2.398280)(1.267960,2.306740)(1.271960,2.344410)(1.275960,2.320790)(1.279960,2.331980)(1.283960,2.347290)(1.287960,2.295200)(1.291960,2.299640)(1.295960,2.344950)(1.299960,2.340190)(1.303960,2.323980)(1.307960,2.346820)(1.311960,2.286920)(1.315960,2.351930)(1.319960,2.332700)(1.323960,2.345370)(1.327960,2.318340)(1.331960,2.313450)(1.335960,2.315270)(1.339960,2.309450)(1.343960,2.287070)(1.347960,2.269010)(1.351960,2.296510)(1.355960,2.292080)(1.359960,2.343480)(1.363960,2.338340)(1.367960,2.286090)(1.371960,2.311170)(1.375960,2.330060)(1.379960,2.267590)(1.383960,2.221880)(1.387960,2.266860)(1.391960,2.293400)(1.395960,2.335120)(1.399960,2.326010)(1.403960,2.307940)(1.407960,2.288080)(1.411960,2.291970)(1.415960,2.298930)(1.419960,2.383800)(1.423960,2.277480)(1.427960,2.301360)(1.431960,2.251990)(1.435960,2.276070)(1.439960,2.292510)(1.443960,2.266560)(1.447960,2.247840)(1.451960,2.234100)(1.455960,2.276630)(1.459960,2.229100)(1.463960,2.251890)(1.467960,2.230560)(1.471960,2.215650)(1.475960,2.252460)(1.479960,2.241490)(1.483960,2.246650)(1.487960,2.288050)(1.491960,2.232920)(1.495960,2.262620)(1.499960,2.217290)(1.503960,2.206190)(1.507960,2.268320)(1.511960,2.229350)(1.515960,2.162200)(1.519960,2.246270)(1.523960,2.309780)(1.527960,2.195440)(1.531960,2.257200)(1.535960,2.185290)(1.539960,2.211370)(1.543960,2.190060)(1.547960,2.167340)(1.551960,2.222130)(1.555960,2.226060)(1.559960,2.187980)(1.563960,2.201420)(1.567960,2.194610)(1.571960,2.189850)(1.575960,2.201120)(1.579960,2.213450)(1.583960,2.246820)(1.587960,2.227160)(1.591960,2.177130)(1.595960,2.199550)(1.599960,2.162390)(1.603960,2.199940)(1.607960,2.184250)(1.611960,2.178690)(1.615960,2.136920)(1.619960,2.185720)(1.623960,2.199650)(1.627960,2.160450)(1.631960,2.136310)(1.635960,2.155280)(1.639960,2.187360)(1.643960,2.180590)(1.647960,2.175290)(1.651960,2.209240)(1.655960,2.163800)(1.659960,2.133660)(1.663960,2.170130)(1.667960,2.159390)(1.671960,2.141500)(1.675960,2.182930)(1.679960,2.145010)(1.683960,2.125840)(1.687960,2.158650)(1.691960,2.098210)(1.695960,2.152580)(1.699960,2.180740)(1.703960,2.142500)(1.707960,2.104870)(1.711960,2.116870)(1.715960,2.184510)(1.719960,2.174520)(1.723960,2.107430)(1.727960,2.119810)(1.731960,2.108900)(1.735960,2.112290)(1.739960,2.148260)(1.743960,2.104420)(1.747960,2.051020)(1.751960,2.077800)(1.755960,2.136460)(1.759960,2.141230)(1.763960,2.076870)(1.767960,2.115740)(1.771960,2.089620)(1.775960,2.145600)(1.779960,2.064760)(1.783960,2.111350)(1.787960,2.062950)(1.791960,2.081000)(1.795960,2.096880)(1.799960,2.132360)(1.803960,2.071270)(1.807960,2.132020)(1.811960,2.056960)(1.815960,2.044670)(1.819960,2.034730)(1.823960,2.066440)(1.827960,2.044650)(1.831960,2.103120)(1.835960,2.068490)(1.839960,2.040940)(1.843960,2.084190)(1.847960,2.032910)(1.851960,2.063890)(1.855960,2.016980)(1.859960,2.036310)(1.863960,2.039010)(1.867960,2.030350)(1.871960,2.073330)(1.875960,2.060370)(1.879960,2.022170)(1.883960,2.053020)(1.887960,2.020320)(1.891960,2.008900)(1.895960,2.043550)(1.899960,2.035620)(1.903960,2.002640)(1.907960,2.023140)(1.911960,1.973200)(1.915960,1.980890)(1.919960,2.027870)(1.923960,2.036870)(1.927960,1.988850)(1.931960,2.013140)(1.935960,1.960650)(1.939960,1.974640)(1.943960,1.969020)(1.947960,1.979430)(1.951960,2.002070)(1.955960,1.998830)(1.959960,1.976330)(1.963960,2.007510)(1.967960,1.956860)(1.971960,1.970360)(1.975960,1.975680)(1.979960,1.982810)(1.983960,1.924000)(1.987960,1.921980)(1.991960,1.949810)(1.995960,1.982790)(1.999960,1.917790)(2.003960,1.917820)(2.007960,1.941630)(2.011960,1.937660)(2.015960,1.967620)(2.019960,1.923110)(2.023960,1.927820)(2.027960,1.950190)(2.031960,1.959080)(2.035960,1.989430)(2.039960,1.963100)(2.043960,1.914650)(2.047960,1.915830)(2.051960,1.907590)(2.055960,1.873020)(2.059960,1.935820)(2.063960,1.872960)(2.067960,1.858950)(2.071960,1.897060)(2.075960,1.885180)(2.079960,1.884680)(2.083960,1.914850)(2.087960,1.871700)(2.091960,1.874490)(2.095960,1.862950)(2.099960,1.888490)(2.103960,1.888990)(2.107960,1.844010)(2.111960,1.829990)(2.115960,1.870640)(2.119960,1.946250)(2.123960,1.868010)(2.127960,1.858150)(2.131960,1.857140)(2.135960,1.808720)(2.139960,1.869100)(2.143960,1.831800)(2.147960,1.812010)(2.151960,1.852760)(2.155960,1.829400)(2.159960,1.791900)(2.163960,1.873940)(2.167960,1.804340)(2.171960,1.827080)(2.175960,1.796640)(2.179960,1.775570)(2.183960,1.770550)(2.187960,1.791790)(2.191960,1.791980)(2.195960,1.804030)(2.199960,1.750110)(2.203960,1.754740)(2.207960,1.800360)(2.211960,1.764480)(2.215960,1.740630)(2.219960,1.757530)(2.223960,1.747940)(2.227960,1.719840)(2.231960,1.746680)(2.235960,1.728920)(2.239960,1.701430)(2.243960,1.715350)(2.247960,1.688970)(2.251960,1.709090)(2.255960,1.686840)(2.259960,1.694160)(2.263960,1.691300)(2.267960,1.686260)(2.271960,1.654630)(2.275960,1.698930)(2.279960,1.671000)(2.283960,1.665030)(2.287960,1.654260)(2.291960,1.661650)(2.295960,1.664000)(2.299960,1.615310)(2.303960,1.617280)(2.307960,1.625170)(2.311960,1.627970)(2.315960,1.630110)(2.319960,1.620050)(2.323960,1.599910)(2.327960,1.624320)(2.331960,1.571200)(2.335960,1.568590)(2.339960,1.575470)(2.343960,1.558250)(2.347960,1.538010)(2.351960,1.525130)(2.355960,1.512810)(2.359960,1.512680)(2.363960,1.489590)(2.367960,1.487970)(2.371960,1.487510)(2.375960,1.463820)(2.379960,1.445990)(2.383960,1.424780)(2.387960,1.405330)(2.391960,1.378520)(2.395960,1.343700)}; 
\addlegendentry{Reference}
\addplot[color=taskyblue,ultra thick,solid] coordinates {(0.003960,5.624070)(0.007960,5.237750)(0.011960,4.951190)(0.015960,4.679450)(0.019960,4.403620)(0.023960,4.288650)(0.027960,4.175540)(0.031960,3.997240)(0.035960,3.875480)(0.039960,3.828550)(0.043960,3.750530)(0.047960,3.570740)(0.051960,3.522030)(0.055960,3.510910)(0.059960,3.539210)(0.063960,3.332040)(0.067960,3.351040)(0.071960,3.343050)(0.075960,3.311760)(0.079960,3.204000)(0.083960,3.213600)(0.087960,3.146940)(0.091960,3.059370)(0.095960,3.069040)(0.099960,3.013780)(0.103960,3.013240)(0.107960,2.945030)(0.111960,2.936750)(0.115960,2.963270)(0.119960,2.956110)(0.123960,3.039970)(0.127960,2.906590)(0.131960,2.855540)(0.135960,2.921680)(0.139960,2.807050)(0.143960,2.806740)(0.147960,2.792090)(0.151960,2.896810)(0.155960,2.837280)(0.159960,2.815580)(0.163960,2.898010)(0.167960,2.822110)(0.171960,2.754750)(0.175960,2.741010)(0.179960,2.743030)(0.183960,2.735620)(0.187960,2.749110)(0.191960,2.804520)(0.195960,2.702500)(0.199960,2.689860)(0.203960,2.661900)(0.207960,2.729080)(0.211960,2.617750)(0.215960,2.633630)(0.219960,2.615960)(0.223960,2.692650)(0.227960,2.636530)(0.231960,2.678190)(0.235960,2.656290)(0.239960,2.668750)(0.243960,2.569020)(0.247960,2.630280)(0.251960,2.620080)(0.255960,2.648330)(0.259960,2.576270)(0.263960,2.586110)(0.267960,2.549590)(0.271960,2.518280)(0.275960,2.546560)(0.279960,2.607610)(0.283960,2.600650)(0.287960,2.607460)(0.291960,2.584990)(0.295960,2.546990)(0.299960,2.504190)(0.303960,2.508640)(0.307960,2.531930)(0.311960,2.568380)(0.315960,2.549270)(0.319960,2.515520)(0.323960,2.497690)(0.327960,2.524870)(0.331960,2.489760)(0.335960,2.528410)(0.339960,2.480560)(0.343960,2.544540)(0.347960,2.472870)(0.351960,2.505240)(0.355960,2.451610)(0.359960,2.542440)(0.363960,2.518780)(0.367960,2.483780)(0.371960,2.475280)(0.375960,2.450040)(0.379960,2.450630)(0.383960,2.532690)(0.387960,2.472160)(0.391960,2.446970)(0.395960,2.519080)(0.399960,2.533630)(0.403960,2.478200)(0.407960,2.504700)(0.411960,2.433540)(0.415960,2.459080)(0.419960,2.452440)(0.423960,2.434430)(0.427960,2.451650)(0.431960,2.472970)(0.435960,2.427560)(0.439960,2.457400)(0.443960,2.436290)(0.447960,2.434170)(0.451960,2.517490)(0.455960,2.462240)(0.459960,2.453920)(0.463960,2.436610)(0.467960,2.387000)(0.471960,2.512560)(0.475960,2.363830)(0.479960,2.395510)(0.483960,2.388650)(0.487960,2.408030)(0.491960,2.369710)(0.495960,2.404020)(0.499960,2.407090)(0.503960,2.412850)(0.507960,2.381260)(0.511960,2.395250)(0.515960,2.421120)(0.519960,2.396410)(0.523960,2.371550)(0.527960,2.398410)(0.531960,2.404070)(0.535960,2.357530)(0.539960,2.394240)(0.543960,2.418070)(0.547960,2.382610)(0.551960,2.395340)(0.555960,2.335210)(0.559960,2.329610)(0.563960,2.400830)(0.567960,2.375880)(0.571960,2.337510)(0.575960,2.356530)(0.579960,2.378790)(0.583960,2.371570)(0.587960,2.406960)(0.591960,2.392620)(0.595960,2.337200)(0.599960,2.386000)(0.603960,2.384260)(0.607960,2.382320)(0.611960,2.346680)(0.615960,2.296960)(0.619960,2.322720)(0.623960,2.333690)(0.627960,2.333510)(0.631960,2.369470)(0.635960,2.352870)(0.639960,2.359180)(0.643960,2.350230)(0.647960,2.397900)(0.651960,2.362730)(0.655960,2.330880)(0.659960,2.293920)(0.663960,2.277010)(0.667960,2.308350)(0.671960,2.273490)(0.675960,2.328370)(0.679960,2.329590)(0.683960,2.397140)(0.687960,2.306410)(0.691960,2.352360)(0.695960,2.276740)(0.699960,2.327980)(0.703960,2.313120)(0.707960,2.358840)(0.711960,2.399870)(0.715960,2.318910)(0.719960,2.392750)(0.723960,2.346060)(0.727960,2.270270)(0.731960,2.317830)(0.735960,2.297630)(0.739960,2.295900)(0.743960,2.282030)(0.747960,2.229650)(0.751960,2.244750)(0.755960,2.283410)(0.759960,2.312730)(0.763960,2.372180)(0.767960,2.268090)(0.771960,2.349820)(0.775960,2.288830)(0.779960,2.275590)(0.783960,2.268440)(0.787960,2.281470)(0.791960,2.190330)(0.795960,2.296970)(0.799960,2.223820)(0.803960,2.295820)(0.807960,2.275270)(0.811960,2.247610)(0.815960,2.252190)(0.819960,2.325250)(0.823960,2.262890)(0.827960,2.251950)(0.831960,2.270630)(0.835960,2.288920)(0.839960,2.268720)(0.843960,2.283570)(0.847960,2.243710)(0.851960,2.269310)(0.855960,2.271190)(0.859960,2.234660)(0.863960,2.211660)(0.867960,2.201350)(0.871960,2.226290)(0.875960,2.270230)(0.879960,2.268980)(0.883960,2.240910)(0.887960,2.237830)(0.891960,2.216020)(0.895960,2.312140)(0.899960,2.304320)(0.903960,2.286160)(0.907960,2.298020)(0.911960,2.249970)(0.915960,2.211190)(0.919960,2.320240)(0.923960,2.272910)(0.927960,2.257730)(0.931960,2.224820)(0.935960,2.267380)(0.939960,2.179360)(0.943960,2.180340)(0.947960,2.251900)(0.951960,2.258060)(0.955960,2.190730)(0.959960,2.217890)(0.963960,2.256470)(0.967960,2.251000)(0.971960,2.186770)(0.975960,2.241660)(0.979960,2.226180)(0.983960,2.175980)(0.987960,2.238440)(0.991960,2.217560)(0.995960,2.222960)(0.999960,2.232930)(1.003960,2.168850)(1.007960,2.221560)(1.011960,2.256670)(1.015960,2.179340)(1.019960,2.201110)(1.023960,2.192190)(1.027960,2.198410)(1.031960,2.213140)(1.035960,2.246780)(1.039960,2.194370)(1.043960,2.205510)(1.047960,2.181290)(1.051960,2.160700)(1.055960,2.178610)(1.059960,2.147980)(1.063960,2.155680)(1.067960,2.239610)(1.071960,2.198930)(1.075960,2.175830)(1.079960,2.138100)(1.083960,2.178740)(1.087960,2.128760)(1.091960,2.161560)(1.095960,2.182370)(1.099960,2.181810)(1.103960,2.156710)(1.107960,2.198170)(1.111960,2.168890)(1.115960,2.174530)(1.119960,2.153970)(1.123960,2.176740)(1.127960,2.190270)(1.131960,2.169930)(1.135960,2.159620)(1.139960,2.113880)(1.143960,2.148510)(1.147960,2.204400)(1.151960,2.158990)(1.155960,2.139520)(1.159960,2.198460)(1.163960,2.162280)(1.167960,2.157810)(1.171960,2.178350)(1.175960,2.276580)(1.179960,2.155100)(1.183960,2.156680)(1.187960,2.201030)(1.191960,2.163450)(1.195960,2.136210)(1.199960,2.124950)(1.203960,2.137780)(1.207960,2.133030)(1.211960,2.161800)(1.215960,2.147430)(1.219960,2.146090)(1.223960,2.119410)(1.227960,2.075310)(1.231960,2.154600)(1.235960,2.125650)(1.239960,2.151100)(1.243960,2.100900)(1.247960,2.092290)(1.251960,2.158780)(1.255960,2.090750)(1.259960,2.158920)(1.263960,2.118850)(1.267960,2.140890)(1.271960,2.158980)(1.275960,2.159580)(1.279960,2.114790)(1.283960,2.089780)(1.287960,2.082870)(1.291960,2.075820)(1.295960,2.151120)(1.299960,2.146270)(1.303960,2.090190)(1.307960,2.036480)(1.311960,2.111610)(1.315960,2.151070)(1.319960,2.103360)(1.323960,2.118720)(1.327960,2.108930)(1.331960,2.089790)(1.335960,2.064530)(1.339960,2.061190)(1.343960,2.074720)(1.347960,2.103570)(1.351960,2.093830)(1.355960,2.076650)(1.359960,2.068540)(1.363960,2.084170)(1.367960,2.058930)(1.371960,2.064780)(1.375960,2.094990)(1.379960,2.065360)(1.383960,2.078490)(1.387960,2.064730)(1.391960,2.022650)(1.395960,2.057330)(1.399960,2.069640)(1.403960,2.071300)(1.407960,2.049390)(1.411960,2.083600)(1.415960,2.072660)(1.419960,2.126260)(1.423960,2.024450)(1.427960,2.101270)(1.431960,2.078670)(1.435960,2.021230)(1.439960,2.037750)(1.443960,2.044810)(1.447960,2.067420)(1.451960,2.051070)(1.455960,2.071260)(1.459960,2.061380)(1.463960,2.048610)(1.467960,2.015460)(1.471960,2.052850)(1.475960,2.042520)(1.479960,2.025010)(1.483960,1.999570)(1.487960,2.017510)(1.491960,2.078260)(1.495960,2.071020)(1.499960,2.063260)(1.503960,2.042850)(1.507960,2.019790)(1.511960,2.084230)(1.515960,1.981130)(1.519960,2.069470)(1.523960,2.055180)(1.527960,2.035580)(1.531960,2.071720)(1.535960,2.035030)(1.539960,2.021720)(1.543960,2.077700)(1.547960,2.017790)(1.551960,2.023310)(1.555960,1.996430)(1.559960,2.009920)(1.563960,1.992750)(1.567960,2.014370)(1.571960,1.971970)(1.575960,2.019160)(1.579960,1.994590)(1.583960,2.080580)(1.587960,1.988940)(1.591960,1.987360)(1.595960,2.012110)(1.599960,1.977790)(1.603960,2.030060)(1.607960,2.009310)(1.611960,1.951860)(1.615960,1.991320)(1.619960,2.005840)(1.623960,2.016770)(1.627960,2.048890)(1.631960,1.981220)(1.635960,1.956940)(1.639960,1.995210)(1.643960,1.985780)(1.647960,2.013440)(1.651960,2.034360)(1.655960,1.946200)(1.659960,1.967090)(1.663960,1.957880)(1.667960,1.983740)(1.671960,1.937700)(1.675960,1.967290)(1.679960,1.950150)(1.683960,1.989340)(1.687960,1.962210)(1.691960,1.952860)(1.695960,1.952510)(1.699960,1.972920)(1.703960,2.032420)(1.707960,1.956280)(1.711960,1.959370)(1.715960,2.021470)(1.719960,1.969760)(1.723960,2.015100)(1.727960,1.931550)(1.731960,1.997750)(1.735960,1.945000)(1.739960,1.967290)(1.743960,1.951910)(1.747960,1.906610)(1.751960,1.939710)(1.755960,1.916880)(1.759960,1.980040)(1.763960,1.936770)(1.767960,1.951690)(1.771960,1.944310)(1.775960,1.937700)(1.779960,1.898310)(1.783960,1.968180)(1.787960,1.905160)(1.791960,1.891750)(1.795960,1.890420)(1.799960,1.942970)(1.803960,1.907990)(1.807960,1.913880)(1.811960,1.868140)(1.815960,1.907230)(1.819960,1.909240)(1.823960,1.891020)(1.827960,1.888870)(1.831960,1.894770)(1.835960,1.942500)(1.839960,1.921680)(1.843960,1.888380)(1.847960,1.874770)(1.851960,1.877120)(1.855960,1.870890)(1.859960,1.896950)(1.863960,1.927080)(1.867960,1.838650)(1.871960,1.928280)(1.875960,1.882280)(1.879960,1.902550)(1.883960,1.880710)(1.887960,1.868590)(1.891960,1.855450)(1.895960,1.871540)(1.899960,1.878660)(1.903960,1.906260)(1.907960,1.870390)(1.911960,1.871390)(1.915960,1.834220)(1.919960,1.823160)(1.923960,1.852310)(1.927960,1.869510)(1.931960,1.879890)(1.935960,1.821000)(1.939960,1.834910)(1.943960,1.841190)(1.947960,1.855450)(1.951960,1.836030)(1.955960,1.873360)(1.959960,1.819570)(1.963960,1.865820)(1.967960,1.865020)(1.971960,1.831560)(1.975960,1.836750)(1.979960,1.818560)(1.983960,1.828830)(1.987960,1.820030)(1.991960,1.862440)(1.995960,1.797140)(1.999960,1.821380)(2.003960,1.818850)(2.007960,1.798960)(2.011960,1.804250)(2.015960,1.830940)(2.019960,1.793190)(2.023960,1.751300)(2.027960,1.801890)(2.031960,1.782460)(2.035960,1.779330)(2.039960,1.793920)(2.043960,1.807400)(2.047960,1.791480)(2.051960,1.798220)(2.055960,1.763760)(2.059960,1.845300)(2.063960,1.773880)(2.067960,1.771330)(2.071960,1.790000)(2.075960,1.760690)(2.079960,1.747720)(2.083960,1.775740)(2.087960,1.730860)(2.091960,1.735960)(2.095960,1.747820)(2.099960,1.775070)(2.103960,1.743470)(2.107960,1.738150)(2.111960,1.766510)(2.115960,1.755750)(2.119960,1.740710)(2.123960,1.724620)(2.127960,1.743020)(2.131960,1.729530)(2.135960,1.718000)(2.139960,1.731930)(2.143960,1.713380)(2.147960,1.732340)(2.151960,1.753930)(2.155960,1.756390)(2.159960,1.684110)(2.163960,1.752440)(2.167960,1.712270)(2.171960,1.719360)(2.175960,1.709940)(2.179960,1.669010)(2.183960,1.667060)(2.187960,1.695680)(2.191960,1.696690)(2.195960,1.718740)(2.199960,1.671240)(2.203960,1.691700)(2.207960,1.696640)(2.211960,1.666930)(2.215960,1.655650)(2.219960,1.668630)(2.223960,1.640650)(2.227960,1.624300)(2.231960,1.644980)(2.235960,1.648620)(2.239960,1.617990)(2.243960,1.612750)(2.247960,1.635740)(2.251960,1.650600)(2.255960,1.621600)(2.259960,1.597520)(2.263960,1.608600)(2.267960,1.600230)(2.271960,1.589080)(2.275960,1.637590)(2.279960,1.614730)(2.283960,1.580630)(2.287960,1.593750)(2.291960,1.586060)(2.295960,1.564830)(2.299960,1.573210)(2.303960,1.565800)(2.307960,1.570690)(2.311960,1.553390)(2.315960,1.539710)(2.319960,1.519370)(2.323960,1.577890)(2.327960,1.545000)(2.331960,1.511720)(2.335960,1.517480)(2.339960,1.488590)(2.343960,1.505690)(2.347960,1.500740)(2.351960,1.481870)(2.355960,1.469430)(2.359960,1.467050)(2.363960,1.446180)(2.367960,1.437510)(2.371960,1.455820)(2.375960,1.435600)(2.379960,1.406200)(2.383960,1.381250)(2.387960,1.380590)(2.391960,1.340160)(2.395960,1.327520)}; 
\addlegendentry{\kmeans (ours)}
\addplot[color=ta3scarletred,ultra thick,dashed] coordinates {(0.003960,5.638340)(0.007960,5.201820)(0.011960,4.910290)(0.015960,4.641020)(0.019960,4.441740)(0.023960,4.295490)(0.027960,4.168020)(0.031960,4.007060)(0.035960,3.843220)(0.039960,3.746150)(0.043960,3.774320)(0.047960,3.592370)(0.051960,3.460490)(0.055960,3.566220)(0.059960,3.456980)(0.063960,3.306760)(0.067960,3.363400)(0.071960,3.347270)(0.075960,3.206970)(0.079960,3.191750)(0.083960,3.174790)(0.087960,3.205160)(0.091960,3.072530)(0.095960,3.085540)(0.099960,3.023140)(0.103960,2.969860)(0.107960,2.940640)(0.111960,2.989610)(0.115960,3.012200)(0.119960,2.986830)(0.123960,3.094630)(0.127960,2.908360)(0.131960,2.895680)(0.135960,2.871210)(0.139960,2.918420)(0.143960,2.796730)(0.147960,2.841480)(0.151960,2.837660)(0.155960,2.847390)(0.159960,2.785030)(0.163960,2.793250)(0.167960,2.774250)(0.171960,2.691680)(0.175960,2.695500)(0.179960,2.706490)(0.183960,2.702600)(0.187960,2.735490)(0.191960,2.747740)(0.195960,2.717050)(0.199960,2.714190)(0.203960,2.665940)(0.207960,2.696450)(0.211960,2.640210)(0.215960,2.687580)(0.219960,2.584610)(0.223960,2.665260)(0.227960,2.617700)(0.231960,2.694660)(0.235960,2.567110)(0.239960,2.686670)(0.243960,2.563970)(0.247960,2.612540)(0.251960,2.554970)(0.255960,2.548950)(0.259960,2.588570)(0.263960,2.636750)(0.267960,2.590420)(0.271960,2.542580)(0.275960,2.489010)(0.279960,2.557000)(0.283960,2.540020)(0.287960,2.550270)(0.291960,2.588060)(0.295960,2.579690)(0.299960,2.480250)(0.303960,2.490990)(0.307960,2.533970)(0.311960,2.608810)(0.315960,2.568590)(0.319960,2.497170)(0.323960,2.452190)(0.327960,2.550850)(0.331960,2.560840)(0.335960,2.481500)(0.339960,2.505060)(0.343960,2.452700)(0.347960,2.540500)(0.351960,2.431980)(0.355960,2.454140)(0.359960,2.469840)(0.363960,2.477020)(0.367960,2.512260)(0.371960,2.431350)(0.375960,2.412710)(0.379960,2.408710)(0.383960,2.419010)(0.387960,2.442960)(0.391960,2.395040)(0.395960,2.414380)(0.399960,2.414810)(0.403960,2.447860)(0.407960,2.457500)(0.411960,2.422650)(0.415960,2.477500)(0.419960,2.464370)(0.423960,2.464890)(0.427960,2.466420)(0.431960,2.357080)(0.435960,2.451360)(0.439960,2.386040)(0.443960,2.460880)(0.447960,2.349270)(0.451960,2.413320)(0.455960,2.433730)(0.459960,2.396380)(0.463960,2.376240)(0.467960,2.378490)(0.471960,2.364550)(0.475960,2.430820)(0.479960,2.363410)(0.483960,2.366310)(0.487960,2.385680)(0.491960,2.341000)(0.495960,2.403690)(0.499960,2.489940)(0.503960,2.394130)(0.507960,2.306180)(0.511960,2.403510)(0.515960,2.356640)(0.519960,2.349230)(0.523960,2.353520)(0.527960,2.412790)(0.531960,2.362840)(0.535960,2.448310)(0.539960,2.429040)(0.543960,2.446050)(0.547960,2.336850)(0.551960,2.417700)(0.555960,2.356710)(0.559960,2.288030)(0.563960,2.322030)(0.567960,2.396290)(0.571960,2.323430)(0.575960,2.314810)(0.579960,2.316680)(0.583960,2.275080)(0.587960,2.347760)(0.591960,2.327250)(0.595960,2.300820)(0.599960,2.345570)(0.603960,2.311400)(0.607960,2.342070)(0.611960,2.288220)(0.615960,2.318060)(0.619960,2.308900)(0.623960,2.349590)(0.627960,2.325040)(0.631960,2.331690)(0.635960,2.383410)(0.639960,2.277200)(0.643960,2.354280)(0.647960,2.352680)(0.651960,2.285580)(0.655960,2.309840)(0.659960,2.315810)(0.663960,2.264060)(0.667960,2.280770)(0.671960,2.315630)(0.675960,2.283120)(0.679960,2.275910)(0.683960,2.328640)(0.687960,2.251670)(0.691960,2.266460)(0.695960,2.248990)(0.699960,2.261310)(0.703960,2.356840)(0.707960,2.328330)(0.711960,2.295880)(0.715960,2.269840)(0.719960,2.265400)(0.723960,2.234470)(0.727960,2.267710)(0.731960,2.268090)(0.735960,2.299700)(0.739960,2.285760)(0.743960,2.245350)(0.747960,2.301310)(0.751960,2.281300)(0.755960,2.201700)(0.759960,2.262270)(0.763960,2.281100)(0.767960,2.263770)(0.771960,2.263620)(0.775960,2.264960)(0.779960,2.224150)(0.783960,2.274240)(0.787960,2.282540)(0.791960,2.229250)(0.795960,2.222360)(0.799960,2.218090)(0.803960,2.275520)(0.807960,2.231140)(0.811960,2.249810)(0.815960,2.229450)(0.819960,2.340850)(0.823960,2.267250)(0.827960,2.244670)(0.831960,2.267650)(0.835960,2.277790)(0.839960,2.246380)(0.843960,2.261790)(0.847960,2.219500)(0.851960,2.247490)(0.855960,2.191020)(0.859960,2.248000)(0.863960,2.192490)(0.867960,2.198460)(0.871960,2.257290)(0.875960,2.183870)(0.879960,2.233540)(0.883960,2.213320)(0.887960,2.253850)(0.891960,2.199250)(0.895960,2.283470)(0.899960,2.209990)(0.903960,2.216270)(0.907960,2.165400)(0.911960,2.242920)(0.915960,2.210160)(0.919960,2.251180)(0.923960,2.168990)(0.927960,2.256070)(0.931960,2.244530)(0.935960,2.151830)(0.939960,2.241500)(0.943960,2.155060)(0.947960,2.210360)(0.951960,2.143950)(0.955960,2.198110)(0.959960,2.161880)(0.963960,2.156010)(0.967960,2.239150)(0.971960,2.202730)(0.975960,2.155070)(0.979960,2.150430)(0.983960,2.152100)(0.987960,2.213200)(0.991960,2.190540)(0.995960,2.289170)(0.999960,2.165950)(1.003960,2.152980)(1.007960,2.189480)(1.011960,2.143870)(1.015960,2.165920)(1.019960,2.166680)(1.023960,2.124860)(1.027960,2.192570)(1.031960,2.196750)(1.035960,2.159970)(1.039960,2.194640)(1.043960,2.117620)(1.047960,2.125830)(1.051960,2.182780)(1.055960,2.174710)(1.059960,2.174410)(1.063960,2.161870)(1.067960,2.123140)(1.071960,2.215270)(1.075960,2.124320)(1.079960,2.122050)(1.083960,2.188480)(1.087960,2.204070)(1.091960,2.146310)(1.095960,2.198690)(1.099960,2.160820)(1.103960,2.123090)(1.107960,2.145280)(1.111960,2.125850)(1.115960,2.158480)(1.119960,2.162140)(1.123960,2.103040)(1.127960,2.149010)(1.131960,2.117280)(1.135960,2.154680)(1.139960,2.103390)(1.143960,2.125150)(1.147960,2.180530)(1.151960,2.148030)(1.155960,2.097090)(1.159960,2.066310)(1.163960,2.192870)(1.167960,2.111070)(1.171960,2.117020)(1.175960,2.090350)(1.179960,2.105150)(1.183960,2.078530)(1.187960,2.122810)(1.191960,2.168050)(1.195960,2.074380)(1.199960,2.100180)(1.203960,2.107740)(1.207960,2.114710)(1.211960,2.158260)(1.215960,2.123000)(1.219960,2.156310)(1.223960,2.111780)(1.227960,2.065710)(1.231960,2.232650)(1.235960,2.116610)(1.239960,2.080510)(1.243960,2.060030)(1.247960,2.119060)(1.251960,2.208420)(1.255960,2.098420)(1.259960,2.143040)(1.263960,2.152550)(1.267960,2.089380)(1.271960,2.136950)(1.275960,2.064830)(1.279960,2.066830)(1.283960,2.104680)(1.287960,2.071600)(1.291960,2.066760)(1.295960,2.079080)(1.299960,2.106960)(1.303960,2.104340)(1.307960,2.037150)(1.311960,2.093300)(1.315960,2.061840)(1.319960,2.087320)(1.323960,2.119480)(1.327960,2.032470)(1.331960,2.086250)(1.335960,2.027880)(1.339960,2.031050)(1.343960,2.046940)(1.347960,2.094470)(1.351960,2.051380)(1.355960,2.041960)(1.359960,2.148550)(1.363960,2.076570)(1.367960,2.026190)(1.371960,2.031980)(1.375960,2.028340)(1.379960,2.019440)(1.383960,2.075750)(1.387960,2.044390)(1.391960,2.051720)(1.395960,2.029870)(1.399960,2.048540)(1.403960,2.043540)(1.407960,2.039200)(1.411960,2.061230)(1.415960,2.085760)(1.419960,2.053680)(1.423960,2.007190)(1.427960,2.074910)(1.431960,2.072410)(1.435960,2.005680)(1.439960,2.097150)(1.443960,2.010800)(1.447960,2.050460)(1.451960,2.088450)(1.455960,2.076330)(1.459960,2.012560)(1.463960,2.019910)(1.467960,2.010290)(1.471960,2.079890)(1.475960,2.048690)(1.479960,2.053830)(1.483960,2.038820)(1.487960,2.041270)(1.491960,1.978070)(1.495960,1.994890)(1.499960,2.033030)(1.503960,1.949110)(1.507960,2.007900)(1.511960,2.027430)(1.515960,1.982950)(1.519960,2.032450)(1.523960,2.043460)(1.527960,2.053250)(1.531960,2.038290)(1.535960,1.974610)(1.539960,1.982020)(1.543960,1.953640)(1.547960,1.992960)(1.551960,2.035410)(1.555960,2.002200)(1.559960,1.987490)(1.563960,2.003470)(1.567960,2.020170)(1.571960,1.966480)(1.575960,1.997480)(1.579960,1.992210)(1.583960,2.056690)(1.587960,1.906780)(1.591960,2.032380)(1.595960,1.977710)(1.599960,1.998620)(1.603960,1.976300)(1.607960,1.999080)(1.611960,1.981490)(1.615960,1.964000)(1.619960,1.937470)(1.623960,1.980010)(1.627960,1.958070)(1.631960,1.942500)(1.635960,1.932230)(1.639960,1.917550)(1.643960,2.017770)(1.647960,1.954690)(1.651960,1.965830)(1.655960,1.993730)(1.659960,1.943510)(1.663960,1.908990)(1.667960,1.916730)(1.671960,1.938470)(1.675960,1.943190)(1.679960,1.935110)(1.683960,1.929360)(1.687960,1.933900)(1.691960,1.918880)(1.695960,1.994400)(1.699960,1.958870)(1.703960,1.948570)(1.707960,1.926720)(1.711960,1.899110)(1.715960,1.955690)(1.719960,1.921690)(1.723960,1.905410)(1.727960,1.952210)(1.731960,1.945250)(1.735960,1.947450)(1.739960,1.980030)(1.743960,1.895070)(1.747960,1.870460)(1.751960,1.931450)(1.755960,1.972090)(1.759960,1.982040)(1.763960,1.956020)(1.767960,1.935580)(1.771960,1.901180)(1.775960,1.944090)(1.779960,1.912550)(1.783960,1.917520)(1.787960,1.905100)(1.791960,1.892500)(1.795960,1.882240)(1.799960,1.947900)(1.803960,1.929970)(1.807960,1.929580)(1.811960,1.851660)(1.815960,1.849340)(1.819960,1.918020)(1.823960,1.870930)(1.827960,1.941150)(1.831960,1.930320)(1.835960,1.957450)(1.839960,1.878700)(1.843960,1.904990)(1.847960,1.905420)(1.851960,1.868690)(1.855960,1.875220)(1.859960,1.846560)(1.863960,1.925140)(1.867960,1.845080)(1.871960,1.890360)(1.875960,1.927800)(1.879960,1.878850)(1.883960,1.876410)(1.887960,1.898690)(1.891960,1.866540)(1.895960,1.906350)(1.899960,1.858470)(1.903960,1.859630)(1.907960,1.855050)(1.911960,1.837950)(1.915960,1.870080)(1.919960,1.803420)(1.923960,1.828780)(1.927960,1.815960)(1.931960,1.892610)(1.935960,1.857350)(1.939960,1.837100)(1.943960,1.850380)(1.947960,1.835930)(1.951960,1.798330)(1.955960,1.820540)(1.959960,1.836030)(1.963960,1.809660)(1.967960,1.835190)(1.971960,1.862810)(1.975960,1.802360)(1.979960,1.820450)(1.983960,1.777450)(1.987960,1.758210)(1.991960,1.794990)(1.995960,1.829960)(1.999960,1.773330)(2.003960,1.803870)(2.007960,1.838870)(2.011960,1.792870)(2.015960,1.824830)(2.019960,1.792850)(2.023960,1.769520)(2.027960,1.794600)(2.031960,1.775710)(2.035960,1.781350)(2.039960,1.762010)(2.043960,1.801930)(2.047960,1.779270)(2.051960,1.776970)(2.055960,1.769350)(2.059960,1.785780)(2.063960,1.759640)(2.067960,1.723260)(2.071960,1.759790)(2.075960,1.736330)(2.079960,1.761890)(2.083960,1.750900)(2.087960,1.804510)(2.091960,1.745890)(2.095960,1.728650)(2.099960,1.754280)(2.103960,1.740580)(2.107960,1.741860)(2.111960,1.706650)(2.115960,1.743410)(2.119960,1.744370)(2.123960,1.755870)(2.127960,1.715420)(2.131960,1.746060)(2.135960,1.725050)(2.139960,1.774640)(2.143960,1.697050)(2.147960,1.707010)(2.151960,1.706770)(2.155960,1.709130)(2.159960,1.661950)(2.163960,1.772520)(2.167960,1.678590)(2.171960,1.740730)(2.175960,1.710320)(2.179960,1.677320)(2.183960,1.686790)(2.187960,1.677190)(2.191960,1.668900)(2.195960,1.728020)(2.199960,1.660910)(2.203960,1.725080)(2.207960,1.699760)(2.211960,1.643910)(2.215960,1.640260)(2.219960,1.651750)(2.223960,1.655700)(2.227960,1.634210)(2.231960,1.624560)(2.235960,1.663100)(2.239960,1.641630)(2.243960,1.624510)(2.247960,1.615760)(2.251960,1.634170)(2.255960,1.620390)(2.259960,1.619450)(2.263960,1.596580)(2.267960,1.609240)(2.271960,1.599330)(2.275960,1.606090)(2.279960,1.571590)(2.283960,1.578790)(2.287960,1.571320)(2.291960,1.596690)(2.295960,1.556830)(2.299960,1.569450)(2.303960,1.565550)(2.307960,1.585830)(2.311960,1.539120)(2.315960,1.520020)(2.319960,1.525140)(2.323960,1.548110)(2.327960,1.532690)(2.331960,1.549350)(2.335960,1.501510)(2.339960,1.503250)(2.343960,1.482790)(2.347960,1.497850)(2.351960,1.493150)(2.355960,1.465040)(2.359960,1.459860)(2.363960,1.452630)(2.367960,1.433440)(2.371960,1.433210)(2.375960,1.442250)(2.379960,1.423650)(2.383960,1.396030)(2.387960,1.381920)(2.391960,1.359360)(2.395960,1.322620)}; 
\addlegendentry{\kmeans, single loss (ours)}
\end{axis}
\end{tikzpicture}

%% file: iclr2016_initialization.bbl
\begin{thebibliography}{20}
\providecommand{\natexlab}[1]{#1}
\providecommand{\url}[1]{\texttt{#1}}
\expandafter\ifx\csname urlstyle\endcsname\relax
  \providecommand{\doi}[1]{doi: #1}\else
  \providecommand{\doi}{doi: \begingroup \urlstyle{rm}\Url}\fi

\bibitem[Agrawal et~al.(2015)Agrawal, Carreira, and Malik]{agrawal2015learning}
Agrawal, Pulkit, Carreira, Joao, and Malik, Jitendra.
\newblock Learning to see by moving.
\newblock \emph{ICCV}, 2015.

\bibitem[Bradley(2010)]{bradley2010learning}
Bradley, David~M.
\newblock Learning in modular systems.
\newblock Technical report, DTIC Document, 2010.

\bibitem[Coates \& Ng(2012)Coates and Ng]{coates2012learning}
Coates, Adam and Ng, Andrew~Y.
\newblock Learning feature representations with k-means.
\newblock In \emph{Neural Networks: Tricks of the Trade}, pp.\  561--580.
  Springer, 2012.

\bibitem[Doersch et~al.(2015)Doersch, Gupta, and
  Efros]{doersch2015unsupervised}
Doersch, Carl, Gupta, Abhinav, and Efros, Alexei~A.
\newblock Unsupervised visual representation learning by context prediction.
\newblock \emph{ICCV}, 2015.

\bibitem[Everingham et~al.(2014)Everingham, Eslami, Van~Gool, Williams, Winn,
  and Zisserman]{everingham2014pascal}
Everingham, Mark, Eslami, SM~Ali, Van~Gool, Luc, Williams, Christopher~KI,
  Winn, John, and Zisserman, Andrew.
\newblock The {P}ascal {V}isual {O}bject {C}lasses challenge: A retrospective.
\newblock \emph{IJCV}, 111\penalty0 (1):\penalty0 98--136, 2014.

\bibitem[Girshick(2015)]{fastrcnn}
Girshick, Ross.
\newblock Fast {R-CNN}.
\newblock \emph{ICCV}, 2015.

\bibitem[Glorot \& Bengio(2010)Glorot and Bengio]{glorot2010understanding}
Glorot, Xavier and Bengio, Yoshua.
\newblock Understanding the difficulty of training deep feedforward neural
  networks.
\newblock In \emph{AISTATS}, pp.\  249--256, 2010.

\bibitem[He et~al.(2015)He, Zhang, Ren, and Sun]{he2015delving}
He, Kaiming, Zhang, Xiangyu, Ren, Shaoqing, and Sun, Jian.
\newblock Delving deep into rectifiers: Surpassing human-level performance on
  {ImageNet} classification.
\newblock In \emph{ICCV}, 2015.

\bibitem[Ioffe \& Szegedy(2015)Ioffe and Szegedy]{ioffe2015batch}
Ioffe, Sergey and Szegedy, Christian.
\newblock Batch normalization: Accelerating deep network training by reducing
  internal covariate shift.
\newblock In \emph{ICML}, 2015.

\bibitem[Jia et~al.(2014)Jia, Shelhamer, Donahue, Karayev, Long, Girshick,
  Guadarrama, and Darrell]{caffe}
Jia, Yangqing, Shelhamer, Evan, Donahue, Jeff, Karayev, Sergey, Long, Jonathan,
  Girshick, Ross~B., Guadarrama, Sergio, and Darrell, Trevor.
\newblock Caffe: Convolutional architecture for fast feature embedding.
\newblock In \emph{ACM Multimedia, {MM}}, 2014.

\bibitem[Kingma \& Ba(2015)Kingma and Ba]{kingma2014adam}
Kingma, Diederik and Ba, Jimmy.
\newblock Adam: A method for stochastic optimization.
\newblock \emph{ICLR}, 2015.

\bibitem[Krizhevsky et~al.(2012)Krizhevsky, Sutskever, and
  Hinton]{krizhevsky2012imagenet}
Krizhevsky, Alex, Sutskever, Ilya, and Hinton, Geoffrey~E.
\newblock Image{N}et classification with deep convolutional neural networks.
\newblock In \emph{NIPS}, 2012.

\bibitem[LeCun et~al.(1998)LeCun, Bottou, Orr, and Muller]{lecun-98b}
LeCun, Y., Bottou, L., Orr, G., and Muller, K.
\newblock Efficient backprop.
\newblock In \emph{Neural Networks: Tricks of the trade}. Springer, 1998.

\bibitem[Russakovsky et~al.(2015)Russakovsky, Deng, Su, Krause, Satheesh, Ma,
  Huang, Karpathy, Khosla, Bernstein, Berg, and Fei-Fei]{imagenet}
Russakovsky, Olga, Deng, Jia, Su, Hao, Krause, Jonathan, Satheesh, Sanjeev, Ma,
  Sean, Huang, Zhiheng, Karpathy, Andrej, Khosla, Aditya, Bernstein, Michael,
  Berg, Alexander~C., and Fei-Fei, Li.
\newblock {ImageNet} large scale visual recognition challenge.
\newblock \emph{IJCV}, 2015.

\bibitem[Saxe et~al.(2013)Saxe, McClelland, and Ganguli]{saxe2013exact}
Saxe, Andrew~M, McClelland, James~L, and Ganguli, Surya.
\newblock Exact solutions to the nonlinear dynamics of learning in deep linear
  neural networks.
\newblock \emph{arXiv preprint}, 2013.

\bibitem[Simonyan \& Zisserman(2015)Simonyan and Zisserman]{vgg}
Simonyan, Karen and Zisserman, Andrew.
\newblock Very deep convolutional networks for large-scale image recognition.
\newblock \emph{ICLR}, 2015.

\bibitem[Sussillo \& Abbot(2015)Sussillo and Abbot]{sussillo2014random}
Sussillo, David and Abbot, Larry.
\newblock Random walk initialization for training very deep feedforward
  networks.
\newblock \emph{ICLR}, 2015.

\bibitem[Szegedy et~al.(2015)Szegedy, Liu, Jia, Sermanet, Reed, Anguelov,
  Erhan, Vanhoucke, and Rabinovich]{googlenet}
Szegedy, Christian, Liu, Wei, Jia, Yangqing, Sermanet, Pierre, Reed, Scott,
  Anguelov, Dragomir, Erhan, Dumitru, Vanhoucke, Vincent, and Rabinovich,
  Andrew.
\newblock Going deeper with convolutions.
\newblock \emph{CVPR}, 2015.

\bibitem[Wang \& Gupta(2015)Wang and Gupta]{wang2015unsupervised}
Wang, Xiaolong and Gupta, Abhinav.
\newblock Unsupervised learning of visual representations using videos.
\newblock \emph{ICCV}, 2015.

\bibitem[Yosinski et~al.(2014)Yosinski, Clune, Bengio, and
  Lipson]{yosinski2014transferable}
Yosinski, Jason, Clune, Jeff, Bengio, Yoshua, and Lipson, Hod.
\newblock How transferable are features in deep neural networks?
\newblock In \emph{NIPS}, 2014.

\end{thebibliography}
